\documentclass{article}

\PassOptionsToPackage{numbers, sort&compress}{natbib} 

\usepackage[preprint]{neurips_2021}

\usepackage[utf8]{inputenc} 
\usepackage[T1]{fontenc}    
\usepackage{url}            
\usepackage{booktabs}       
\usepackage{amsfonts}       
\usepackage{nicefrac}       
\usepackage{microtype}      
\usepackage{dsfont}         
\usepackage{bm}             
\usepackage{xcolor}         

\usepackage[]{amsthm}       
\usepackage[]{amssymb}      
\usepackage[]{mathtools}    

\usepackage[many]{tcolorbox}
\tcolorboxenvironment{proof}{
  blanker,
  before skip=12pt,
  after skip=12pt,
  borderline west={0.4pt}{0.4pt}{black},
  breakable,
  left=8pt,
}

\usepackage[]{tikz}         
\usepackage{graphicx}       
\usepackage{wrapfig}        
\usepackage{subcaption}     
\usepackage{float}          
\usepackage[font={footnotesize}]{caption}
\usepackage{booktabs}       
\usepackage{refcount}       

\usepackage{pgfplots}
\DeclareUnicodeCharacter{2212}{−}
\usepgfplotslibrary{groupplots,dateplot}
\usetikzlibrary{patterns,shapes.arrows}
\pgfplotsset{compat=newest}

\usepackage[pdfencoding=auto, hypertexnames=false, psdextra]{hyperref} 
\usepackage[capitalise]{cleveref}    
\crefformat{footnote}{#2\footnotemark[#1]#3}

\usepackage{ulem}       
\newcommand{\renewtheorem}[1]{%
  \expandafter\let\csname #1\endcsname\relax
  \expandafter\let\csname c@#1\endcsname\relax
  \expandafter\let\csname end#1\endcsname\relax
  \newtheorem{#1}%
}

\theoremstyle{plain}

\renewtheorem{thm}{Theorem}
\renewtheorem{cor}{Corollary}[thm]
\renewtheorem{lem}{Lemma}[thm]
\renewtheorem{obs}{Observation}[thm]
\renewtheorem{fact}{Fact}[thm]
\renewtheorem{prop}[thm]{Proposition}
\renewtheorem{asm}{Assumption}

\theoremstyle{definition}

\renewtheorem{rem}[thm]{Remark}
\renewtheorem{rems}[thm]{Remarks}
\renewtheorem{exa}[thm]{Example}
\renewtheorem{exas}[thm]{Examples}
\renewtheorem{defi}[thm]{Definition}
\renewtheorem{subdefi}{Definition}[thm]
\renewtheorem{conv}[thm]{Convention}
\renewtheorem{conj}[thm]{Conjecture}
\renewtheorem{prob}[thm]{Problem}
\renewtheorem{oprob}[thm]{Open Problem}
\renewtheorem{algo}[thm]{Algorithm}
\renewtheorem{qu}[thm]{Question}
\renewtheorem{pty}[thm]{Property}
\renewtheorem{subcor}{Corollary}[thm]

\newcommand{\NB}{\emph{Note}}  
\newcommand{\eg}{\emph{e.g.}}  
\newcommand{\ie}{\emph{i.e.}}  
\newcommand{\etc}{\emph{etc.}} 
\newcommand{\cf}{\emph{cf. }}  
\newcommand{\viz}{\emph{viz.}} 

\newcommand{\define}{\coloneqq}

\renewcommand{\d}[2][{}]{\mathop{d^{#1}#2}}
\newcommand{\dd}[3][{}]{\frac{\d[#1]{#2}}{\d{{#3}^{#1}}}}
\newcommand{\pp}[3][{}]{\frac{\partial^{#1}{#2}}{\partial {#3}^{#1}}}

\newcommand{\avg}{\overline} 
\newcommand{\pr}{p} 

\DeclareMathOperator*{\argmax}{argmax}

\DeclareMathOperator{\sign}{sgn}

\newcommand{\RR}{\mathbf{R}} 

\newcommand{\seq}{\rvert_{{\rm eq}}}
\newcommand{\eq}{\Big\rvert_{{\rm eq}}}
\newcommand{\beq}{\Bigg\rvert_{{\rm eq}}}
\newcommand{\weq}{W_{\rm eq}}

\newcommand{\bNPF}{\textbf{F}\textsubscript{\textbf{NP}}}
\newcommand{\NPF}{F\textsubscript{NP}}
\newcommand{\bLTF}{\textbf{F}\textsubscript{\textbf{LT}}}
\newcommand{\LTF}{F\textsubscript{LT}}

\makeatletter
\newcommand\nocaption{%
    \renewcommand\p@subfigure{}
    \renewcommand\thesubfigure{\thefigure\alph{subfigure}}
}
\makeatother

\title{
  Unintended Selection: Persistent Qualification Rate Disparities and Interventions
}

\author{%
  Reilly Raab \\
  Computer Science and Engineering\\
  University of California, Santa Cruz\\
  Santa Cruz, CA 95064 \\
  \texttt{reilly@ucsc.edu} \\
  \And
  Yang Liu \\
  Computer Science and Engineering\\
  University of California, Santa Cruz\\
  Santa Cruz, CA 95064 \\
  \texttt{yangliu@ucsc.edu}
}

\begin{document}

\maketitle

\begin{abstract}
Realistically---and equitably---modeling the dynamics of group-level disparities
in machine learning remains an open problem. In particular, we desire models
that do \emph{not} suppose inherent differences between artificial groups of
people---but rather endogenize disparities by appeal to unequal \emph{initial
conditions} of insular subpopulations.  In this paper, agents each have a
real-valued feature $X$ (\eg, credit score) informed by a ``true'' binary label
$Y$ representing \emph{qualification} (\eg, for a loan).  Each agent alternately
(1) receives a binary classification label $\hat{Y}$ (\eg, loan approval) from a
Bayes-optimal machine learning classifier observing $X$ and (2) may update their
qualification $Y$ by imitating successful \emph{strategies} (\eg, seek a raise)
within an isolated group $G$ of agents to which they belong. We consider the
disparity of qualification rates $\Pr(Y=1)$ between different groups and how
this disparity changes subject to a sequence of Bayes-optimal classifiers
repeatedly retrained on the global population.  We model the evolving
qualification rates of each subpopulation (group) using the replicator equation,
which derives from a class of imitation processes.  We show that differences in
qualification rates between subpopulations can persist indefinitely for a set of
non-trivial equilibrium states due to uniformed classifier deployments, even
when groups are identical in all aspects except initial qualification densities.
We next simulate the effects of commonly proposed fairness interventions on this
dynamical system along with a new feedback control mechanism capable of
permanently eliminating group-level qualification rate disparities.  We conclude
by discussing the limitations of our model and findings and by outlining
potential future work.
\end{abstract}

\vspace{-0.1in}
\section{Introduction} \label{sec:introduction}

Algorithmic prediction is increasingly used for socially consequential decisions
and may determine individual access to information, education, employment,
credit, housing, medical treatment, freedom from incarceration, or freedom from
military targeting
\cite{obermeyer2019dissecting, Hao.2020, Metz.2020, Newton.2021, Hernandez.2021}.
This situation raises technical challenges and ethical concerns, particularly
regarding the dynamics of systemic inequalities and attendant harms to society
\cite{crawford2016there, chaney2018algorithmic, fuster2018predictably}.
Nonetheless, realistically---and equitably---modeling the dynamics of disparity
in machine learning remains an open problem.

Research historically considered the fairness of algorithmic predictions in
terms of statistical (in)consistencies
\cite{corbett2018measure}
(\eg, across groups
\cite{dwork2012fairness, zemel2013learning, hardt2016equality,
  zafar2017fairness, chouldechova2017fair, feldman2015certifying,
  kleinberg2016inherent} or between similar individuals
\cite{dwork2012fairness, zemel2013learning}),
preference guarantees
\cite{zafar17, ustun2019fairness, kusner2017counterfactual},
or causal considerations \cite{kusner2017counterfactual, kasy2021fairness} but
ignored the \emph{response} of a population to new prediction policies. For
instance, the proportions of potential loan applicants in each group that will
seek higher wages, falsify income, or forego application might \emph{change} if
banks use new policies to approve or deny loans, possibly counteracting fair
intent. We refer this class of fairness definitions as \emph{normative present
fairness}.

Efforts to model such population response
\cite{coate1993will, d2020fairness, zhang2020fair, heidari2019on,
  wen2019fairness, liu2019disparate, hu2018short, mouzannar2019fair,
  williams2019dynamic, liu2018delayed, hu2019disparate} and the autonomous
  dynamical systems arising from mutual recursion with myopically updating
  prediction policies
\cite{coate1993will, d2020fairness, zhang2020fair, heidari2019on,
  wen2019fairness, liu2019disparate, hu2018short, mouzannar2019fair,
  williams2019dynamic} have intensified, but it has remained to plausibly
  explain persistent disparities under group-independent prediction
  policies---\ie, those that do not discriminate on the basis of group
  membership---without assuming a setting that is structurally imbalanced
  between groups.  Our paper contributes to these efforts and considers the
  long-term consequences of machine learning on inter-group disparities when a
  sequence of classifiers induces dynamics within the rates of strategy adoption
  in each group.  Upon adopting a dynamical framework, we note that yet another
  operationalization of fairness arises: the asymptotic equality of latent
  variables (\ie, those causally responsible for outcome disparities) between
  groups. This notion of
\emph{long-term fairness} need not be consistent with \emph{normative present
fairness}, which may actively combat it, highlighting a tension
  between \emph{ends} and \emph{means} for fairness considerations.

\vspace{-0.05in}
\subsection{Our contributions}
\vspace{-0.05in}

Herein, we describe an \emph{equitable} model of population response: one which
does not suppose inherent differences between groups of people but endogenizes
disparities by appeal to unequal \emph{initial conditions}, accounting for
group-specific environmental conditions as dynamical variables.  We reform our
notion of ``groups'' (\ie, \emph{subpopulations}) to appeal to natural
boundaries of information exchange rather than artificially imposed classes of
people. We thus offer a potentially more meaningful way to group individuals in
discussions of fairness, asserting that, when considering such networks of peer
exchange, ``sensitive attributes'' such as race, sex, color, \etc{ }might
not correspond to meaningful divisions of people, which depend on social context.
Finally, we recognize fairness interventions as dynamical control policies that
(un)intentionally select the future trajectories of a given system.
We therefore allow ourselves to consider interventions that explicitly
incorporate \emph{feedback} from dynamical variables---rather than relying on
fixed, prescriptive modifications of predictor loss functions.

Our first contribution is to propose a model of \emph{classifier-induced}
group-level strategy adoption that is (1) equitable, \ie, free from structurally
asymmetric assumptions as described above, (2) capable of explaining persistent
disparities under Bayes-optimal, group-independent policies, and (3) derivable
from plausible, localized information exchange between individuals.
Specifically, we appeal to the replicator equation, an established model for
evolutionary phenomena without mutation, to model how competing \emph{strategies
for qualification} (which determine true machine learning labels \(\{0, 1\}\),
affecting agent utilities) replicate within groups (\ie, isolated subpopulations
that differ only in size and initial proportions of qualified individuals).

We ground statements with a running example involving loan applications
(elaborated upon in \cref{sec:replication}) for which qualification (label $Y =
1$), interpreted as being in the public interest, implies future repayment of a
loan  for an applicant with feature profile $X$. As we avoid assuming inherent
differences between groups, we consider the label-conditioned feature
distribution \(\Pr(X \mid Y = y)\) as group-independent and define qualification
disparity in terms of differences in group qualification rates $\Pr(Y = 1)$. We
formulate our model in \cref{sec:formulation}, emphasizing that only
the \emph{profile of strategies} in each subpopulation is subject to
evolution---narrowly qualified by the competition between strategies for
replicative success---rather than the subpopulations themselves. The persistence
of disparity is thus attributed to classifier policy.

Our second contribution, in \cref{sec:dynamics}, is a rigorous examination of
the dynamical system formed by the replicator equation and an updating,
group-independent, Bayes-optimal classifier policy, including a characterization
of its equilibrium states with linear stability analysis. We identify the set of
stable interior states of the system as a stable hyperplane and show that any
initial state with non-zero total qualification disparity, defined in
\cref{sec:dynamics}, will continue to exhibit non-zero disparity asymptotically
if the state attracts to the stable hyperplane (\cref{thm:yangsthm}). In this
sense, we claim that qualification rate disparity persists indefinitely for this
setting.

Our final contribution, in \cref{sec:interventions} is to consider a
dynamics-aware fairness intervention based on feedback control that
parametrically violates classifier group-independence (and therefore, in our
setting,
\emph{equalized odds}
\cite{hardt2016equality, zafar2017fairness, chouldechova2017fair}
and \emph{envy-freeness}
\cite{zafar17, ustun2019fairness})
to achieve long-term fairness. We use simulation to contrast this feedback
control policy to a group-independent classifier; a policy subject to
demographic parity \cite{dwork2012fairness, zemel2013learning}; and
``laissez-faire'', \emph{group-specific} policies. We conclude by discussing the
limitations of our model and our findings and by outlining potential future
work.

\vspace{-0.05in}
\subsection{Related work}
\vspace{-0.05in}

Our work chiefly contributes to the literature on fairness in machine learning
but also builds on prior work on ``statistical discrimination''.  The most
relevant publications are those that have highlighted the importance of studying
the dynamics and long-term consequences of machine learning, fairness
constraints, and models of population response.  In particular,
\citet{liu2018delayed}
use Markov transitions to model agent responses to classification without
considering classifier retraining;
\citet{d2020fairness} and \citet{zhang2020fair} reapply Markov transitions to agent attributes in the presense of classifier retraining;
\citet{zhang2019group}
model agents' decisions of whether to engage with classification based on
perceived accuracy and intra-group disparity;
\citet{coate1993will, hu2018short}, and \citet{liu2019disparate}
considered economical ``best-response'' models to agent labels with classifier
updates; and
\citet{heidari2019on}
considered an imitation-based model of social learning in which agents choose
between the strategies of other agents to maximize utility and minimize effort.
\citet{tang2020fair} also studied the delayed and accumulated impacts of past
deployed policies, but did not study the fairness implication of such impacts.
Similarly, literature on ``fair bandit/reinforcement learning"
\cite{joseph2016fairness,liu2017calibrated,jabbari2017fairness} has largely
focused on technical aspects of imposing normative present fairness in a
sequential setting.

Our proposed model synthesizes prior conceptual innovations: First,
\citet{coate1993will}, \citet{hu2018short}, and \citet{ensign2018runaway}
each considered incomplete information available to a classifier as a means to
equitably endogenize persistent predictor bias, but did not consider incomplete
information available to individual agents. Second, the class of response
functions considered by
\citet{mouzannar2019fair}
allows group-level strategic responses to depend on existing qualification rates
and may be used to endogenize persistent disparity under group-independent
policies; the cited work does not explore this direction, but, like us, the
authors assume ``groups are ex-ante equal in all respect except for their
qualification profiles...and any potential coupling between groups can only
happen through the different and interacting selection rates induced by the
policies'' \cite[p. 362]{mouzannar2019fair}. Atop this foundation, we provide a
plausible mechanism of imitation, motivated by incomplete information available
to individual agents, to justify replicator dynamics as a special case of such
response functions, and we extend a dynamical analysis for a classifier forced
to contend with misclassification errors.

To support our use of the replicator equation to model group-level responses to
classification, we cite the imitation-based derivation(s) of the replicator
equation by \citet{bjornerstedt1994nash}; the characterization of evolutionarily
stable strategies conducted by \citet{taylor1978evolutionary}; the analogy of
memes as attributed to \citet{dawkins1976selfish}; and the extensive application
of the replicator equation in game-theoretic contexts as explored by
\citet{friedman2016evolutionary}.

\section{Formulation} \label{sec:formulation}

\emph{We defer all proofs and provide them in \cref{Asec:proofs} of the
supplementary material.}

We consider countably many \emph{agents}, \(n \geq 2\) \emph{groups}, and a
single \emph{classifier}. Until \cref{sec:replication}, our setting matches that
of \citet{coate1993will} but treats \(n\) groups and a more granular classifier
utility function. We ground statements with a running example: a regional bank
(classifier) serving several isolated communities (groups, subpopulations) by
offering standardized loans for which every individual (agent) applies.
Alternative examples include hiring decisions \cite{coate1993will} or college
admissions.

Agents belong to \textbf{groups}, interpreted in \cref{sec:replication} and
consistent with \emph{isolated communities} in our running example, with known
relative frequencies \(\mu_{g} \in (0, 1)\). We vectorize these frequencies as
\(\boldsymbol{\mu}\).
\begin{align}
 \label{eq:def-mu}
 \mathcal{G} \define \{1, 2, ..., n\}; \quad
  \forall g\in\mathcal{G},~\mu_{g} &\define \Pr(G = g); \quad
  \sum_{g \in \mathcal{G}} \mu_{g} = 1; \quad
  \boldsymbol{\mu} \define (\mu_{1}, \mu_{2}, ..., \mu_{n})
\end{align}
\emph{For all statements of probability, we assign uniform probability mass to
each agent}.

In addition to relative size \(\mu_g\), each group has a \textbf{qualification
  rate} \(s_g \in (0, 1)\), which we vectorize as our \textbf{state} variable
\(\mathbf{s}\). We denote the global qualification rate as \(\avg{s}\):
\begin{align}
  \label{eq:def-sg}
  s_{g} \define \Pr(Y = 1 \mid G = g);\quad
  \mathbf{s} \define (s_{1}, s_{2}, ..., s_{n});\quad
  \avg{s} \define \sum_{g \in \mathcal{G}} \mu_{g} s_{g}
  = \langle \boldsymbol{\mu}, \mathbf{s} \rangle
\end{align}
\begin{asm} \label{asm:s-interior}
  No community is completely (un)qualified;
\(\quad \forall g,~s_{g} \in (0, 1)\).
\end{asm}

In our banking example, a qualified (\(Y = 1\)) individual will repay a loan in
full if accepted (\(\hat{Y} = 1\)), and we presume this outcome to be desirable.
The fraction of qualified individuals in community \(g\) is represented by
\(s_g\). \cref{asm:s-interior} states that no community is completely
(un)qualified, and, because \(\mu_{g} \in (0, 1)\), neither is  the total
population, \ie, \(\avg{s} \in (0, 1)\).

\begin{table}[ht]
  \caption{Agent-specific variables forming a Markov chain.}
  \label{table:variables}
  \centering
  \begin{tabular}{lllll}
    \toprule
    Variable & Meaning      & Domain    & Realizations  \\
    \midrule
    \(G\) & group & \(\mathcal{G} = \{1, 2, ..., n\}\) & \(g, h, i, j\) \\
    \(Y\) & qualification & \(\{0, 1\}\) \ie, \{unqualified, qualified\} & \(y\)  \\
    \(X\) & feature & \((-\infty, \infty)\) & \(x\) \\
    \(\hat{Y}\) & classification & \(\{0, 1\}\) \ie, \{reject, accept\} & \(\hat{y}\) \\
    \bottomrule
  \end{tabular}
\end{table}

The \textbf{feature} \(X\) of an agent qualified as (\(Y = y\)) is sampled
according to a probability density function \(q_y\). In our banking example, we
may interpret \(X\) as a ``credit score'' known to the bank.
\begin{align} \label{eq:def-q}
  q_{y}(x) &\define \pr_{X}(x \mid Y = y); \quad y \in \{0, 1\}
\end{align}
\begin{asm} \label{asm:q-group-independent}
  The qualification-conditioned distribution of features \(q_y(x)\) is
  group-independent.
\end{asm}
\vspace{-0.05in} \cref{asm:q-group-independent} ensures that qualified
individuals are statistically indistinguishable in terms of feature \(X\) across
different communities---as are unqualified individuals. Given an agent's
qualification \(Y\), learning \(G\) gives no additional information about \(X\).

\begin{asm} \label{asm:q1-q0-monotonic}
  \(q_y\) is differentiable and strictly positive for each \(y\). The values of
  \(X\) are ordered and unified such that \(q_{1}(x) / q_{0}(x)\) is strictly
  increasing in \(X\):
  \begin{align} \label{eq:q1-q0-monotonic}
    \forall x, y,~ q_{y}(x) \in (0, \infty); \quad \dd{}{x} \bigg(
    \frac{q_{1}(x)}{q_{0}(x)} \bigg) > 0
  \end{align}
\end{asm}
\cref{asm:q1-q0-monotonic} ensures that the feature \(X\) is ``well-behaved'':
In our example, as credit scores increase as \(x\), the odds that individuals
with that credit score \(x\) will pay off loans also increases.

Finally, a \textbf{classifier} observes the feature \(X\) of each agent, from
which it must predict the agent's correct label \(Y\) using a deterministic
policy \(\pi\) that, unless otherwise stated, remains ignorant of \(G\).
\begin{asm} \label{asm:learning}
  The classifier learns the true distribution \(\Pr(Y \mid X)\) before choosing
  policy \(\pi\).\footnote{In practice, this distribution may be learned from
  sufficient data.}
\end{asm}
\begin{asm} \label{asm:u-linear}
  The classifier maximizes its expected utility \(u\) with risk-neutral
  preferences.  This utility \(u\) is linear in each outcome fraction \(\Pr(Y =
  y, \hat{Y} = \hat{y})\), and the coefficients\footnote{We will abuse notation
  to write, \eg, \(V_{y,\hat{y}}\) as \(V_{1 \hat{0}}\), to disambiguate the
  order of indices on \(V\) and, later, \(U\).} \(V_{y, \hat{y}} \in(-\infty,
  \infty)\) are independent of feature value \(X\) and group membership \(G\).
  The classifier receives higher utility from correct predictions (\(\hat{Y} =
  Y\)).
  \begin{align}
    \hat{Y} \define \pi(X)
    ; \quad
    u(\pi) &\define \sum_{y,\hat{y} = 0}^{1} V_{y, \hat{y}} \Pr(Y = y, \pi(X) = \hat{y})
    ; \quad
    V_{y = \hat{y}} > V_{y \neq \hat{y}}
  \end{align}
\end{asm}
\vspace{-0.05in} In our example, \cref{asm:u-linear} is consistent with a bank
maximizing expected net profit, where the bank expects net profit proportional
to \(V_{y, \hat{y}}\) from each individual qualified as \(y\) and approved as
\(\hat{y}\), independent of credit score \(X\) or community \(G\).  By
\cref{asm:learning}, the bank selects policy \(\pi\) knowing the stochastic
relationship between qualification \(Y\) and credit score \(X\) for the region
it serves.

With group-independent classifier policies, having excised assumptions of
inherent differences between groups in our formulation, we emphasize that
unequal group qualification rates cause any statistical group-level disparities
of prediction outcomes. We therefore consider eliminating differences in group
qualification rates as a realization of long-term fairness in this setting.

\begin{thm}\label{thm:phi-threshold}
  Discounting sets of measure zero, the \(u\)-maximizing, group-independent
  policy \(\pi\) is parameterized by the \textbf{feature threshold} \(\phi \in
  [-\infty, \infty]\) such that \(\pi(x) = 1\) if and only if \(x > \phi\),
  where \(\phi\) depends only on the global qualification rate \(\bar{s}\).
\begin{align} \label{eq:q1-q0-s-theta}
  \hat{y} = \pi(x) = \begin{cases}
  1 & x > \phi \\
  0 & \text{otherwise} \end{cases}
  ; \quad
  \frac{q_{1}(\phi)}{q_{0}(\phi)} = \xi \cdot \frac{1 - \avg{s}}{\avg{s}}
  ; \quad
  \xi \define \dfrac{V_{0\hat{0}} - V_{0\hat{1}}}{V_{1\hat{1}} - V_{1\hat{0}}}
 \end{align}
When a solution in \(\phi\) to the \textbf{threshold equation},
\cref{eq:q1-q0-s-theta}, does not exist, \(\phi\) is either \(\pm \infty\).
\end{thm}
\newcommand{\corddphis}{ The classifier's feature threshold \(\phi\) responds
  inversely to \(\avg{s}\): \(\dd{\phi}{\avg{s}} < 0, ~~~ \dd{\avg{s}}{\phi} <
  0.\) }
\begin{cor}\label{cor:ddphis}
\corddphis
\end{cor}
\begin{asm} \label{asm:xi-domain}
  \(V\) is such that \( \xi \in (0, \infty) \) ($\xi $ is defined in Theorem
  \ref{thm:phi-threshold}, \cref{eq:q1-q0-s-theta}).
\end{asm}

The threshold equation, \cref{eq:q1-q0-s-theta}, is a reprise of
\citet{coate1993will} restricted to group-independent classifier policies. In
our example, the bank maximizes its utility by approving individuals with credit
scores greater than \(\phi\) and denying everyone else. Interpreting
\cref{asm:xi-domain}, there exist populations for which the bank prefers to
accept some applicants and reject others.

\subsection{Time-dependence}

We model our system in discrete time for semantic reasons, acknowledging that a
learning process consistent with \cref{asm:learning} requires time, although the
mathematics generalize to continuous time without issue.\footnote{The continuous
replicator equation appears in \citet{bjornerstedt1994nash} \&
\citet{friedman2016evolutionary}.}  Where required, we will denote
time-dependence in square brackets \([t]\). Where we omit this explicit
dependence, as in all prior expressions, it is understood that all variables in
an expression correspond to the same time \(t\).

\begin{asm}
  The relative sizes of groups \(\mu_g\), qualification-conditioned feature
  distributions \(q_y\), and classifier utility coefficients \(V_{y,\hat{y}}\)
  are all time-independent. All prior assumptions hold independently for each
  time step.
\end{asm}

\subsection{Replicator dynamics} \label{sec:replication}

Anticipating algorithmic classification, how do agents decide whether to become
qualified? In our banking example, we imagine that individuals must ``invest''
or ``apply capital'' to be able to pay back loans and that the ``rationality''
of doing so depends on what they know about the classification policy and
potential outcomes---which they must estimate from incomplete information
provided by peer examples. To deal with the uncertainty of limited examples, we
imagine the emergent heuristic of updating personal qualification by
\emph{imitating} the strategies of others based on popularity and ``success''.
For example, if your friend chose to become qualified for a loan and now runs a
small business, the success of the business may induce you to seek qualification
yourself by first building credit history; if many of your neighbors receive
loans despite being unqualified and appear successful investing in speculative
assets, you may infer that qualification is a waste of resources.

\begin{figure}[ht]
  \centering
  \includegraphics[width=\textwidth, trim=13 3 3 3, clip]{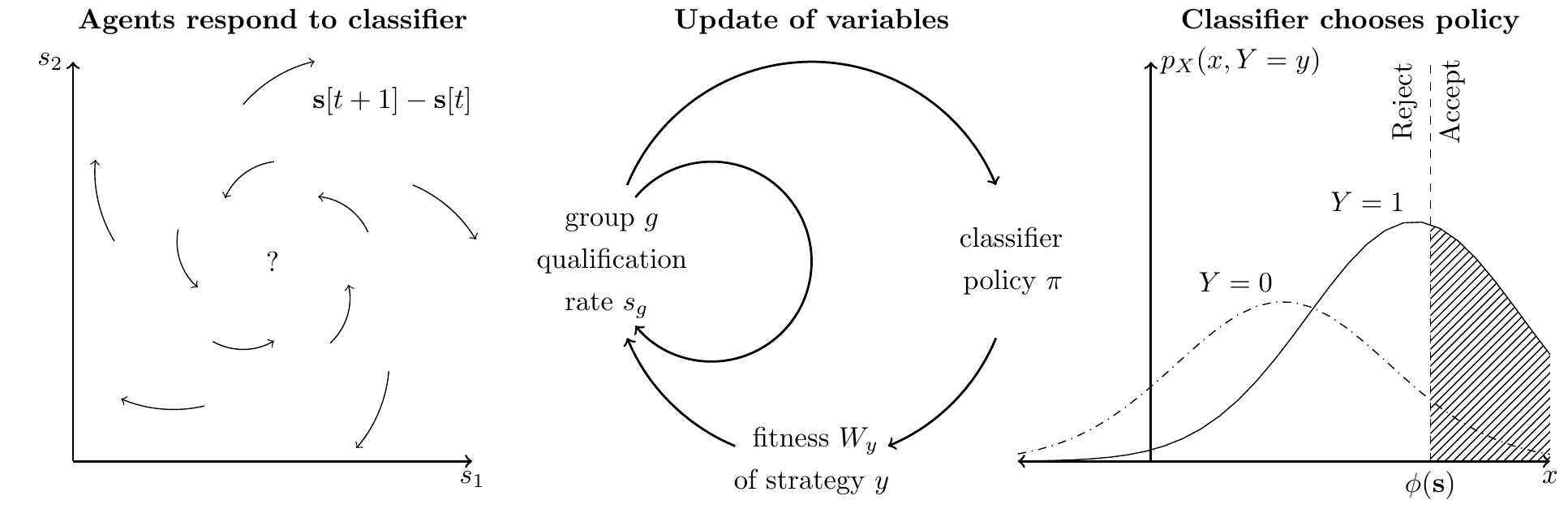}
  \caption{
    Our model appeals to the replicator equation \cref{eq:replicator} to model
    population response and considers a Bayes-optimal, group-independent
    classifier policy \(\pi\) with feature threshold \(\phi\)
    (\cref{eq:q1-q0-s-theta}, right pane). When coupled (middle pane), these
    equations give rise to an autonomous dynamical system. We wish to understand
    how the vector of group qualification rates \(\mathbf{s}\), as our state
    variable, changes in time (left pane).}
\label{fig:overview}
\end{figure}

\citet{bjornerstedt1994nash} have shown that \emph{imitation} in this form,
whereby agents stochastically update to strategies weighted by success and
popularity\footnote{Weighting by popularity effects ``preferential attachment''
for the success of a strategy in a subpopulation and thus introduces dynamical
inertia.  Replicator dynamics also arise when agents update strategies with
(Poisson distributed) expected periodicity that is affine in the success of
one's current strategy \cite{bjornerstedt1994nash}.}  yields the (continuous
time) \textbf{replicator equation}, which we use in its discrete time form, as
detailed by \citet{friedman2016evolutionary}:
\begin{align}
  s_{g}[t + 1] = s_{g}[t] \frac{W_{1}[t]}{\avg{W}_{g}[t]} \label{eq:replicator}; \quad
  \avg{W}_{g} &\define W_{1} s_{g} + W_{0} (1 - s_{g}); \quad \forall y,~W_y \geq 0
\end{align}
Here, \(W_{y}\) is the \textbf{fitness} of strategy \(y\), which, by
\cref{thm:WQ}, we model as independent of feature \(X\) and group \(G\).
Following \citet{bjornerstedt1994nash}, we may derive the fitness \(W_y\) in
terms of expected ``success'' \(U_{y, \hat{y}}\) of each
qualification-classification outcome \((y, \hat{y})\):
\begin{asm} \label{asm:W-group-independent}
  The fitness of strategy \((Y = y)\), denoted as \(W_{y}^{g}\), is affine in
  the average success \(U_{y, \hat{y}}\) of qualification \((Y = y)\) with
  classification \((\hat{Y} = \hat{y})\). \(U_{y, \hat{y}}\) is time-, feature-
  and group-independent. Without loss of generality, we restrict \(U_{y,
    \hat{y}} \in [0, \infty)\) and drop the constant bias term from each
    \(W_{y}^{g}\).
  \begin{align} \label{eq:W-group-independent}
    W_{y}^{g} &\define
    \sum_{\hat{y}=0}^{1} \Pr(\hat{Y} = \hat{y} \mid Y = y, G = g) U_{y, \hat{y}}
  \end{align}
\end{asm}
\newcommand{\thmWQ}{
    The fitness \(W_{y}^{g}\) of strategy \(Y = y\) in group \(g\) is
    feature- and group-independent.
  \begin{align}
    \forall y, g,~W_{y}^{g} = W_{y}
  \end{align}
}
\begin{thm} \label{thm:WQ}
\thmWQ
\end{thm}
\vspace{-0.05in}
Intuitively, ``success'' \(U\) may be interpreted as \emph{utility} or
\emph{payoff} to each agent when agents align strategy adoption with personal
incentives, but, fundamentally, \(W\) corresponds to the relative success of the
\emph{strategy} in replicating, \ie, spreading between individuals. The
strategies of (non)qualification are thus subject to evolutionary pressures,
competing to out-replicate each other in an environment shaped by perceptions of
classifier policy. Notably, the fitness of a strategy depends only on the
classifier---not group membership \(G\). Agents remain identically modelled
across all groups.
\begin{asm} \label{asm:UUneq}
  The success of (non)qualification is sensitive to classification, and the
  expected success for qualified individuals increases with classifier
  acceptance:
  \(
  U_{0\hat{1}} \neq U_{0\hat{0}};~~ U_{1\hat{1}} > U_{1\hat{0}}~.
\)
\end{asm}
\begin{asm}\label{asm:isolated}
  Each group \(g\) has the properties of a closed population in which
  \emph{qualification}, as a strategy or meme \cite{dawkins1976selfish},
  competes with \emph{non-qualification} free from exchange with other groups.
\end{asm}
\vspace{-0.05in}
It is significant that we model population updates as independent for
\emph{closed} populations, as this restricts our interpretation of groups, which
must be functionally impermeable to the exchange of qualification strategies. To
precisely delineate between real-world examples of ``groups'' is akin to
disassociating ``cultures'', which also imply boundaries of exchange but
generally intersect. Having noted that ``sensitive attributes'' such as race,
sex, color, \etc{ }may not correspond to meaningful divisions between people
(which depend on social context), we instead qualify a \textbf{group} by the
extent to which it satisfies \cref{asm:isolated}. As an open question, we ask
whether imposing arbitrary demographic-dependent policies may catalyze the
formation of groups of strategic peers, but we will consider insular social
groups or isolated communities as canonical examples.
\vspace{-0.05in}

\section{Dynamics} \label{sec:dynamics}

The threshold equation, \cref{eq:q1-q0-s-theta}: \(\phi[t](\mathbf{s}[t])\), and
the replicator equation, \cref{eq:replicator}: \(\mathbf{s}[t+1](\phi[t],
\mathbf{s}[t])\), may be coupled to yield an autonomous dynamical system
\(\mathbf{s}[t+1](\mathbf{s}[t])\) that evolves in time.  To analyze it, we
first generate a useful set of coordinates to compliment \(\avg{s}\) and track
qualification rate disparities, defined by the differences in \(s_g\) between
groups.  We then note the importance of \(W_{1}(\phi) - W_{0}(\phi)\) to the
overall dynamics of the system, and use it to identify non-trivial equilibrium
states.  To interpret this section for our example, we ask how
community-specific loan qualification rates change as individuals imitate
successful strategies in their isolated communities, while assuming that the
bank maximizes profit using group-independent credit thresholds for loan
approval.

\begin{defi} Define the (signed) \textbf{qualification distance} from group \(h\)
  to group \(g\) as
  \begin{align} \label{eq:def-delta}
   \delta(g, h) \define s_{g} - s_{h}, \quad g, h \in \{1, 2, ..., n\}
  \end{align}
\end{defi}
We next define the vector \(D\) comprising \((n-1)\) linearly-independent
qualification distances between sequential pairs of subpopulations:
\begin{align}
  \label{eq:Dcoords}
  D \define \Big( \delta(1, 2), ~\delta(2, 3), ~...,~ \delta(n - 1, n) \Big)
\end{align}
The components of \(D\) and value of \(\avg{s}\) together yield a complete set
of coordinates to describe the state of the dynamical system, which we may
exchange for the original vector of qualification rates \(\mathbf{s} = (s_{1},
s_{2}, ..., s_{n})\) via a non-orthogonal, linear change of basis (See
\cref{Asec:proofs}):
\begin{align} \label{eq:changecoords}
    s_{g}
    &= \avg{s} + \sum_{h=g}^{n-1} \delta(h, h+1)
    - \sum_{h=1}^{n-1} \sum_{k=1}^{h} \mu_{k} \delta(h, h + 1)
      \quad \forall g \in \mathcal{G}
\end{align}
Let us denote the state vector in our new coordinate system as \(\mathbf{r}
\define \big(\delta(1,2), \delta(2,3),...,\delta(n-1,n), \avg{s}\big)\).

\begin{subdefi}
  For \(p \geq 1\), define a state's \textbf{\(\boldsymbol{p}\)-total
    qualification rate disparity} as the \(p\)-norm of \(D\):
  \begin{equation}
    \big\|D\big\|_p \define
    \Big( \sum_{g=1}^{n-1} \big|\delta(g,g+1)\big|^p \Big)^{1/p}
  \end{equation}
\end{subdefi}

\newcommand{\remhyperplane}{
States \(\mathbf{s}\) with a common \(\avg{s}\) value form a \textbf{hyperplane}
\(\avg{s} = \langle \boldsymbol{\mu}, \mathbf{s}\rangle\) (\cref{eq:def-sg}), by
definition.}
\begin{rem}\label{rem:hyperplane}
\remhyperplane
\end{rem}

\newcommand{\rempersistence}
{
  The nullity of any \(p\)-total qualification rate disparity is preserved in
  time.
  \begin{align}
    p \geq 1; \quad
\big\|D[t]\big\|_p = 0 \iff \big\|D[t + 1]\big\|_p = 0
  \end{align}
}
\begin{thm} \label{rem:persistence}
\rempersistence
\end{thm}
\vspace{-0.1in}
\cref{rem:persistence} highlights a weak notion of the persistence of disparity
within the system sans intervention: Any state that possesses some non-zero
total qualification disparity (defined as some chosen \(p\)-norm of \(D\)) must
always exhibit some non-zero total qualification disparity with any finite time
horizon. In our example, if some communities start more qualified than others,
the qualification rates of different communities will not naturally equalize in
any given lifetime. Note that this statement is insufficient to address the
limit \(t \to \infty\), however. For a stronger result that includes this limit
(\cref{thm:yangsthm}), we first characterize the system's equilibrium states.

\subsection{Equilibrium}

\begin{defi} \label{defi:equilibrium}
The system as a whole is \textbf{at equilibrium} when, for all \(g \in
\mathcal{G}\) simultaneously, \(s_{g}\) is stationary in time:
\vspace{-0.1in}
\begin{align} \label{eq:def-eq}
  \text{at equilibrium} \overset{\text{def}}{\iff} \forall g \in \mathcal{G},~~
  \exists t_{0} ~\text{s.t.}~\forall t \geq t_{0},\quad s_{g}[t] = s_{g}[t_0]
\end{align}
\end{defi}
\NB: replicator dynamics is an instance of the more general family of monotone
dynamics, with which all equilibria are shared \cite{bjornerstedt1994nash,
  friedman2016evolutionary}.
\newcommand{\thmppstplus}{
  Disregarding boundary states by \cref{asm:s-interior}, the replicator
  equation, \cref{eq:replicator}, implies
  \begin{equation}
  \begin{aligned}
    \sign\big(\avg{s}[t + 1] - \avg{s}[t] \big)
    = \sign\big(W_{1}(\phi [t]) - W_{0}(\phi [t])\big)
  \end{aligned}
  \end{equation}
}
\begin{thm}\label{thm:ppstplus}
  \thmppstplus
\end{thm}
\newcommand{\thmeqWW}{
  It is necessary and sufficient for a system at equilibrium that \(W_{1} =
  W_{0}$ or for the system to occupy some vertex of the state space.
\begin{align}
  \text{\rm at equilibrium} \iff \begin{cases}
    W_{1} = W_{0} & \text{(internal equilibrium)} \\
    \forall g \in \mathcal{G}, \quad s_{g} \in \{0, 1\}
    & \text{(trivial equilibrium)}
  \end{cases}
\end{align}
}
\begin{thm} \label{thm:eqW1W0}
  \thmeqWW
\end{thm}
\begin{wrapfigure}{r}{0.4\textwidth}
  \centering
  \includegraphics[width=0.4\textwidth]{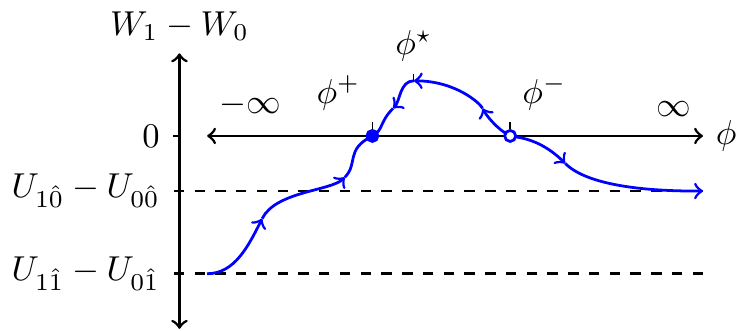}
    \caption{$W_{1}(\phi) - W_{0}(\phi)$ (blue curve) is a strictly
      quasi-concave function of $\phi$. $\phi^{\star}$ denotes the unique local
      extremum.  The direction of the arrows is a consequence of
      \cref{thm:ppstplus} and \cref{cor:ddphis}.  }
  \vspace{-0.05in}
  \label{fig:zeros}
\end{wrapfigure}

\cref{thm:eqW1W0} indicates that the conditions for internal equilibrium are
described by the \emph{zeros} of the function \(W_{1}(\phi) - W_{0}(\phi)\), as
depicted in \cref{fig:zeros}, and, by the threshold equation,
\cref{eq:q1-q0-s-theta}, \(\phi\) has dynamical dependence only on \(\avg{s}\).
It follows that only certain values of \(\avg{s}\) support internal equilibrium,
and each value corresponds to a hyperplane in state space
(\cref{rem:hyperplane}).

\newcommand{\WWquasi}{
  \(W_{1}(\phi) - W_{0}(\phi)\) is strictly quasi-concave in \(\phi\).
  This guarantees that no more than two zeros of the function \(W_{1} - W_{0}$
  exist.
}
\begin{thm}\label{thm:WW-quasi}
\WWquasi
\end{thm}
We denote the possible zeros of \(W_1 - W_0\) as \(\phi^+\) and \(\phi^-\),
where the sign in the superscript indicates the local slope of the function.
These zeroes correspond to parallel hyperplanes in state space that comprise all
interior equilibria of the system. Whether \(\phi^\pm\) corresponds to an
(un)stable equilibrium hyperplane may be determined by \cref{thm:ppstplus} and
the sign of \(\pp{}{\phi}(W_{1} - W_{0})\): Only \(\avg{s}(\phi^{+})\) is
stable, and we will verify this fact with linear stability analysis.
\newcommand{\thmyangsthm}{
    If the state of the system asymptotically approaches an internal
    equilibrium, the nullity of $p$-total qualification rate disparity is
    preserved in the limit of infinite time.
    \begin{equation}
      \begin{aligned}
        p \geq 1; \quad \lim_{t' \to \infty} (W_{1} - W_{0}) = 0 \implies
        \Big(
        \big\|D[t]\big\|_p = 0 \iff
        \lim_{t' \to \infty} \big\|D[t']\big\|_p = 0 \Big)
      \end{aligned}
    \end{equation}
}

\begin{thm} \label{thm:yangsthm}
\thmyangsthm
\end{thm}

\cref{thm:yangsthm} formalizes the critical observation that any state that
attracts to the stable equilibrium hyperplane, unless initially free from
qualification disparity, will forever exhibit some total qualification
disparity. This is a more robust notion of the \textbf{persistence of disparity}
in our system than \cref{rem:persistence}.

\subsection{Stability}

For our regional banking example, we may imagine that qualification rates settle
into a stable pattern in which some communities have a higher average
qualification rate than others. How robust is this pattern of inequality to
small fluctuations of qualification rates?  Using \textbf{linear stability
  analysis} (\ie, linearizing the response of the system to small perturbations
about equilibrium and asking ``do perturbations amplify or dissipate?''), we
show that only the \(\phi^+\)-hyperplane acts as a stable attractor.

First, let us denote the evaluation of an expression \emph{at equilibrium} by
placing a vertical line to the right of the expression with ``eq'' as a
subscript. In light of \cref{thm:eqW1W0} and \cref{eq:replicator}, we also
introduce the shorthand \(\weq\) to denote an equilibrium value of \(W_{1}\),
\(W_{0}\), or, equivalently, any \(\avg{W}_{g}\). It should be noted that the
value of \(\weq \in [0, \infty)\) still depends on the particular equilibrium
  state of the system.
\begin{align}
  \weq &\define W_{0} \eq = W_{1} \eq = \avg{W}_{g} \eq
  \quad \forall g \in \mathcal{G}
\end{align}
We linearize the system at equilibrium by constructing the Jacobian \(J \in
\RR^{n \times n}\) corresponding to discrete time-evolution and identifying its
eigenvectors and eigenvalues:
\begin{align}
  J &\define \begin{bsmallmatrix}
    \displaystyle
    \pp{\mathbf{r}}{\delta(1,2)} &
    \displaystyle
    \pp{\mathbf{r}}{\delta(2,3)}&
    \displaystyle
    ... &
    \displaystyle
    \pp{\mathbf{r}}{\delta(n-1,n)} &
    \displaystyle
    \pp{\mathbf{r}}{\avg{s}}
  \end{bsmallmatrix}
\end{align}
where \(\mathbf{r}\), the state vector in \((D, \avg{s})\) coordinates, is
interpreted as a column vector.

\newcommand{\thmjacobian}{
    The Jacobian \(J\) simplifies to a scalar multiplied by a matrix with a
    single non-zero column \(\mathbf{v}\) in the last position.
  \begin{align} \label{eq:jacobian}
    J \beq
    =\frac{1}{\weq}
      \bigg( \dd{\phi}{\avg{s}}\bigg)
                 \bigg( \dd{}{\phi} (W_{1} - W_{0}) \bigg)
      \Bigg[ \mathbf{0}^{(n \times n - 1)} \Bigg| \mathbf{v }\Bigg], ~~~
    \mathbf{v}
    \define \begin{bsmallmatrix}
    \delta(1,2)(1 - s_{1} - s_{2})\\
    \delta(2,3)(1 - s_{2} - s_{3})\\
    ... \\
    \delta(n-1,n)(1 - s_{n-1} - s_{n})\\
     \sum_{g \in \mathcal{G}} \mu_{g} s_{g}( 1 - s_{g} )
  \end{bsmallmatrix}
  \end{align}
}
\begin{thm} \label{thm:jacobian}
\thmjacobian
\end{thm}
The eigenvalues of \(J\) determine the stability of the system at equilibrium.
\begin{cor} \label{cor:jacob-inspect}
  At equilibrium, any state displacement vector with zero \(\avg{s}\) component
  is an eigenvector of \(J\) with eigenvalue 0, while \(\mathbf{v}\) is an
  eigenvector of \(J\) with eigenvalue \(\lambda\):
  \begin{align}\label{eq:lambda}
    \lambda \define \bigg( \sum_{g \in \mathcal{G}} \mu_{g} s_{g}( 1 - s_{g} )\bigg)
    \frac{1}{\weq} \bigg( \dd{\phi}{\avg{s}}\bigg)
    \bigg( \dd{}{\phi} (W_{1} - W_{0}) \bigg) \beq
  \end{align}
\end{cor}
Perturbing (displacing) a state vector \(\mathbf{r}\) at an internal equilibrium
by altering any combination of coordinates appearing in \(D\)---while leaving
\(\avg{s}\) fixed---specifies motion on the \(\avg{s}\) hyperplane occurring in
neutrally stable equilibrium (\ie, a displacement vector with zero \(\avg{s}\)
component has null eigenvalues at internal equilibrium. See
\citet{strogatz2018nonlinear}).  An internal equilibrium is stable to
perturbations in \(\mathbf{v}\), leaving the hyperplane, iff \(\lambda\) is
negative (and, in discrete-time, \(>-2\) to forbid over-corrections)
\cite{strogatz2018nonlinear}.

\newcommand{\corphiplusstable}{
As a consequence of \cref{cor:ddphis}, which states \(\dd{\phi}{\avg{s}} < 0\),
the eigenvalue \(\lambda\) in \cref{eq:lambda} is negative, (and the associated
equilibrium hyperplane stable) iff
\(
    \dd{}{\phi}(W_{1} - W_{0}) \seq > 0
\).
This prescribes precisely the value \(\phi^{+}\) for the stable equilibrium
hyperplane.
}
\begin{cor} \label{cor:phiplusstable}
  \corphiplusstable
\end{cor}

\section{Interventions} \label{sec:interventions}

In the dynamical setting we have characterized, we now explore ``fairness
interventions'', which substitute the set of policies that the classifier may
choose from, possibly permitting group-specific decision rules \(\pi_g\). We
first observe that for the default policy with a group-independent feature
threshold \(\phi\), one commonly cited standard of normative present fairness is
automatically satisfied.
\begin{defi}\label{defi:eo}
  \textbf{Equalized Odds}
  \cite{hardt2016equality, zafar2017fairness, chouldechova2017fair} requires
  that a classifier's decisions \(\hat{Y}\), given by policy \(\pi\),
  misclassify (un)qualified agents at equal rates across groups:
  \begin{align}
    \forall g, h \in \mathcal{G},~ \forall y, \hat{y} \in \{0, 1\}, \quad
    \Pr(\hat{Y} = \hat{y} \mid Y = y, G = g) = \Pr(\hat{Y} = \hat{y} \mid Y = y,
    G = h)
  \end{align}
\end{defi}
\newcommand{\thmeqopt}{
For policies defined by group-specific thresholds \(\phi_g\), the equivalence of
these feature thresholds (\(\forall g, \phi_g = \phi\)) is necessary and
sufficient to satisfy Equalized Odds given the group-independence of each
\(q_y\) (\cref{asm:q-group-independent}).
}
\begin{thm} \label{thm:eq-opt}
\thmeqopt
\end{thm}
\vspace{-0.05in}
By \cref{thm:eq-opt}, a group-independent policy satisfies Equalized Odds (\eg,
the bank accepts/rejects (un)qualified loan applicants at group-independent
rates), yet disparities may persist (\cref{thm:yangsthm}). This indicates a
counter-example to reliance on Equalized Odds for long-term fairness in our
model, \viz, the optimal group-independent threshold classifier we have studied
so far.
\begin{subcor} \label{cor:eq-odds}
Equalized Odds does not imply long-term fairness in our model.
\end{subcor}
\vspace{-0.05in}
We next ask whether a small displacement a group-independent threshold \(\phi\)
near the $\phi^{+}$-hyperplane, which we interpret as a \textbf{universal
  subsidy} (or penalty), can diminish qualification rate disparities.
\newcommand{\thmintervention}{
   \(\Theta(\epsilon)\) perturbations of a group-independent $\phi$ at internal
   equilibrium induce motion, which, to first-order approximation (\ie, ignoring
   \(\mathcal{O}(\epsilon^2)\) terms), is parallel to the eigenvector
   $\mathbf{v}$.  }
\begin{thm} \label{thm:intervention}
\thmintervention
\end{thm}
As a consequence of \cref{thm:intervention}, while \(\mathbf{v}\) need not be
orthogonal to the equilibrium hyperplane, and a universal subsidy may decrease
qualification rate disparity while applied (settling on a new equilibrium
hyperplane with different, though persistent disparities), the system is stable
to such perturbations at \(\phi^+\) as characterized by linear system response
and will relax to the original equilibrium state when the intervention is
removed. To permanently change qualification disparities, a temporary universal
subsidy (penalty) must rely on the \textbf{non-linear response} of the system
and is therefore liable to require \emph{large} perturbations to the
classifier's threshold \(\phi\). This finding compels us to consider
interventions with group-dependent threshold perturbations---or group-dependent
thresholds. To this end, we hereafter generalize our classifier such that it
independently classifies each group \(g\) according to a group-specific
threshold \(\phi_g\). We denote the vector of these thresholds as \(\Phi \define
(\phi_{1}, \phi_{2}, ..., \phi_{n})\) and assume that, prior to some
perturbative intervention, \(\phi_{g} = \phi\) for each \(g \in \mathcal{G}\).

\begin{defi}\label{defi:dp}
  \textbf{Demographic Parity} \cite{dwork2012fairness, zemel2013learning}
  requires that a classifier's decisions \(\hat{Y}\), given by policy \(\pi\),
  are positive (\(\hat{Y} = 1\), \eg, accepting a loan application) at equal
  rates for all groups:
  \vspace{-0.05in}
  \begin{align}
    \forall g, h \in \mathcal{G}, \quad
    \Pr(\hat{Y} = 1 \mid G = g) = \Pr(\hat{Y} = 1 \mid G = h)
  \end{align}
\end{defi}
\begin{defi}\label{defi:lz}
  \textbf{Laissez-Faire} allows a separate, \(u\)-maximizing  threshold
  \(\phi_g\) for each group.
\end{defi}
\newcommand{\thmdemparity}{
  Demographic parity requires sign-heterogeneous, group-dependent changes to the
  Laissez-Faire values of \(\phi_g\)  when \(\pi\) is non-trivial (does not
  uniformly accept (reject)).
}
\begin{thm} \label{thm:dem-parity}
\thmdemparity
\end{thm}
\vspace{-0.05in}
Satisfying demographic parity in our setting requires the solution of a
differential equation in $q_{y}$ (\cref{Asec:proofs}), which we do not rigidly
constrain. We therefore rely on numerical simulation, rather than analytical
tools, to evaluate this intervention for our system.

\vspace{-0.1in}
\paragraph{Feedback control}
Arbitrary state transitions in the equilibrium hyperplane may be permanently
effected by group-dependent perturbations to $\Phi$, which we derive from linear
system response at equilibrium. Specifically, to diminish a specific
qualification distance $\delta(g, g+1)$ for given $g$, $\Phi$ may be perturbed
by a vector quantity
\(\Delta_{g}\Phi = (\Delta_g \phi_{1}, \Delta_g \phi_{2}, ..., \Delta_g \phi_{n})\).
\newcommand{\thmhyperplaneintervention}{
    On the stable internal equilibrium hyperplane, infinitesimal perturbation of
    $\Phi$ by
    \vspace{-0.05in}
\begin{subequations}
  \begin{align}
  \Delta_{g} \Phi &\define -\epsilon  \delta(g, g + 1) \Big(
  \frac{\alpha_{g}}{s_{1}(1 - s_{1})},
  ...,
  \frac{\alpha_{g}}{s_{g}(1 - s_{g})},
  \frac{\beta_{g}}{s_{g+1}(1 - s_{g+1})},
  ...,
  \frac{\beta_{g}}{s_{n}(1 - s_{n})}
                    \Big) \\
    &\begin{aligned}
  \alpha_{g} &\define (\mu_{g+1} + \mu_{g+2} + ... + \mu_{n}), \quad &
  \beta_{g} &\define -(\mu_{1} + \mu_{2} + ... + \mu_{g})
  \end{aligned}
  \end{align}
\end{subequations}
will induce motion in the system preserving $\avg{s}$ and each $\delta(h, h +
1)$ for $h \neq g$. The value of $\delta(g, g+1)$ will be diminished by a ratio
proportional to the \textbf{strength parameter} $\epsilon > 0$.
}
\begin{thm} \label{thm:hyperplane-intervention}
\thmhyperplaneintervention
\end{thm}
Perturbations of the form $\Delta_g \Phi$ may be composed linearly for multiple
values of $g$. In particular, when $\epsilon$ is a universal quantity, we may
determine the total perturbation to $\Phi$ necessary to simultaneously and
proportionately decrease all qualification distances for any given state on the
stable equilibrium hyperplane. Let us denote this total perturbation as $\Delta
\Phi \define \sum_{g\in \mathcal{G}} \Delta_{g} \Phi = (\Delta \phi_{1}, \Delta
\phi_{2}, ..., \Delta \phi_{n})$. Component-wise, $\Delta \Phi$ is given by
\vspace{-0.1in}
\begin{align}
  \Delta \phi_{g}
  = \frac{-\epsilon}{s_{g}(1-s_{g})} \bigg(
  \sum_{h=g}^{n-1} \alpha_h \delta(h,h+1) + \sum_{h=1}^{g-1} \beta_h \delta(h,h+1)
  \bigg)
\end{align}
We note that the proposed feedback control mechanism depends only on the known
constants $\mu_{g}$ and feedback in terms of current qualification distances
\(\delta\). In addition, the force of the intervention can be tuned by setting
the strength parameter $\epsilon$.  Finally, we remark that this mechanism can
be composed with global perturbations of \(\phi\), \ie, universal subsidy in the
manner of \cref{thm:intervention}, to intervene without rejecting any agents
that would have been accepted under a group-independent policy.

\begin{figure}[ht] 
  \centering
  \includegraphics[width=\textwidth, trim=7 9 7 9, clip]{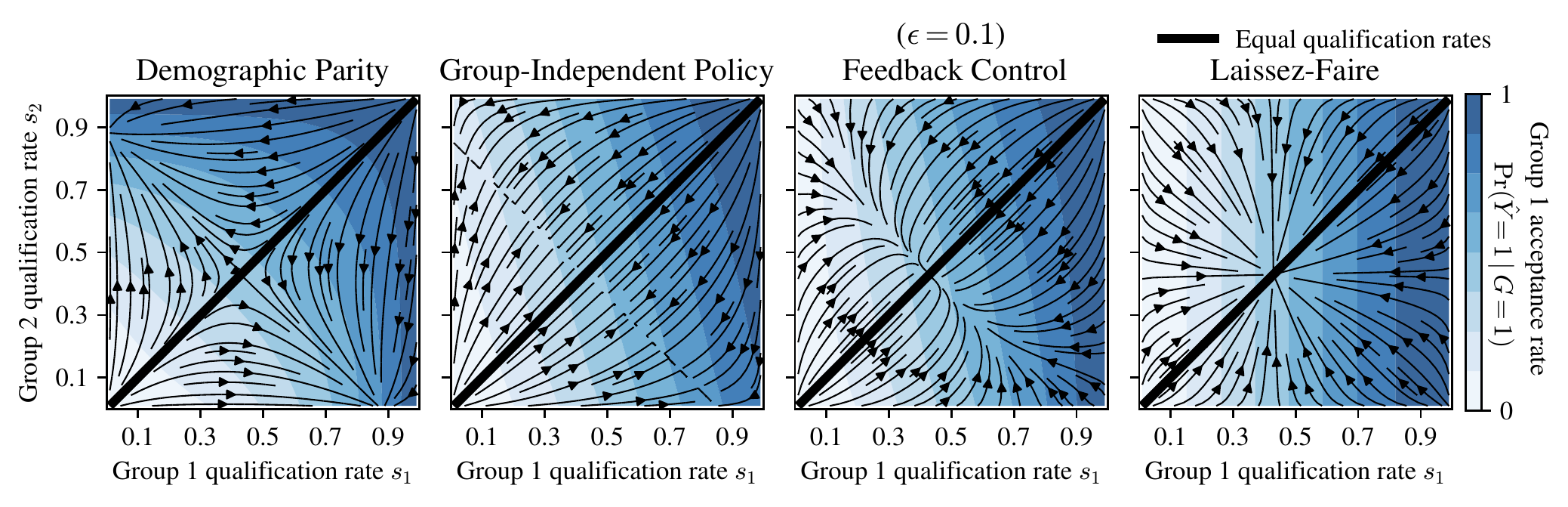}
  \caption{ Simulated dynamics for two groups of equal size, subject to
    different global interventions.  Streamlines approximate system time
    evolution.  $q_{0}$ and $q_{1}$ are Gaussians with unit variance and have
    means $-1$ and $1$, respectively.  Other examples and rendered dynamical
    variables are provided in \cref{Asec:figures}. For this example,
    \(
    (U_{0\hat{0}}= 0.1\); \(U_{0\hat{1}}=5.5\); \(U_{1\hat{0}}=0.5\); \(U_{1\hat{1}}=1.0;
    V_{0\hat{0}}=0.5\); \(V_{0\hat{1}}=-0.5\); \(V_{1\hat{0}}=-0.25\); \(V_{1\hat{1}}=1.0
    )\).
  }\label{fig:dynamics}
    \vspace{-0.2in}

\end{figure}
We compare interventions by appeal to simulation, choosing a setting that
guarantees a single, stable average qualification rate \(\avg{s}^\star\) under
group-independent policies (\textbf{GI}) (\cref{fig:dynamics}). We consider
trade-offs between \emph{normative present fairness} (\bNPF) (\eg, demographic
parity (\textbf{DP}) or equalized odds (\textbf{EO})) and \emph{long-term
fairness} (\bLTF), for which the dynamics must converge to the line demarcating
equal qualification rates.  Darker shading (blue) implies a higher absolute
acceptance rate for Group 1, which, by the setting's symmetry, is the same for
Group 2 when reflected across the aforementioned line; reflectional asymmetry
violates DP.  Under compulsory DP (first pane), the system violates \LTF,
settling into a ``patronizing equilibrium under affirmative action'', as coined
by \citet{coate1993will}, in which agents from a less-qualified group are
\emph{patronized} (\eg, granted loans despite nonqualification) by the
classifier (\cf the upper-left corner, with low qualification and high
acceptance rates for Group 1). States under GI (second pane), which satisfies
the EO notion of \NPF{ }by \cref{thm:eq-opt}, converge to a line of constant
\(\avg{s}\) (\cf \cref{rem:hyperplane}) while preserving qualification
disparities (\cf \cref{thm:yangsthm}).  \LTF{ }is expected from a
\textbf{laissez-faire} (\textbf{LZ}) policy (last pane), which adopts
group-specific policies and thus decouples all group dynamics: Each group must
converge to \(\avg{s}^\star\) separately. Still, LZ satisfies neither the DP (by
reflectional asymmetry) nor EO (by \cref{thm:eq-opt}) notions of \NPF. In
contrast, feedback control (third pane) achieves \LTF{ }by conceding
\(\epsilon\)-small, parametric violations of \NPF{ }(EO) (See
\cref{Asec:figures} for plots of classifier error rates in this setting).

\vspace{-0.2cm}
\section{Discussion and limitations} \label{sec:discussion}
\vspace{-0.1cm}
The novelty of our contribution is the demonstration of persistent qualification
rate disparities in a symmetric setting consistent with plausible mechanisms of
population response---sustained by the careless deployment of machine learning
and myopic fairness interventions.  We submit that, given the many charitable
assumptions of our model to achieve perfect structural equality between groups,
any reasonable fairness intervention should succeed in responsibly rectifying
disparities here, if anywhere.  Moreover, we have laid bare inherent tensions
that can exist between the \emph{means} and \emph{ends} of fairness
considerations in a dynamical context, demonstrating the potential
incompatibility of immediate and long-term notions of fairness.

We acknowledge that our model is simplistic, but such simple cases must be
well-understood as a first step towards further, equitable models of population
response. We regard the requirement of strictly isolated groups as the most
tenuous assumption of our model and conjecture that even relatively weak
inter-group exchange of strategies should lead to long-term fairness in our
default setting.  Nonetheless, we believe that a program based on incomplete
agent information can successfully endogenize persistent disparities in
symmetric settings more robustly. Specifically, future work may consider
multiple classifiers with different task domains affecting a common population;
we expect this extension to readily endogenize broken symmetries between group
environments and conditions. We also trust that voluntary participation, as
considered by \citet{zhang2019group}, may be modelled as an additional strategy
within our framework. Regarding empirical falsifiability, we note that the
dynamics of social disparity are not exclusive to algorithmic
classifiers \cite{hu2018short}, and ask whether our model's predictions may be
contrasted with existing and historical resource allocation problems.

We invite readers to consider both our model and  application of control theory
to society through algorithmic classification, with care.  We intend our work
to \emph{reform} the misapplication of machine learning, inappropriate modelling
assumptions, and myopic notions of fairness.

\begin{ack}
We believe our paper benefited significantly from NeurIPS 2021 review process.
We thank our peer-reviewers for their helpful feedback and constructive
criticism of the originally submitted version of this paper. We also thank the
NeurIPS Area Chairs, Senior Area Chair, Ethics Chairs, and Program Chairs for
their discussion of relevant sociotechnical issues and helpful recommendations
in preparation of this published version, in which we believe our contributions
are better framed and clarified.

This work is partially supported by the National
Science Foundation (NSF) Program on Fairness in Artificial Intelligence in
Collaboration with Amazon (FAI) under grant IIS-2040800 and by NSF grant
CCF-2023495.
\end{ack}
{
  \bibliographystyle{unsrtnat}
  \bibliography{references.bib}

\begin{thebibliography}{43}
\providecommand{\natexlab}[1]{#1}
\providecommand{\url}[1]{\texttt{#1}}
\expandafter\ifx\csname urlstyle\endcsname\relax
  \providecommand{\doi}[1]{doi: #1}\else
  \providecommand{\doi}{doi: \begingroup \urlstyle{rm}\Url}\fi

\bibitem[Obermeyer et~al.(2019)Obermeyer, Powers, Vogeli, and
  Mullainathan]{obermeyer2019dissecting}
Ziad Obermeyer, Brian Powers, Christine Vogeli, and Sendhil Mullainathan.
\newblock Dissecting racial bias in an algorithm used to manage the health of
  populations.
\newblock \emph{Science}, 366\penalty0 (6464):\penalty0 447--453, 2019.

\bibitem[Hao(2020)]{Hao.2020}
Karen Hao.
\newblock The coming war on the hidden algorithms that trap people in poverty.
\newblock \emph{MIT Technology Review}, 2020.

\bibitem[Metz and Satariano(2020)]{Metz.2020}
Cade Metz and Adam Satariano.
\newblock An algorithm that grants freedom, or takes it away.
\newblock \emph{The New York Times}, 2020.

\bibitem[Newton(2021)]{Newton.2021}
Derek Newton.
\newblock From admissions to teaching to grading, {AI} is infiltrating higher
  education.
\newblock \emph{The Hechinger Report}, 2021.

\bibitem[Hernandez(2021)]{Hernandez.2021}
Joe Hernandez.
\newblock A military drone with a mind of its own was used in combat, {U}.{N}.
  says.
\newblock \emph{National Public Radio}, 2021.

\bibitem[Crawford and Calo(2016)]{crawford2016there}
Kate Crawford and Ryan Calo.
\newblock There is a blind spot in {A}{I} research.
\newblock \emph{Nature News}, 538\penalty0 (7625):\penalty0 311, 2016.

\bibitem[Chaney et~al.(2018)Chaney, Stewart, and
  Engelhardt]{chaney2018algorithmic}
Allison~JB Chaney, Brandon~M Stewart, and Barbara~E Engelhardt.
\newblock How algorithmic confounding in recommendation systems increases
  homogeneity and decreases utility.
\newblock In \emph{Proceedings of the 12th ACM Conference on Recommender
  Systems}, pages 224--232. ACM, 2018.

\bibitem[Fuster et~al.(2018)Fuster, Goldsmith-Pinkham, Ramadorai, and
  Walther]{fuster2018predictably}
Andreas Fuster, Paul Goldsmith-Pinkham, Tarun Ramadorai, and Ansgar Walther.
\newblock Predictably unequal? {T}he effects of machine learning on credit
  markets.
\newblock \emph{The Effects of Machine Learning on Credit Markets}, 2018.

\bibitem[Corbett-Davies and Goel(2018)]{corbett2018measure}
Sam Corbett-Davies and Sharad Goel.
\newblock The measure and mismeasure of fairness: A critical review of fair
  machine learning.
\newblock \emph{arXiv preprint arXiv:1808.00023}, 2018.

\bibitem[Dwork et~al.(2012)Dwork, Hardt, Pitassi, Reingold, and
  Zemel]{dwork2012fairness}
Cynthia Dwork, Moritz Hardt, Toniann Pitassi, Omer Reingold, and Richard Zemel.
\newblock Fairness through awareness.
\newblock In \emph{Proceedings of the 3rd innovations in theoretical computer
  science conference}, pages 214--226, 2012.

\bibitem[Zemel et~al.(2013)Zemel, Wu, Swersky, Pitassi, and
  Dwork]{zemel2013learning}
Rich Zemel, Yu~Wu, Kevin Swersky, Toni Pitassi, and Cynthia Dwork.
\newblock Learning fair representations.
\newblock In \emph{International conference on machine learning}, pages
  325--333. PMLR, 2013.

\bibitem[Hardt et~al.(2016)Hardt, Price, and Srebro]{hardt2016equality}
Moritz Hardt, Eric Price, and Nathan Srebro.
\newblock Equality of opportunity in supervised learning, 2016.

\bibitem[Zafar et~al.(2017{\natexlab{a}})Zafar, Valera, Gomez~Rodriguez, and
  Gummadi]{zafar2017fairness}
Muhammad~Bilal Zafar, Isabel Valera, Manuel Gomez~Rodriguez, and Krishna~P
  Gummadi.
\newblock Fairness beyond disparate treatment \& disparate impact: Learning
  classification without disparate mistreatment.
\newblock In \emph{Proceedings of the 26th international conference on world
  wide web}, pages 1171--1180, 2017{\natexlab{a}}.

\bibitem[Chouldechova(2017)]{chouldechova2017fair}
Alexandra Chouldechova.
\newblock Fair prediction with disparate impact: A study of bias in recidivism
  prediction instruments.
\newblock \emph{Big data}, 5\penalty0 (2):\penalty0 153--163, 2017.

\bibitem[Feldman et~al.(2015)Feldman, Friedler, Moeller, Scheidegger, and
  Venkatasubramanian]{feldman2015certifying}
Michael Feldman, Sorelle~A Friedler, John Moeller, Carlos Scheidegger, and
  Suresh Venkatasubramanian.
\newblock Certifying and removing disparate impact.
\newblock In \emph{proceedings of the 21th ACM SIGKDD international conference
  on knowledge discovery and data mining}, pages 259--268, 2015.

\bibitem[Kleinberg et~al.(2016)Kleinberg, Mullainathan, and
  Raghavan]{kleinberg2016inherent}
Jon Kleinberg, Sendhil Mullainathan, and Manish Raghavan.
\newblock Inherent trade-offs in the fair determination of risk scores.
\newblock \emph{arXiv preprint arXiv:1609.05807}, 2016.

\bibitem[Zafar et~al.(2017{\natexlab{b}})Zafar, Valera, Rodriguez, Gummadi, and
  Weller]{zafar17}
Muhammad~Bilal Zafar, Isabel Valera, Manuel~Gomez Rodriguez, Krishna~P.
  Gummadi, and Adrian Weller.
\newblock From parity to preference-based notions of fairness in
  classification.
\newblock In \emph{Proceedings of the 31st International Conference on Neural
  Information Processing Systems}, NIPS'17, page 228–238. Curran Associates
  Inc., 2017{\natexlab{b}}.
\newblock ISBN 9781510860964.

\bibitem[Ustun et~al.(2019)Ustun, Liu, and Parkes]{ustun2019fairness}
Berk Ustun, Yang Liu, and David Parkes.
\newblock Fairness without harm: Decoupled classifiers with preference
  guarantees.
\newblock In \emph{International Conference on Machine Learning}, pages
  6373--6382. PMLR, 2019.

\bibitem[Kusner et~al.(2017)Kusner, Loftus, Russell, and
  Silva]{kusner2017counterfactual}
Matt~J Kusner, Joshua~R Loftus, Chris Russell, and Ricardo Silva.
\newblock Counterfactual fairness.
\newblock \emph{arXiv preprint arXiv:1703.06856}, 2017.

\bibitem[Kasy and Abebe(2021)]{kasy2021fairness}
Maximilian Kasy and Rediet Abebe.
\newblock Fairness, equality, and power in algorithmic decision-making.
\newblock In \emph{Proceedings of the 2021 ACM Conference on Fairness,
  Accountability, and Transparency}, pages 576--586, 2021.

\bibitem[Coate and Loury(1993)]{coate1993will}
Stephen Coate and Glenn~C Loury.
\newblock Will affirmative-action policies eliminate negative stereotypes?
\newblock \emph{The American Economic Review}, pages 1220--1240, 1993.

\bibitem[D'Amour et~al.(2020)D'Amour, Srinivasan, Atwood, Baljekar, Sculley,
  and Halpern]{d2020fairness}
Alexander D'Amour, Hansa Srinivasan, James Atwood, Pallavi Baljekar, D~Sculley,
  and Yoni Halpern.
\newblock Fairness is not static: Deeper understanding of long term fairness
  via simulation studies.
\newblock In \emph{Proceedings of the 2020 Conference on Fairness,
  Accountability, and Transparency}, pages 525--534, 2020.

\bibitem[Zhang et~al.(2020)Zhang, Tu, Liu, Liu, Kjellstr{\"o}m, Zhang, and
  Zhang]{zhang2020fair}
Xueru Zhang, Ruibo Tu, Yang Liu, Mingyan Liu, Hedvig Kjellstr{\"o}m, Kun Zhang,
  and Cheng Zhang.
\newblock How do fair decisions fare in long-term qualification?
\newblock \emph{arXiv preprint arXiv:2010.11300}, 2020.

\bibitem[Heidari et~al.(2019)Heidari, Nanda, and Gummadi]{heidari2019on}
Hoda Heidari, Vedant Nanda, and Krishna~P. Gummadi.
\newblock On the long-term impact of algorithmic decision policies: Effort
  unfairness and feature segregation through social learning.
\newblock \emph{the International Conference on Machine Learning (ICML)}, 2019.

\bibitem[Wen et~al.(2019)Wen, Bastani, and Topcu]{wen2019fairness}
Min Wen, Osbert Bastani, and Ufuk Topcu.
\newblock Fairness with dynamics.
\newblock \emph{arXiv preprint arXiv:1901.08568}, 2019.

\bibitem[Liu et~al.(2020)Liu, Wilson, Haghtalab, Kalai, Borgs, and
  Chayes]{liu2019disparate}
Lydia~T Liu, Ashia Wilson, Nika Haghtalab, Adam~Tauman Kalai, Christian Borgs,
  and Jennifer Chayes.
\newblock The disparate equilibria of algorithmic decision making when
  individuals invest rationally.
\newblock In \emph{Proceedings of the 2020 Conference on Fairness,
  Accountability, and Transparency}, pages 381--391, 2020.

\bibitem[Hu and Chen(2018)]{hu2018short}
Lily Hu and Yiling Chen.
\newblock A short-term intervention for long-term fairness in the labor market.
\newblock In \emph{Proceedings of the 2018 World Wide Web Conference on World
  Wide Web}, pages 1389--1398. International World Wide Web Conferences
  Steering Committee, 2018.

\bibitem[Mouzannar et~al.(2019)Mouzannar, Ohannessian, and
  Srebro]{mouzannar2019fair}
Hussein Mouzannar, Mesrob~I Ohannessian, and Nathan Srebro.
\newblock From fair decision making to social equality.
\newblock In \emph{Proceedings of the Conference on Fairness, Accountability,
  and Transparency}, pages 359--368. ACM, 2019.

\bibitem[Williams and Kolter(2019)]{williams2019dynamic}
Joshua Williams and J~Zico Kolter.
\newblock Dynamic modeling and equilibria in fair decision making.
\newblock \emph{arXiv preprint arXiv:1911.06837}, 2019.

\bibitem[Liu et~al.(2018)Liu, Dean, Rolf, Simchowitz, and
  Hardt]{liu2018delayed}
Lydia~T Liu, Sarah Dean, Esther Rolf, Max Simchowitz, and Moritz Hardt.
\newblock Delayed impact of fair machine learning.
\newblock In \emph{International Conference on Machine Learning}, pages
  3150--3158. PMLR, 2018.

\bibitem[Hu et~al.(2019)Hu, Immorlica, and Vaughan]{hu2019disparate}
Lily Hu, Nicole Immorlica, and Jennifer~Wortman Vaughan.
\newblock The disparate effects of strategic manipulation.
\newblock In \emph{Proceedings of the Conference on Fairness, Accountability,
  and Transparency}, pages 259--268, 2019.

\bibitem[Zhang et~al.(2019)Zhang, Khalili, Tekin, and Liu]{zhang2019group}
Xueru Zhang, Mohammad~Mahdi Khalili, Cem Tekin, and Mingyan Liu.
\newblock Group retention when using machine learning in sequential decision
  making: {T}he interplay between user dynamics and fairness.
\newblock In \emph{Advances in Neural Information Processing Systems}, pages
  15243--15252, 2019.

\bibitem[Tang et~al.(2020)Tang, Ho, and Liu]{tang2020fair}
Wei Tang, Chien-Ju Ho, and Yang Liu.
\newblock Bandit learning with delayed impact of actions.
\newblock \emph{arXiv preprint arXiv:2002.10316}, 2020.

\bibitem[Joseph et~al.(2016)Joseph, Kearns, Morgenstern, and
  Roth]{joseph2016fairness}
Matthew Joseph, Michael Kearns, Jamie~H Morgenstern, and Aaron Roth.
\newblock Fairness in learning: Classic and contextual bandits.
\newblock In \emph{Advances in Neural Information Processing Systems}, pages
  325--333, 2016.

\bibitem[Liu et~al.(2017)Liu, Radanovic, Dimitrakakis, Mandal, and
  Parkes]{liu2017calibrated}
Yang Liu, Goran Radanovic, Christos Dimitrakakis, Debmalya Mandal, and David~C
  Parkes.
\newblock Calibrated fairness in bandits.
\newblock \emph{arXiv preprint arXiv:1707.01875}, 2017.

\bibitem[Jabbari et~al.(2017)Jabbari, Joseph, Kearns, Morgenstern, and
  Roth]{jabbari2017fairness}
Shahin Jabbari, Matthew Joseph, Michael Kearns, Jamie Morgenstern, and Aaron
  Roth.
\newblock Fairness in reinforcement learning.
\newblock In \emph{Proceedings of the 34th International Conference on Machine
  Learning-Volume 70}, pages 1617--1626. JMLR. org, 2017.

\bibitem[Ensign et~al.(2018)Ensign, Friedler, Neville, Scheidegger, and
  Venkatasubramanian]{ensign2018runaway}
Danielle Ensign, Sorelle~A Friedler, Scott Neville, Carlos Scheidegger, and
  Suresh Venkatasubramanian.
\newblock Runaway feedback loops in predictive policing.
\newblock In \emph{Conference of Fairness, Accountability, and Transparency},
  2018.

\bibitem[Bj{\"o}rnerstedt and Weibull(1994)]{bjornerstedt1994nash}
Jonas Bj{\"o}rnerstedt and J{\"o}rgen~W Weibull.
\newblock Nash equilibrium and evolution by imitation.
\newblock Technical report, IUI Working Paper, 1994.

\bibitem[Taylor and Jonker(1978)]{taylor1978evolutionary}
Peter~D Taylor and Leo~B Jonker.
\newblock Evolutionary stable strategies and game dynamics.
\newblock \emph{Mathematical biosciences}, 40\penalty0 (1-2):\penalty0
  145--156, 1978.

\bibitem[Dawkins(1976)]{dawkins1976selfish}
Richard Dawkins.
\newblock \emph{The Selfish Gene}.
\newblock Oxford University Press, 1976.

\bibitem[Friedman and Sinervo(2016)]{friedman2016evolutionary}
Daniel Friedman and Barry Sinervo.
\newblock \emph{Evolutionary Games in Natural, Social, and Virtual Worlds}.
\newblock Oxford University Press, 2016.

\bibitem[Strogatz(2018)]{strogatz2018nonlinear}
Steven~H Strogatz.
\newblock \emph{Nonlinear Dynamics and Chaos with Student Solutions Manual:
  With Applications to Physics, Biology, Chemistry, and Engineering}.
\newblock CRC press, 2018.

\bibitem[Boyd et~al.(2004)Boyd, Boyd, and Vandenberghe]{boyd2004convex}
Stephen Boyd, Stephen~P Boyd, and Lieven Vandenberghe.
\newblock \emph{Convex optimization}.
\newblock Cambridge university press, 2004.

\end{thebibliography}
}

\newpage
\appendix

The Appendices are organized according to \cref{table:appendices}.

\begin{table}[h]
    \caption{Organization of
    appendices} \label{table:appendices} \centering \begin{tabular}{cl} \toprule

    Appendix & Content  \\ \midrule

\cref{Asec:notation} & Summary of notation used in the main paper \\

\cref{Asec:proofs} & Proofs of all theorems, remarks, and corollaries as well as additional lemmas \\

\cref{Asec:figures} & Figures like \cref{fig:dynamics}, exploring different settings and variables of interest \\

    \bottomrule \end{tabular} \end{table}

\newpage

\section{Notation}\label{Asec:notation}

\begin{table}[h]
    \caption{Choice of
    notation} \label{table:notation} \centering \begin{tabular}{cl} \toprule

       \multicolumn{2}{l}{\underline{Parameters}: } \\
      
       $n$ & Number of groups \\ $\mathcal{G}$ & Set of groups $\{1, 2, ...,
       n\}$ \\ $\mu_g$ & Fraction of total population in group $g$  \\
       $\boldsymbol{\mu}$ & Vector $(\mu_1, \mu_2, ... , \mu_n)$ \\ $V$ &
       $2 \times 2$ matrix of classifier utilities, indexed by $(Y, \hat{Y})$
       pairs \\ $\theta$ & Classifier's probability threshold
       (\cref{thm:theta-threshold}) \\ $U$ & $2 \times 2$ matrix of agent
       fitnesses, indexed by $(Y, \hat{Y})$ pairs \\
    
       $q_y$ & Probability density function of $X$ given $Y=y$ \\ $Q_y$ &
       Cumulative distribution function of $X$ given $Y=y$ \\
       
       \cmidrule(r){1-2}     \multicolumn{2}{l}{\underline{Random Variables}:
       } \\
     
       $G$ & Group to which an agent belongs \\ $X$ & Real-valued feature of an
       agent \\ $Y$ & Actual binary label (qualification) of an agent \\
       $\hat{Y}$ & Predicted binary label (qualification) of an agent \\
       
       \cmidrule(r){1-2}  \multicolumn{2}{l}{\underline{Indices}: } \\
     
       $g, h, i, j$ & used to indicate a group \\ $x$ & used to indicate a
       feature value \\ $y$ & used to indicate a binary label (qualification) \\
       $\hat{y}$ & used to indicate a predicted binary label (qualification) \\
      
       \cmidrule(r){1-2}      \multicolumn{2}{l}{\underline{Dynamical
       Variables}: } \\

       $t$ & Discrete time \\ $\cdot[t]$ & Restriction of a dynamical variable
       to time $t$ \\ $s_g$ & Fraction of qualified agents in group $g$ \\
       $\mathbf{s}$ & Vector $(s_1, s_2, ..., s_n)$ \\ $\avg{s}$ & Fraction of
       qualified agents in total population \\ $\delta(g,h)$ & Difference in
       qualification rates between groups: $s_g - s_h$ \\ $D$ & A set of $n-1$
       linearly independent qualification distances \(\delta\) \\ $\mathbf{r}$ &
       Vector with the elements of $D$ prepending $\avg{s}$ as components \\
      
       $\pi$ & Classifier's policy mapping $X$ to $\hat{Y}$ \\ $\phi$ &
       Classifier's feature threshold (\cref{thm:phi-threshold}) \\
      
       $W_y$ & Average agent fitness conditioned on $Y = y$ \\ $\avg{W}_g$ &
       Average agent fitness conditioned on $G = g$ \\

       \cmidrule(r){1-2}       \multicolumn{2}{l}{\underline{Miscellaneous}:
       } \\
      
       $J$ & Jacobian matrix for dynamical system. \\ $\lambda$ & A specific
       eigenvalue of $J$ \\ $\mathbf{v}$ & A specific eigenvector of $J$ \\
       $\Phi$ & Vector of group-specific feature thresholds \(\phi_g\) \\

      \bottomrule \end{tabular} \end{table}

\newpage
\section{Proofs}\label{Asec:proofs}

\begin{center}
\textbf{Proof of \cref{thm:phi-threshold}}
\end{center}
\setcounter{lem}{0} \setcounterref{thm}{thm:phi-threshold}

\newcommand{\thmthetathreshold}{ Discounting sets of measure zero, the
  $u$-maximizing policy $\pi$ is parameterized by a single \textbf{probability
    threshold} $\theta \in [0, 1]$ such that

\begin{align} \label{eqA:probability-threshold}
  \pi(x) = \begin{cases} 1 & \Pr(Y = 1 \mid X = x) > \theta \\ 0 &
    \text{otherwise}
  \end{cases}
\end{align}
where
\begin{align}
  \theta = \frac{V_{0\hat{0}} - V_{0\hat{1}}}{V_{1\hat{1}} - V_{0\hat{1}} +
    V_{0\hat{0}} - V_{0\hat{1}}}
\end{align}
}
\begin{lem}\label{thm:theta-threshold}
\thmthetathreshold
\end{lem}

\begin{proof}[Proof of \cref{thm:theta-threshold}]
  Discounting sets of measure zero (\ie, rejecting the possibility of strict
  equality as infinitely unlikely), a classifier will accept an agent with
  feature value $X = x$ if and only if the expected utility of doing so is
  greater than rejecting.
\begin{equation}
  \Bigg( \sum_{y=0}^{1} \Pr(Y = y \mid X = x) V_{y\hat{1}} \Bigg) > \Bigg(
  \sum_{y=0}^{1} \Pr(Y = y \mid X = x) V_{y\hat{0}} \Bigg)
\end{equation}
This reduces algebraically to
\begin{align} \label{eqA:xi-criterion}
  \frac{\Pr(Y = 1 \mid X = x)}{\Pr(Y = 0 \mid X = x)} > \xi
\end{align}
where
\begin{align}\label{eqA:def-xi}
  \xi \define \dfrac{V_{0\hat{0}} - V_{0\hat{1}}}{V_{1\hat{1}} - V_{1\hat{0}}}
\end{align}
and by change of variables
\begin{align} \theta \define \frac{\xi}{1 + \xi} =
           \frac{V_{0\hat{0}} - V_{0\hat{1}}}{V_{1\hat{1}} - V_{1\hat{0}} +
             V_{0\hat{0}} - V_{0\hat{1}}}, \quad \xi = \frac{\theta}{1 -
             \theta} \label{eqA:xi-theta}
\end{align}
to
\begin{align} \label{eqA:criterion-theta}
  \Pr(Y = 1 \mid X = x) > \theta
\end{align}
\cref{eqA:criterion-theta} is thus the sole criterion for accepting an agent
with feature value $X = x$, and our proof is complete.
\end{proof}

\begin{lem}\label{lem:pryx}
  \begin{align}
    \Pr(Y = 1 \mid X = x) = \frac{\avg{s} q_{1}(x)}{\avg{s} q_{1}(x) + (1 -
      \avg{s}) q_{0}(x)}
  \end{align}
\end{lem}
\begin{proof}[Proof of \cref{lem:pryx}]
     By Bayes's Theorem,
  \begin{subequations}
     \begin{align}
     \Pr(G = g, Y = 1 \mid X = x) &= \frac{\pr_{X}(x \mid G = g, Y = 1) \Pr(Y =
       1 \mid G = g) \Pr(G = g)}{\pr_{X}(x)} \\ &= \frac{ q_{1}(x)s_{g}\mu_{g}
     }{ \sum_{h \in \mathcal{G}} \big(s_{h}q_{1}(x) + (1 - s_{h})q_{0}(x)
       \big)\mu_{h} }
     \end{align}
   \end{subequations}
   By marginalizing over groups $g$, it follows that
   \begin{align}
     \Pr(Y = 1 \mid X = x) = \frac{ \sum_{g\in \mathcal{G}} s_{g}q_{1}(x)
       \mu_{g} }{ \sum_{h\in \mathcal{G}} \big(s_{h}q_{1}(x) + (1 -
       s_{h})q_{0}(x) \big)\mu_{h} }
   \end{align}
   This expression may be simplified to the target statement by substituting
   from \cref{eq:def-mu}:
   \begin{align}
     \avg{s} \define \sum_{g \in \mathcal{G}} \mu_{g} s_{g}
   \end{align}
  \begin{subequations}
   \begin{align}
     (1 - \avg{s}) &= \sum_{g} \mu_{g} - \sum_{g \in \mathcal{G}} \mu_{g} s_{g}
     \\ &= \sum_{g \in \mathcal{G}} \mu_{g} (1 - s_{g})
   \end{align}
 \end{subequations}
 \end{proof}

\begin{lem}\label{lem:pryxsupport}
  $\Pr(Y=1 \mid X=x)$ and $\Pr(Y=0 \mid X=x)$ have support for all $x$.
  \begin{align}
    \forall x \in (-\infty, \infty),~ 0 < \Pr(Y = 1 \mid X = x) < 1
  \end{align}
\end{lem}

\begin{proof}[Proof of \cref{lem:pryxsupport}]
  By \cref{asm:q1-q0-monotonic} and \cref{asm:s-interior}, ${\avg{s}q_{1}(x)}$
  and ${(1 - \avg{s})q_{0}(x)}$ must both be strictly positive.  As an immediate
  consequence of \cref{lem:pryx} and the identity ${\Pr(Y=1 \mid X=x) = 1 -
    \Pr(Y=0 \mid X=x)}$, we conclude that both ${\Pr(Y=1 \mid X=x)}$ and
  ${\Pr(Y=0 \mid X=x)}$ are greater than zero.
 \end{proof}

\begin{lem}\label{lem:pryxmonotonic}
  \(\Pr(Y = 1 \mid X = x)\) is monotonically increasing in \(X\).
\end{lem}

\begin{proof}[Proof of \cref{lem:pryxmonotonic}]
   By \cref{asm:q1-q0-monotonic}, we have that
   \[
     \forall y, x, \quad q_{y}(x) \in (0, \infty) \text{ and } \dd{}{x} \Bigg(
     \frac{q_{1}(x)}{q_{0}(x)} \Bigg) > 0
   \]
   While from \cref{lem:pryx},
   \[
     \Pr(Y = 1 \mid X = x) = \frac{\avg{s} q_{1}(x)}{\avg{s} q_{1}(x) + (1 -
       \avg{s}) q_{0}(x)}
   \]

   By the differentiability and strict positivity of each $q$
   (\cref{asm:q1-q0-monotonic}) as well as the strict psitivity of \(\Pr(Y = 1
   \mid X = x)\) by \cref{lem:pryxsupport}, it is sufficient to show strict
   positivity of the first derivative of $\Pr(Y=1 \mid X=x)$ to prove
   monotonicity. We therefore wish to show
  \begin{align}
    \dd{}{x} \Bigg( \frac{q_{1}(x)}{q_{0}(x)} \Bigg) > 0 \implies \dd{}{x}
    \Bigg( \frac{ \avg{s} q_{1}(x) }{ \avg{s} q_{1}(x) + (1 - \avg{s}) q_{0}(x)
    } \Bigg) > 0
  \end{align}

  First, let us define
  \begin{align}
    a(x) \define \avg{s} q_{1}(x) ; \quad b(x) \define (1 - \avg{s}) q_{0}(x)
  \end{align}
  Our objective may therefore be rewritten as
  \begin{align}
    \Bigg( \frac{ (1 - \avg{s}) }{ \avg{s} } \Bigg) \dd{}{x} \Bigg(
    \frac{a(x)}{b(x)} \Bigg) > 0 \implies \dd{}{x} \Bigg( \frac{a(x)}{a(x) +
      b(x)} \Bigg) > 0
  \end{align}
  Performing explicit differentiation,
  \begin{align}
    \dd{}{x} \Bigg( \frac{a(x)}{b(x)} \Bigg) &= \Bigg(\frac{a(x) +
      b(x)}{b(x)}\Bigg)^{2} \dd{}{x} \Bigg( \frac{a(x)}{a(x) + b(x)} \Bigg)
  \end{align}
  we again rewrite our objective as
  \begin{align}
         \frac{ (1 - \avg{s}) }{ \avg{s} } \Bigg(\frac{a(x) +
           b(x)}{b(x)}\Bigg)^{2} \dd{}{x} \Bigg( \frac{a(x)}{a(x) + b(x)} \Bigg)
         > 0 &\implies \dd{}{x} \Bigg( \frac{a(x)}{a(x) + b(x)} \Bigg) > 0
  \end{align}
  since $\frac{1 - \avg{s}}{\avg{s}} > 0$ by \cref{asm:s-interior}, this is a
  necessarily true statement.
\end{proof}

\textbf{\cref{thm:phi-threshold} Statement}.  Discounting sets of measure zero,
the \(u\)-maximizing, group-independent policy \(\pi\) is parameterized by the
feature threshold \(\phi \in [-\infty, \infty]\) such that \(\pi(x) = 1\) if and
only if \(x > \phi\), where \(\phi\) depends only on the global qualification
rate \(\bar{s}\) as
\begin{align} \label{eqA:q1-q0-s-theta}
  \xi \define \dfrac{V_{0\hat{0}} - V_{0\hat{1}}}{V_{1\hat{1}} - V_{1\hat{0}}};
  \quad \frac{q_{1}(\phi)}{q_{0}(\phi)} = \xi \bigg( \frac{1 - \avg{s}}{\avg{s}}
  \bigg)
 \end{align}
When a solution in \(\phi\) to the threshold equation, \cref{eqA:q1-q0-s-theta},
does not exist, \(\phi\) is either \(\pm \infty\).

\begin{proof}[Proof of \cref{thm:phi-threshold}]
  By \cref{thm:theta-threshold}, a threshold value for $\Pr(Y = 1 \mid X = x)$
  (\ie, $\theta$) is sufficient to characterize the optimal classifier policy
  $\pi$:
  \begin{align}
  \pi(x) = \begin{cases} 1 & \Pr(Y = 1 \mid X = x) > \theta \\ 0 &
    \text{otherwise}
  \end{cases}
 \end{align}
 \cref{lem:pryxmonotonic} concludes that $\Pr(Y = 1 \mid X = x)$ is monotonic in
 $X$, and so it follows that whenever $\Pr(Y=1\mid X=x)$ achieves the threshold
 value of $\theta$, it does so at a corresponding, unique feature-threshold
 value $X = \phi$
  \begin{align} \label{eqA:phi-theta}
    \forall x,~\Pr(Y = 1 \mid X = x) = \theta \implies x = \phi
  \end{align}
  such that the classifier's policy is given by
  \begin{align} \label{eqA:pi-phi}
    \pi(x) = \begin{cases} 1 & x > \phi \\ 0 & \text{otherwise}
    \end{cases}
  \end{align}
  When such equality between $\Pr(Y=1\mid X=x)$ and $\theta$ never occurs, we
  are free to define
  \begin{align}
    \phi =
    \begin{cases}
      -\infty & \min_{x} \big\{\Pr(Y = 1 \mid X = x)\big\} > \theta  \\ \infty &
      \max_{x} \big\{\Pr(Y = 1 \mid X = x)\big\} < \theta \\
    \end{cases}
  \end{align}
  so that \cref{eqA:pi-phi} remains valid in all cases.

  Finally, when equality between $\Pr(Y=1\mid X=x)$ and $\theta$ does occur, we
  may solve for finite $\phi$ by re-expressing $\Pr(Y=1 \mid X=x)$ according to
  \cref{lem:pryx}:
  \begin{align}
    \label{eqA:theta-from-s-deriv}
    \theta &= \frac{\avg{s} q_{1}(\phi)}{\avg{s} q_{1}(\phi) + (1 - \avg{s})
      q_{0}(\phi)}
  \end{align}
  Algebraic manipulations are sufficient to derive \cref{eqA:q1-q0-s-theta},
  where we appeal to \cref{asm:s-interior} (\(s_g \in (0, 1)\)) and
  \cref{asm:xi-domain} (\(\xi \in (0, \infty)\), thus \(\theta \in (0, 1)\) by
  \cref{eqA:xi-theta}) to ensure that we do not divide by 0.
\end{proof}

\begin{center}
\textbf{Proof of \cref{cor:ddphis}}
\end{center}

\textbf{\cref{cor:ddphis} Statement}.  The classifier's feature threshold $\phi$
responds inversely to $\avg{s}$:
\begin{align}
  \dd{\phi}{\avg{s}} < 0, \quad \dd{\avg{s}}{\phi} < 0
\end{align}

\begin{proof}[Proof of \cref{cor:ddphis}]
  Let us differentiate \cref{eqA:q1-q0-s-theta} with respect to $\avg{s}$. By
  the chain rule,
  \begin{subequations}
  \begin{align}
      \dd{}{\phi} \bigg( \frac{q_{1}(\phi)}{q_{0}(\phi)} \bigg)
      \bigg(\dd{\phi}{\avg{s}} \bigg) &= \bigg( \frac{\theta}{1 - \theta} \bigg)
      \dd{}{\avg{s}}\bigg( \frac{1 - \avg{s}}{\avg{s}} \bigg) \\ &= \bigg(
      \frac{\theta}{1 - \theta} \bigg)
      \bigg(\frac{-1}{(\avg{s})^{2}}\bigg) \label{eqA:factors}
  \end{align}
\end{subequations}
By \cref{asm:q1-q0-monotonic}, \cref{asm:xi-domain}, we observe
  \begin{align}
    \dd{}{\phi} \bigg( \frac{q_{1}(\phi)}{q_{0}(\phi)} \bigg) > 0; \quad \bigg(
    \frac{\theta}{1 - \theta} \bigg) > 0; \quad
    \bigg(\frac{-1}{(\avg{s})^{2}}\bigg) < 0
  \end{align}
  Therefore, by accounting for the sign of each factor in \cref{eqA:factors} and
  the relationship between derivatives of inverse functions, we conclude that
  \begin{align}
    \dd{\phi}{\avg{s}} < 0, \quad \dd{\avg{s}}{\phi} < 0
  \end{align}
\end{proof}

\begin{center}
\textbf{Proof of \cref{thm:WQ}}
\end{center}
\setcounter{lem}{0} \setcounterref{thm}{thm:WQ}

\begin{subdefi} \label{defi:Q}
  Define the \textbf{cumulative distribution functions} \(Q_y\) such that
  \begin{align} \label{eqA:def-Q}
  Q_{y}(\phi) \define \int_{-\infty}^{\phi} q_{y}(x) \d{x}, ~~~ y \in \{0, 1\}
\end{align}
\end{subdefi}

\begin{lem} \label{lem:Q-independent}
  \(Q_y(\phi)\) is group-independent.
  \begin{equation}
    Q_y(\phi) = \Pr(\hat{Y} = 0 \mid Y = y)
  \end{equation}
\end{lem}

\begin{proof}[Proof of \cref{lem:Q-independent}]
  \begin{subequations}
  \begin{align}
    Q_{i}(\phi) &\define \int_{-\infty}^{\phi} q_{i}(x) \d{x} \\ &=\int_{\{x
      \colon \pi(x) = 0\}} \pr_{X}(x \mid Y =
    i) \label{eqA:defiQpb}\\ &=\int_{\{x \colon \pi(x) = 0\}} \pr_{X}(x \mid Y =
    i, G = g), \quad \forall g \in \mathcal{G}\label{eqA:defiQpc} \\ &=
    \Pr(\hat{Y} = 0 \mid Y = i, G = g), \quad \forall g \in \mathcal{G} \\ &=
    \Pr(\hat{Y} = 0 \mid Y = y)
  \end{align}
\end{subequations}
where \cref{eqA:defiQpb} is a consequence of \cref{thm:phi-threshold}, and
\cref{eqA:defiQpc} follows from \cref{asm:q-group-independent}.
\end{proof}

\textbf{\cref{thm:WQ} Statement.}  \thmWQ
\begin{proof}[Proof of \cref{thm:WQ}]
By \cref{asm:W-group-independent} and \cref{lem:Q-independent},
  \begin{align}
    W_{y}^{g} = U_{y\hat{1}} + (U_{y\hat{0}} - U_{y\hat{1}}) Q_{y}(\phi), \quad
    y \in \{0, 1\}
  \end{align}
  This expression is also group-independent, and we may denote
  \begin{align}
    \forall y, g,~W_y^g = W_y
  \end{align}
\end{proof}

\begin{center}
\textbf{Verification of \cref{eq:changecoords}}
\end{center}
\textbf{\cref{eq:changecoords} Statement.}
\begin{align*}
  s_g = \avg{s} + \sum_{h=g}^{n-1} \delta(h, h+1) - \sum_{h=1}^{n-1}
  \sum_{k=1}^{h} \mu_{k} \delta(h, h + 1)
\end{align*}

\begin{proof}[Verification of \cref{eq:changecoords}]

We will verify from \cref{eq:changecoords} directly that
\begin{align}
  \sum_g s_g \mu_g = \avg{s}; \quad s_{g} - s_{g+1} = \delta(g, g + 1)
\end{align}

First, let us verify that \(\sum_g s_g \mu_g = \avg{s}\), recalling
\(\sum_{g=1}^n \mu_g = 1\).  We recognize that thes first and third unexpanded
terms in \cref{eq:changecoords} are unvarying with \(g\), while the second term,
when summed, negates the third. That is, despite prescribing a different order
of summation, precisely the same values are summed in

\begin{align}
  \sum_{g=1}^{n} \mu_g \sum_{h=g}^{n-1} \delta(h, h+1) = \sum_{h=1}^{n-1}
  \sum_{k=1}^{h} \mu_{k} \delta(h, h + 1) = \sum_{1 \leq i \leq j < n} \mu_i
  \delta(j, j+1)
\end{align}
and so, as desired,
\begin{align}
  \sum_{g=1}^n \mu_g s_g = \avg{s}
\end{align}

Next, we rewrite the definitions of \(\alpha_g\) and \(\beta_g\) appearing in
\cref{thm:hyperplane-intervention}
\begin{align}
  \alpha_{g} \define \sum_{h = g+1}^n \mu_h &= (\mu_{g+1} + \mu_{g+2} + ... +
  \mu_{n}) \\ \beta_{g} \define -\sum_{h = 1}^g \mu_h &= -(\mu_{1} + \mu_{2} +
  ... + \mu_{g})
\end{align}

We note that \(\alpha_{g} - \beta_{g} = 1\) and may rewrite the change of
coordinates given in \cref{eq:changecoords} as
\begin{subequations}
\begin{align}
    s_{g} &= \avg{s} + \sum_{h=g}^{n-1} \delta(h, h+1) - \sum_{h=1}^{n-1}
    \sum_{k=1}^{h} \mu_{k} \delta(h, h + 1) \\ &= \avg{s} + \sum_{h=g}^{n-1}
    \delta(h, h+1) + \sum_{h=1}^{n-1} \beta_h \delta(h, h + 1) \\ &= \avg{s} +
    \sum_{h=g}^{n-1} \alpha_h \delta(h, h + 1) + \sum_{h=1}^{g-1}\beta_h
    \delta(h, h+ 1)
\end{align}
\end{subequations}

from which we may verify that
\begin{align}
  s_g - s_{g + 1} &= \alpha_g \delta(g, g + 1) - \beta_g \delta(g, g + 1) =
  \delta(g, g+1)
\end{align}
It follows that \cref{eq:changecoords} inverts the linear change of coordinates
prescribed by the definitions of \(\avg{s}\) and \(\delta(g, g+1)\).
\end{proof}

\begin{center}
\textbf{Proof of \cref{rem:hyperplane}}
\end{center}
\setcounterref{thm}{rem:hyperplane} \textbf{\cref{rem:hyperplane} Statement.}
\remhyperplane

\begin{proof}[Proof of \cref{rem:hyperplane}]
This follows from the definition of a hyperplane and \cref{eq:def-sg}
\begin{align}
     \avg{s} \define \sum_{g \in \mathcal{G}} \mu_{g} s_{g} = \langle
     \boldsymbol{\mu}, \mathbf{s} \rangle
\end{align}
\end{proof}

\begin{center}
\textbf{Proof of \cref{rem:persistence}}
\end{center}
\setcounterref{thm}{rem:persistence} \textbf{\cref{rem:persistence} Statement.}
\rempersistence

\begin{proof}[Proof of \cref{rem:persistence}]
  We first prove the forward direction (for any $p$), $\mathcal{H} = \{1, 2,
  ..., n - 1\}$

  \begin{subequations}
  \begin{align}
    \sum_{g=1}^{n-1} \| \delta(g, g+1)[t] \|_{p} = 0 &\implies \delta(h, h +
    1)[t] = 0 &\forall h \in \mathcal{H} \\ &\implies s_{h}[t] = s_{h+1}[t],
    \quad \avg{W}_{h}[t] = \avg{W}_{h + 1}[t] &\forall h \in \mathcal{H}
    \\ &\implies s_{h}[t] \frac{W_{1}}{\avg{W}_{h}[t]} = s_{h+1}[t]
    \frac{W_{1}}{\avg{W}_{h+1}[t]} &\forall h \in \mathcal{H} \\ &\implies
    s_{h}[t + 1] = s_{h+1}[t + 1] &\forall h \in \mathcal{H} \\ &\implies
    \delta(h, h + 1)[t + 1] = 0 &\forall h \in \mathcal{H} \\ &\implies
    \sum_{g=1}^{n-1} \| \delta(g, g+1)[t + 1] \|_{p} = 0
  \end{align}
  \end{subequations}

  Next, we prove the reverse direction (for any $p$):
  \begin{subequations}
  \begin{align}
    \sum_{g=1}^{n-1} \| \delta(g, g+1)[t + 1] \|_{p} = 0 &\implies \delta(h, h +
    1)[t + 1] = 0 &\forall h \in \mathcal{H} \\ &\implies s_{h}[t + 1] =
    s_{h+1}[t + 1] &\forall h \in \mathcal{H} \\ &\implies s_{h}[t]
    \frac{W_{1}}{\avg{W}_{h}[t]} = s_{h+1}[t] \frac{W_{1}}{\avg{W}_{h+1}[t]}
    &\forall h \in \mathcal{H} \\
    \label{eqA:reverse-norm}
    &\implies s_{h}[t] = s_{h+1}[t] &\forall h \in \mathcal{H} \\ &\implies
    \delta(h, h + 1)[t] = 0 &\forall h \in \mathcal{H} \\ &\implies
    \sum_{g=1}^{n-1} \| \delta(g, g+1)[t] \|_{p} = 0
  \end{align}
\end{subequations}
where \cref{eqA:reverse-norm} follows from
\begin{subequations}
  \begin{align}
    &\frac{s_{g}}{s_{g}W_{1} + (1-s_{g})W_{0}} = \frac{s_{h}}{s_{h}W_{1} +
      (1-s_{h})W_{0}} \\ \implies& s_{g}(s_{h}W_{1} + (1 - s_{h})W_{0}) =
    s_{h}(s_{g}W_{1} + (1 - s_{g})W_{0}) \\ \implies& W_{0}s_{g} = W_{0}s_{h}
    \\ \implies& s_{g} = s_{h}
  \end{align}
\end{subequations}
\end{proof}

\begin{center}
\textbf{Proof of \cref{thm:ppstplus}}
\end{center}
\setcounterref{thm}{thm:ppstplus} \textbf{\cref{thm:ppstplus} Statement.}

\thmppstplus

\begin{proof}[Proof of \cref{thm:ppstplus}]
  There are three mutually exclusive cases we must consider by appeal to the
  replicator equation (\cref{eq:replicator}) and \cref{asm:s-interior} ($s_{g}
  \in (0, 1)$ for all $g$ in $\mathcal{G}$).  Specifically, we verify that

  \begin{subequations}
  \begin{align}
    W_{1} > W_{0} \implies \forall g \in \mathcal{G},\quad
    \frac{W_{1}}{\avg{W}_{g}} > 1 &\implies \avg{s}[t + 1] > \avg{s}[t] \\ W_{1}
    = W_{0} \implies \forall g \in \mathcal{G},\quad \frac{W_{1}}{\avg{W}_{g}} =
    1 &\implies \avg{s}[t + 1] = \avg{s}[t]\\ W_{1} < W_{0} \implies \forall g
    \in \mathcal{G},\quad \frac{W_{1}}{\avg{W}_{g}} < 1 &\implies \avg{s}[t + 1]
    < \avg{s}[t]
  \end{align}
\end{subequations}

The sign of $W_{1} - W_{0}$ therefore determines the sign of $\avg{s}[t + 1] -
\avg{s}[t]$ directly. Likewise, the sign of $\avg{s}[t + 1] - \avg{s}[t]$
implies the sign of $W_{1} - W_{0}$, for any discrepancy would imply a
contradiction:
  \begin{subequations}
  \begin{align}
\avg{s}[t + 1] > \avg{s}[t] &\implies W_{1} > W_{0}  \\ \avg{s}[t + 1] =
\avg{s}[t]  &\implies W_{1} = W_{0} \\ \avg{s}[t + 1] < \avg{s}[t]  &\implies
W_{1} < W_{0}
  \end{align}
\end{subequations}
\end{proof}

\begin{center}
\textbf{Proof of \cref{thm:eqW1W0}}
\end{center}
\setcounterref{thm}{thm:eqW1W0} \textbf{\cref{thm:eqW1W0} Statement.}

\thmeqWW

  \begin{proof}[Proof of \cref{thm:eqW1W0}]
    By the definition of equilibrium (\cref{defi:equilibrium}), the forward
    direction implies equality between $s_{g}[t]$ and $s_{g}[t_{0} + 1]$ for all
    $g \in \mathcal{G}$. It follows from the replicator equation
    (\cref{eq:replicator}) that at least one condition must be met to guarantee
    this equality, and at least one is consequent:
  \begin{subequations}
  \begin{align}
    &\forall g \in \mathcal{G}, \\ \quad s_{g}[t_{0} + 1] = s_{g}[t_{0}] &\iff
    \bigg( \frac{W_{1}}{W_{1} s_{g} + W_{0}(1 - s_{g})}[t_{0}] = 1 ~~\vee~~
    s_{g}[t_{0}] = 0 \bigg)
  \end{align}
  \end{subequations}
  We consider the first case further and note that it is true if and only if
  $W_{1} = W_{0}$ or $s_{g} = 1$.
  \begin{subequations}
  \begin{align}
    \frac{W_{1}}{W_{1} s_{g} + W_{0}(1 - s_{g})} = 1 &\iff W_{1} = W_{1} s_{g} +
    W_{0}(1 - s_{g}) \\ &\iff (1 - s_{g}) W_{1} = W_{0}(1 - s_{g}) \\ &\iff
    \bigg( W_{1} = W_{0} ~~\vee~~s_{g} = 1 \bigg)
  \end{align}
  \end{subequations}
  Thus the tree of cases terminates with, for each $g \in \mathcal{G}$, any one
  of $s_{g} = 0$, $s_{g} = 1$, or $W_{0} = W_{1}$, as necessary and sufficient
  for the state to be at equilibrium. We note that if $W_{0} = W_{1}$, no other
  conditions must be considered separately for different values of $g$. We may
  therefore re-express these conditions for equilibrium succinctly as the two
  cases we set out to show:
  \begin{align}
  \text{at equilibrium} \iff \bigg( W_{1} = W_{0} ~~\vee~~ \forall g \in
  \mathcal{G},~ s_{g} \in \{0, 1\} \bigg)
  \end{align}
  \emph{Note: It is possible that $W_{1} = W_{0}$ at some trivial equilibrium,
  and so our use of the term \emph{internal} equilibrium does not strictly limit
  us to consideration of states in the \emph{interior} of the state space (\ie,
  points removed from the boundary).}
  \end{proof}

\begin{center}
\textbf{Proof of \cref{thm:WW-quasi}}
\end{center}
\setcounterref{thm}{thm:WW-quasi} \textbf{\cref{thm:WW-quasi} Statement.}
\WWquasi

\begin{proof}[Proof of \cref{thm:WW-quasi}]
  We proceed by characterizing the function $W_{1}(\phi) - W_{0}(\phi)$,
  starting with its zeros. By \cref{thm:WQ} and \cref{lem:Q-independent}, the
  values of $\phi$ for which $W_{1}(\phi) - W_{0}(\phi) = 0$ must satisfy
    \begin{align}
    \Big( U_{1\hat{1}} + (U_{1\hat{0}} - U_{1\hat{1}}) Q_{1}(\phi) \Big) - \Big(
    U_{0\hat{1}} + (U_{0\hat{0}} - U_{0\hat{1}}) Q_{0}(\phi) \Big) = 0
  \end{align}
  Next, we consider the first derivative of $W_{1} - W_{0}$ with respect to
  $\phi$:
  \begin{subequations}
  \begin{align} \label{eqA:accidentallypermuted}
    \dd{}{\phi} \Big( W_{1}(\phi) - W_{0}(\phi) \Big) &=
    q_{1}(\phi)(U_{1\hat{0}} - U_{1\hat{1}}) - q_{0}(\phi)(U_{0\hat{0}} -
    U_{0\hat{1}}) \\ &= \bigg( \frac{q_{1}(\phi)}{q_{0}(\phi)} -
    \frac{U_{0\hat{0}} - U_{0\hat{1}}}{U_{1\hat{0}} - U_{1\hat{1}}} \bigg)
    \bigg( q_{0}(\phi) (U_{1\hat{0}} - U_{1\hat{1}}) \bigg)
  \end{align}
\end{subequations}

  Recall that $U_{i\hat{0}} \neq U_{i\hat{1}}$ and \(U_{1\hat{0}} <
  U_{1\hat{1}}\) (\cref{asm:UUneq}). By the strict (increasing) monotonicity of
  $\frac{q_{1}(\phi)}{q_{0}(\phi)}$ in $\phi$ and strict positivity of
  $q_{0}(\phi)$, both guaranteed by \cref{asm:q1-q0-monotonic}, the sign of this
  expression can change at most once as $\phi$ is varied from $-\infty$ to
  $\infty$. We denote the value of $\phi$ at which the sign of this first
  derivative changes as $\phi^{\star}$:
  \begin{align}
    \frac{q_{1}(\phi^{\star})}{q_{0}(\phi^{\star})} = \frac{U_{0\hat{0}} -
      U_{0\hat{1}}}{U_{1\hat{0}} - U_{1\hat{1}}}
  \end{align}
  Moreover, it follows that
  \begin{subequations}
  \begin{align}
    \phi < \phi^{\star} &\implies \dd{}{\phi} W_{1} - W_{0} > 0 \\ \phi >
    \phi^{\star} &\implies \dd{}{\phi} W_{1} - W_{0} < 0
  \end{align}
  \end{subequations}
  $W_{1} - W_{0}$ is therefore strictly quasi-concave, from which it follows
  that only two zeros of the function can exist (\emph{By contradiction, more
  than two zeros would require the function, which has no discontinuities, to
  invert its slope more than once.})

  For completeness, we may also take a second derivative of $W_{1} - W_{0}$ with
  respect to $\phi$:
   \begin{equation}
  \begin{aligned}
    \dd[2]{}{\phi} \Big( W_{1}(\phi) - W_{0}(\phi) \Big) &= \dd{}{\phi}\bigg(
    \frac{q_{1}(\phi)}{q_{0}(\phi)} \bigg) \bigg( q_{0}(\phi) (U_{1\hat{0}} -
    U_{1\hat{1}}) \bigg) \\ &~+ \bigg( \frac{q_{1}(\phi)}{q_{0}(\phi)} -
    \frac{U_{0\hat{0}} - U_{0\hat{1}}}{U_{1\hat{0}} - U_{1\hat{1}}} \bigg)
    \bigg(\dd{}{\phi} q_{0}(\phi) \bigg)
  \end{aligned}
\end{equation}
Doing so, we observe that $W_{1} - W_{0}$ may have any number of inflection
points, but $\phi^{\star}$ cannot be one of them. We see this because the second
term of the expression above evaluated at $\phi^{\star}$ must be zero, but the
first term must be non-zero by \cref{asm:q1-q0-monotonic} and \cref{asm:UUneq}.
It follows that $\phi^{\star}$ is the unique occurrence of a local extremum and
therefore a global extremum of $W_{1} - W_{0}$.

  \end{proof}
\begin{center}
\textbf{Proof of \cref{thm:yangsthm}}
\end{center}
\textbf{\cref{thm:yangsthm} Statement.}  \thmyangsthm

\begin{proof}[Proof of \cref{thm:yangsthm}]

  Assuming that a state asymptotically approaches the equilibrium hyperplane
  ($\lim_{t' \to \infty} (W_{1} - W_{0}) = 0$), Let us first prove the forward
  direction of the desired mutual implication.  If a state starts with zero
  total qualfication rate disparity, it follows by \cref{rem:persistence} that
  the total qualification rate disparity remains zero for all time, and so
  \begin{equation}
            p \geq 1; \quad \lim_{t' \to \infty} (W_{1} - W_{0}) = 0 \implies
            \Big( \big\|D[t]\big\|_p = 0 \implies \lim_{t' \to \infty}
            \big\|D[t']\big\|_p = 0 \Big)
  \end{equation}
  For the reverse direction, let us define $\mathbf{s}^{\star}$ as the unique
  disparity-free state on the stable internal equilibrium hyperplane, such that
  \begin{align}
    \big\|D(\mathbf{s}^\star)\big\|_p = 0
  \end{align}

  We may phase the assumptions $\lim_{t' \to \infty} (W_{1} - W_{0}) = 0$ and
  $\lim_{t' \to \infty} \|D[t']\|_p= 0$ jointly as the condition
  \begin{align}
    \lim_{t' \to \infty} \mathbf{s} = \mathbf{s^{\star}}
  \end{align}

  By the Weierstrass definition of a limit, this is
  \begin{align}
    \forall \varepsilon > 0, \exists t_{0}, \forall t > t_{0}, \quad \|
    \mathbf{s} -  \mathbf{s^{\star}} \|_{p} < \varepsilon.
  \end{align}

  For any $\varepsilon$, we have thus assumed that there exists some time
  $t_{0}$ beyond which $\mathbf{s}$ is within $\varepsilon$ of
  $\mathbf{s^{\star}}$.  In particular, we are free to choose $\varepsilon$
  small enough that the local dynamics of the system are well approximated by
  the linearization undertaken in the proof of \cref{thm:jacobian}:

  Because the system is well approximated to first order within any
  sufficiently-small $\varepsilon$-neighborhood of the equilibrium hyperplane,
  the preimage of $\mathbf{s^{\star}}$ in the infinite-time limit within this
  neighborhood lies along the line through $\mathbf{s^{\star}}$ parallel to the
  \emph{sole} eigenvector of the Jacobian with non-zero eigenvalue:
  $\mathbf{v}$.

When $\forall g, \delta(g, g +1) = 0$, $\mathbf{v}$ is orthogonal to the
internal equilibrium hyperplane (\cref{eq:jacobian}), therefore, all states in
the preimage of \(\mathbf{s}^\star\) also satisfy $\forall g, \delta(g, g +1) =
0$ and exhibit zero \(p\)-total qualification rate disparity. We may then appeal
to induction and \cref{rem:persistence} to note that the entire trajectory of of
the state must have had zero total disparity.
  \begin{equation}
            p \geq 1; \quad \lim_{t' \to \infty} (W_{1} - W_{0}) = 0 \implies
            \Big( \big\|D[t]\big\|_p = 0 \impliedby \lim_{t' \to \infty}
            \big\|D[t']\big\|_p = 0 \Big)
  \end{equation}
  This completes the proof.
\end{proof}

\begin{center}
\textbf{A Series of Lemmas for Linear Stability Analysis}
\end{center}
\setcounter{lem}{0} \setcounterref{thm}{thm:jacobian}

\begin{lem}\label{lem:phiWW}
  \begin{align}
    \pp{}{\phi} \frac{W_{1}}{\avg{W}_{g}} \beq =
    \frac{1}{\weq}(1-s_{g})\pp{}{\phi}(W_{1} - W_{0}) \eq
  \end{align}
\end{lem}

\begin{proof}[Proof of \cref{lem:phiWW}]
  We directly differentiate the expression evaluated at equilibrium, recalling
  that \({\weq = W_{0} \seq = W_{1} \seq = \avg{W}_{g} \seq ~ \forall g \in
    \mathcal{G}}\) and \({W_g = s_g W_1 + (1 - s_g) W_0}\).
  \begin{subequations}
    \begin{align}
    \pp{}{\phi} \frac{W_{1}}{\avg{W}_{g}} \beq &= \frac{1}{\weq}
    \pp{W_{1}}{\phi} \beq - \frac{1}{\weq} \pp{\avg{W}_g }{\phi} \beq \\ &=
    \frac{1}{\weq} \pp{}{\phi} \Big(W_1 - s_g W_1 - (1 - s_g) W_0 \Big) \beq
    \\ &= \frac{1}{\weq}(1-s_{g})\pp{}{\phi}(W_{1} - W_{0}) \eq
    \end{align}
  \end{subequations}
\end{proof}

\begin{lem} \label{lem:ppchcoords}
  $\forall g \in \mathcal{G}$,
  \begin{align}
    \pp{\avg{s}}{s_{g}} = \mu_{g}, \quad \pp{\delta(g, g+1)}{s_{g}} = 1, \quad
    \pp{\delta(g-1, g)}{s_{g}} = -1
  \end{align}
\end{lem}
\begin{proof}[Proof of \cref{lem:ppchcoords}]
  By \cref{eq:def-sg} and \cref{eq:def-delta}, the result is immediate.
\end{proof}

\begin{lem} \label{lem:ppchcoords2}
  \begin{align}
    \pp{s_{g}}{\avg{s}} = 1
  \end{align}
\end{lem}
\begin{proof}[Proof of \cref{lem:ppchcoords2}]
  By \cref{eq:changecoords}, the result is immediate.
\end{proof}
\begin{lem} \label{lem:ppchcoords3}
  \begin{align}
    \pp{s_{g}}{\delta(g,h)} = \begin{cases} 1 - \mu_{1} - \mu_{2} - ... -
      \mu_{g} & h = g + 1 \\ \mu_{1} + \mu_{2} + ... + \mu_{g - 1} & h = g - 1
    \end{cases}
    = \begin{cases} 1 + \beta_g & h = g + 1 \\ -\mu_g - \beta_g & h = g - 1
    \end{cases}
  \end{align}
  \begin{align}
    \pp{s_{h}}{\delta(g,h)} = \begin{cases} - \mu_{1} - \mu_{2} - ... - \mu_{g}
      & h = g + 1 \\ - 1 + \mu_{1} + \mu_{2} + ... + \mu_{g - 1} & h = g - 1
  \end{cases}
    = \begin{cases} \beta_g & h = g + 1 \\ -\mu_g -\beta_g - 1 & h = g - 1
    \end{cases}
  \end{align}
\end{lem}
\begin{proof}[Proof of \cref{lem:ppchcoords3}]
  The result follows from \cref{eq:changecoords}, noting ${\delta(g,h) =
    -\delta(h, g)}$.
\end{proof}

\begin{lem} \label{lem:ppsg}
  Taking a partial derivative with respect to $s_{g}$ while holding all other
  $s_{h}, h \neq g$ fixed,
  \begin{align}
    \pp{}{s_{g}[t]} s_{g}[t+1] \beq = 1 + \mu_{g} \frac{1}{\weq} \bigg(
    \pp{\phi}{\avg{s}} \bigg) s_{g}(1-s_{g}) \pp{}{\phi} (W_{1} - W_{0}) \beq
  \end{align}
  Holding $\phi$ constant as well,
  \begin{align}
    \bigg(\pp{}{s_{g}[t]}\bigg)_\phi s_{g}[t+1] \beq = 1
  \end{align}
  When $\phi$ is held constant when taking a partial derivative with respect to
  $s_{g}$, we shall denote the partial derivative with $\phi$ in the subscript,
  as in the equation above, and omit this subscript otherwise.
\end{lem}

\begin{proof}[Proof of \cref{lem:ppsg}]
  Let us begin by proving the second equality, observing first that
  \begin{align}
    \pp{}{s_{g}}\avg{W}_{g} = \pp{}{s_{g}}\bigg(s_{g}W_{1} + (1 - s_{g}) W_{0}
    \bigg) = W_{1} - W_{0}
  \end{align}
  With \(\phi\) fixed, \(W_1\) does not depend on \(s_g\). Therefore,
  substituting \(s_g[t+1] = s_g \frac{W_1}{\avg{W}_g}\),
  \begin{subequations}
  \begin{align}
    \bigg( \pp{}{s_{g}}\bigg)_{\phi} \bigg( s_{g} \frac{W_{1}}{\avg{W}_{g}}
    \bigg) \beq &= \frac{W_{1}}{\avg{W}_{g}} \beq -
    s_{g}\frac{W_{1}}{\avg{W}_{g}^{2}}(W_{1} - W_{0}) \beq \\ &=
    \frac{\weq}{\weq} - 0 \\ &= 1
  \end{align}
  \end{subequations}
  We next address the first equality. By \cref{lem:ppchcoords},
  \begin{align}
    \bigg( \pp{\phi}{s_{g}} \bigg) &= \bigg( \pp{\avg{s}}{s_{g}} \bigg) \bigg(
    \pp{\phi}{\avg{s}} \bigg) = \mu_{g} \bigg( \pp{\phi}{\avg{s}} \bigg)
  \end{align}
  By \cref{lem:phiWW},
  \begin{subequations}
  \begin{align}
    \bigg(\pp{}{s_{g}}\bigg) \bigg( s_{g} \frac{W_{1}}{\avg{W}_{g}} \bigg) \beq
    &= \bigg( \pp{}{s_{g}}\bigg)_\phi \bigg( s_{g} \frac{W_{1}}{\avg{W}_{g}}
    \bigg) \beq  + s_{g}\bigg( \pp{\phi}{s_{g}} \bigg) \bigg( \pp{}{\phi}
    \frac{W_{1}}{\avg{W_{g}}} \bigg) \beq  \\ &= 1 + \mu_{g} \frac{s_{g}}{\weq}
    \bigg( \pp{\phi}{\avg{s}} \bigg) (1-s_{g}) \pp{}{\phi} (W_{1} - W_{0}) \beq
  \end{align}
  \end{subequations}
\end{proof}

\begin{fact}\label{fact:chainrule}
  We note when differentiating an expression $\mathfrak{g}$ with respect to an
  expression $\mathfrak{f}$, each involving each $s_{g}$ and $\phi$ (which
  depends on each $s_{g}$), we may invoke the chain rule to treat $\phi$ as an
  independent function input from the beginning, or we may treat the effect on
  $\phi$ due to perturbation of each $s_{g}$ separately. It is for this reason
  that we have been explicit about which variables are fixed in the partial
  derivatives of \cref{lem:ppsg}.
  \begin{align}
    \pp{}{s_{g}} \mathfrak{f}(s_{g}, \phi) &=
    \pp{\phi}{s_{g}}\pp{\mathfrak{f}}{\phi} + \bigg( \pp{\mathfrak{f}}{s_{g}}
    \bigg)_{\phi} \\ \pp{\phi}{\mathfrak{f}} &= \sum_{g \in \mathcal{G}}
    \pp{s_{g}}{\mathfrak{f}} \pp{\phi}{s_{g}}
  \end{align}
  Therefore,
  \begin{subequations}
  \begin{align}
    \pp{}{\mathfrak{f}} \mathfrak{g}(s_{g}, \phi) &= \sum_{g \in \mathcal{G}}
    \bigg( \pp{s_{g}}{\mathfrak{f}} \bigg) \pp{}{s_{g}} \mathfrak{g}(s_{g},
    \phi) \\ &= \sum_{g \in \mathcal{G}} \bigg( \pp{s_{g}}{\mathfrak{f}} \bigg)
    \Bigg( \pp{\phi}{s_{g}}\pp{\mathfrak{g}}{\phi} + \bigg(
    \pp{\mathfrak{g}}{s_{g}} \bigg)_{\phi} \Bigg) \\ &= \sum_{g \in \mathcal{G}}
    \bigg( \pp{s_{g}}{\mathfrak{f}} \pp{\phi}{s_{g}} \bigg)
    \pp{\mathfrak{g}}{\phi} + \sum_{g \in \mathcal{G}}  \bigg(
    \pp{s_{g}}{\mathfrak{f}} \bigg) \bigg( \pp{\mathfrak{g}}{s_{g}}
    \bigg)_{\phi} \\ &= \pp{\phi}{\mathfrak{f}} \pp{\mathfrak{g}}{\phi} +
    \sum_{g \in \mathcal{G}}  \bigg( \pp{s_{g}}{\mathfrak{f}} \bigg) \bigg(
    \pp{\mathfrak{g}}{s_{g}} \bigg)_{\phi} \label{eqA:independentchain}
  \end{align}
  \end{subequations}
  For convenience, we will treat $\phi$ as an independent function input (\ie,
  we will invoke the chain rule as in \cref{eqA:independentchain}) when proving
  \cref{lem:savgsavg}, \cref{lem:deltasavg}, and \cref{lem:delta-neutral}.
\end{fact}

\begin{lem} \label{lem:savgsavg}
  \begin{align} \label{eqA:savgsavg}
    \pp{(\avg{s}[t + 1] - \avg{s}[t])}{\avg{s}[t]} \beq = \frac{1}{\weq} \bigg(
    \pp{\phi}{\avg{s}} \bigg) \bigg( \sum_{g\in\mathcal{G}} \mu_{g}s_{g}(1 -
    s_{g}) \bigg) \pp{}{\phi}(W_{1} - W_{0}) \eq
  \end{align}
\end{lem}

\begin{proof}[Proof of \cref{lem:savgsavg}]
  Noting that $\avg{s}[t+1]$ depends on each $s_{g}$ and $\phi$,
  \begin{align}
    \avg{s}[t + 1] = \sum_{g\in\mathcal{G}} \mu_{g} s_{g}
    \frac{W_{1}(\phi)}{\avg{W}_{g}(\phi)}
  \end{align}
  we may use the chain rule (\cref{fact:chainrule}),
  \begin{align}
    \pp{}{\avg{s}} f(s_{1}, s_{2}, ..., s_{n}, \phi) &= \sum_{g\in\mathcal{G}}
    \pp{s_{g}}{\avg{s}} \bigg( \pp{f}{s_{g}} \bigg)_{\phi} + \pp{\phi}{\avg{s}}
    \pp{f}{\phi}
  \end{align}
  to compute, referencing \cref{lem:phiWW}, \cref{lem:ppchcoords},
  \cref{lem:ppchcoords2}, and \cref{lem:ppsg},
  \begin{subequations}
  \begin{align}
    &\pp{(\avg{s}[t + 1] - \avg{s}[t])}{\avg{s}[t]} \beq = \pp{\avg{s}[t +
        1]}{\avg{s}[t]} - 1  \beq \\ &\quad= \Bigg( \sum_{g\in\mathcal{G}}
    \bigg(\pp{s_{g}}{\avg{s}} \bigg) \bigg( \pp{}{s_{g}} \bigg)_{\phi}
    \bigg(\mu_{g} s_{g} \frac{W_{1}}{\avg{W}_{g}} \bigg) + \bigg(
    \pp{\phi}{\avg{s}} \bigg) \bigg( \pp{}{\phi} \sum_{g\in\mathcal{G}} \mu_{g}
    s_{g} \frac{W_{1}}{\avg{W}_{g}} \bigg) - 1 \Bigg) \beq \\ &\quad=
    \sum_{g\in\mathcal{G}} \mu_{g} + \bigg( \pp{\phi}{\avg{s}} \bigg)
    \frac{1}{\weq} \sum_{g\in\mathcal{G}} \mu_{g}s_{g}(1 - s_{g})
    \pp{}{\phi}(W_{1} - W_{0})\beq - 1 \\ &\quad= \frac{1}{\weq} \bigg(
    \pp{\phi}{\avg{s}} \bigg) \bigg( \sum_{g\in\mathcal{G}} \mu_{g}s_{g}(1 -
    s_{g}) \bigg) \pp{}{\phi}(W_{1} - W_{0}) \eq
  \end{align}
\end{subequations}
\end{proof}

\begin{lem} \label{lem:deltasavg}
    \begin{subequations}
    \begin{align} \label{eqA:deltasavg}
      &\pp{}{\avg{s}[t]} \bigg(\delta(g,h)[t + 1] - \delta(g,h)[t]\bigg) \beq
      \\ &\quad= \frac{1}{\weq} \bigg(\pp{\phi}{\avg{s}}\bigg) \delta(g, h)(1 -
      s_{g} - s_{h}) \pp{}{\phi}(W_{1} - W_{0}) \beq
    \end{align}
    \end{subequations}
\end{lem}

\begin{proof}[Proof of \cref{lem:deltasavg}]
  Since $\avg{s}$ and $\delta(g,h)$ are independent coordinates, the partial
  derivative of one with respect to the other at the same time is identically
  zero.
  \begin{align}
    \pp{}{\avg{s}} \delta(g,h) = 0
  \end{align}

  The left hand side of the target equality is therefore equal to
  \begin{align*}
    \pp{}{\avg{s}[t]} \delta(g,h)[t+ 1] \beq
  \end{align*}
  From the chain rule (\cref{fact:chainrule}), \cref{lem:phiWW},
  \cref{lem:ppchcoords2}, and \cref{lem:ppsg}, it follows that
  \begin{subequations}
  \begin{align}
    &\pp{}{\avg{s}[t]} \delta(g,h)[t + 1] \eq \\ &= \Bigg(
    \bigg(\pp{\phi}{\avg{s}}\bigg) \pp{\delta(g, h)[t + 1]}{\phi}  +
    \pp{s_{g}}{\avg{s}}  \bigg(\pp{s_{g}[t+1]}{s_{g}[t]}\bigg)_\phi -
    \pp{s_{h}}{\avg{s}}  \bigg(\pp{s_{h}[t+1]}{s_{h}[t]}\bigg)_\phi \Bigg) \beq
    \\ &= \bigg(\pp{\phi}{\avg{s}}\bigg) \pp{}{\phi} \delta(g, h)[t + 1] \beq
    \\ &= \bigg(\pp{\phi}{\avg{s}}\bigg) \pp{}{\phi} \Big(
    s_{g}\frac{W_{1}}{\avg{W}_{g}} -  s_{h}\frac{W_{1}}{\avg{W}_{h}} \Big) \beq
    \\ &= \frac{1}{\weq} \bigg(\pp{\phi}{\avg{s}}\bigg) \Big( s_{g}(1-s_{g}) -
    s_{h}(1-s_{h}) \Big) \pp{}{\phi}(W_{1} - W_{0}) \eq \\ &= \frac{1}{\weq}
    \bigg(\pp{\phi}{\avg{s}}\bigg) \delta(g, h)(1 - s_{g} - s_{h})
    \pp{}{\phi}(W_{1} - W_{0}) \eq
  \end{align}
  \end{subequations}
\end{proof}

\begin{lem}\label{lem:delta-neutral}
  \begin{align} \label{eqA:ppsavgdelta}
    \bigg( \pp{\avg{s}[t+1] - \avg{s}[t]}{\delta(g,h)[t]} \bigg) \beq &= 0 \\
\label{eqA:deltadeltasame}
    \bigg( \pp{\delta(g,h)[t + 1] - \delta(g,h)[t]}{\delta(g,h)[t]} \bigg) \beq
    &= 0 \\
\label{eqA:deltadeltamixed}
    \bigg( \pp{\delta(h,h+1)[t+1] - \delta(h, h+1)[t]}{\delta(g,g+1)[t]} \bigg)
    \beq &= 0, \quad \forall h \neq g
  \end{align}

\end{lem}

\begin{proof}[Proof of \cref{lem:delta-neutral}]

  By \cref{thm:phi-threshold}, \(\phi\) depends only on \(\avg{s}\), which is
  held constant during partial differentiation by \(\delta(g, h)\). Therefore,
    \begin{align} \label{eqA:phithresholdzero}
      \pp{\phi}{\delta(g,h)} = 0
    \end{align}
    Consider any expression $\mathfrak{f}$ which depends \emph{linearly} on each
    $s_{g}[t+1]$.
    \begin{align}
      \mathfrak{f}\big( s_{g} \colon g \in \mathcal{G} \big) \define \sum_{g \in
        \mathcal{G}} f_{g} s_{g}, \quad f_{g} \in \RR
    \end{align}
    where we introduce a ``vector builder'' notation $\mathbf{s} = \big(s_{g}
    \colon g \in \mathcal{G} \big)$ for brevity. We may use the linearity of
    differentiation to concisely deal with derivatives of linear combinations of
    \(\mathfrak{f}\). We consider will expressions without explicit time
    dependence to correspond to time \([t]\). By the chain rule
    (\cref{fact:chainrule}), \cref{eqA:phithresholdzero}, and \cref{lem:ppsg},
    \begin{subequations}
  \begin{align}
  &\pp{}{\delta(g,h)} \Big(\mathfrak{f}[t + 1] - \mathfrak{f}[t]\Big) \beq
    \\ &\quad = \bigg( \sum_{i\in\mathcal{G}} \pp{s_{i}}{\delta(g, h)}
    \bigg(\pp{}{s_{i}}\bigg)_{\phi} + \pp{\phi}{\delta(g,h)} \pp{}{\phi} \bigg)
    \mathfrak{f} \Big( s_{i}[t+1] - s_{i}[t] \colon i \in \mathcal{G}\Big) \beq
    \\ &\quad = \mathfrak{f}\bigg( \sum_{i\in\mathcal{G}} \pp{s_{i}}{\delta(g,
      h)} \bigg( \pp{s_{i}[t+1]}{s_{i}[t]} - \pp{s_{i}[t]}{s_{i}[t]} \bigg)
    \colon i \in \mathcal{G} \bigg) \\ &\quad = \mathfrak{f}(0 \colon i \in
    \mathcal{G}) = 0
  \end{align}
  \end{subequations}
    We conclude that perturbing any $\delta$ while holding $\avg{s}$ constant
    has no effect on the evolution of dynamical variables that are linear in
    $\mathbf{s}$ at equilibrium. This includes each $\delta$ and $\avg{s}$.

\end{proof}

\begin{center}
\textbf{Proof of \cref{thm:jacobian}}
\end{center}
\setcounterref{thm}{thm:jacobian} \textbf{\cref{thm:jacobian} Statement.}  The
Jacobian \(J\) simplifies to a scalar multiplied by a matrix with a single
non-zero column \(\mathbf{v}\) in the last position.
  \begin{align} \label{eqA:jacobian}
    J \beq =\frac{1}{\weq} \bigg( \dd{\phi}{\avg{s}}\bigg) \bigg( \dd{}{\phi}
    (W_{1} - W_{0}) \bigg) \Bigg[ \mathbf{0}^{(n \times n - 1)} \Bigg| \mathbf{v
      }\Bigg], ~~~ \mathbf{v} \define \begin{bsmallmatrix} \delta(1,2)(1 - s_{1}
      - s_{2})\\ \delta(2,3)(1 - s_{2} - s_{3})\\ ... \\ \delta(n-1,n)(1 -
      s_{n-1} - s_{n})\\ \sum_{g \in \mathcal{G}} \mu_{g} s_{g}( 1 - s_{g} )
  \end{bsmallmatrix}
  \end{align}

\begin{proof}[Proof of \cref{thm:jacobian}]
  The zero entries in the Jacobian matrix are a consequence of
  \cref{lem:delta-neutral}.

\cref{lem:deltasavg} and \cref{lem:savgsavg} provide us with the last column of
the matrix $J$ in the desired form:
\begin{align}
      \pp{\delta(g,h)[t + 1] - \delta(g,h)[t]}{\avg{s}[t]} \eq &= \frac{1}{\weq}
      \bigg(\pp{\phi}{\avg{s}}\bigg) \delta(g, h)(1 - s_{g} - s_{h})
      \pp{}{\phi}(W_{1} - W_{0}) \eq \\ \pp{(\avg{s}[t + 1] -
        \avg{s}[t])}{\avg{s}[t]} \beq &= \frac{1}{\weq} \bigg(
      \pp{\phi}{\avg{s}} \bigg) \bigg( \sum_{g\in\mathcal{G}} \mu_{g}s_{g}(1 -
      s_{g}) \bigg) \pp{}{\phi}(W_{1} - W_{0}) \eq
\end{align}
\end{proof}

\begin{center}
\textbf{Proof of \cref{cor:jacob-inspect}}
\end{center}
\textbf{\cref{cor:jacob-inspect} Statement.}

At equilibrium, any state displacement vector with zero \(\avg{s}\) component is
an eigenvector of \(J\) with eigenvalue 0, while \(\mathbf{v}\) is an
eigenvector of \(J\) with eigenvalue \(\lambda\):
  \begin{align}\label{eqA:lambda}
    \lambda \define \bigg( \sum_{g \in \mathcal{G}} \mu_{g} s_{g}( 1 - s_{g}
    )\bigg) \frac{1}{\weq} \bigg( \dd{\phi}{\avg{s}}\bigg) \bigg( \dd{}{\phi}
    (W_{1} - W_{0}) \bigg) \beq
  \end{align}

\begin{proof}[Proof of \cref{cor:jacob-inspect}]
\textbf{\cref{cor:jacob-inspect}} follows by inspection of $J$ in
\cref{eqA:jacobian}.
\end{proof}

\begin{center}
\textbf{Proof of \cref{cor:phiplusstable}}
\end{center}
\textbf{\cref{cor:phiplusstable} Statement.}  \corphiplusstable

\begin{proof}[Proof of \cref{cor:phiplusstable}]
This is a consequence of \cref{cor:ddphis} and \cref{cor:jacob-inspect} given
\cref{asm:s-interior} (each \(s_g\) is interior) and the restriction of $W_y \in
     [0, \infty)$ as specified in the replicator equation
       (\cref{eq:replicator}). The eigenvalue $\lambda$ is negative, (and the
       associated equilibrium hyperplane stable) iff
\begin{align}
    \dd{}{\phi}(W_{1} - W_{0}) \eq > 0
\end{align}
  This prescribes precisely the value $\phi^{+}$ for the stable equilibrium
  hyperplane.
\end{proof}

\begin{center}
\textbf{Proof of \cref{thm:eq-opt}}
\end{center}
\setcounterref{thm}{thm:eq-opt} \textbf{\cref{thm:eq-opt} Statement.}  \thmeqopt

\begin{proof}[Proof of \cref{thm:eq-opt}]
  The forward direction (group-independence satisfies Equalized Odds) follows
  from the group-independence of \(Q_y\) (\cref{defi:Q}). The reverse direction
  follows from the same; specifically, as functions of \(\phi\),
  \begin{align}
    \Pr(\hat{Y} = 0 \mid Y = y) &= Q_y(\phi) \\ \Pr(\hat{Y} = 1 \mid Y = y) &=
    (1 - Q_y(\phi))
  \end{align}
  are each monotonic, and any specified value of \(\Pr(\hat{Y} = \hat{y} \mid Y
  = y)\) corresponds to a unique \(\phi\) value that must be shared by all
  groups.
\end{proof}

\begin{center}
\textbf{Proof of \cref{cor:eq-odds}}
\end{center}
\textbf{\cref{cor:eq-odds} Statement.}  Equalized Odds does not imply long-term
fairness in our model.

\begin{proof}[Proof of \cref{cor:eq-odds}]
  By contradiction, we have shown that a group-independent threshold policy
  satisfies Equalized Odds (\cref{thm:eq-opt}), yet long-term fairness is
  violated by persistent qualification rate disparities (\cref{thm:yangsthm}).
\end{proof}

\begin{center}
\textbf{Proof of \cref{thm:intervention}}
\end{center}
\setcounterref{thm}{thm:intervention} \textbf{\cref{thm:intervention}
  Statement.}  \thmintervention

\begin{proof}[Proof of \cref{thm:intervention}]
  We note that the a perturbation to $\phi$ at internal equilibrium causes a
  change in state vector parallel to the eigenvector $\mathbf{v}$, where
  \begin{equation}
    \mathbf{v} = \pp{\mathbf{r}}{\avg{s}} =
    \begin{bmatrix}
    \delta(1,2)(1 - s_{1} - s_{2})\\ \delta(2,3)(1 - s_{2} - s_{3})\\ ...
    \\ \delta(n-1,n)(1 - s_{n-1} - s_{n})\\ \sum_{g \in \mathcal{G}} \mu_{g}
    s_{g}( 1 - s_{g} )
  \end{bmatrix}
  \end{equation}
  By use of the chain rule with \cref{lem:deltasavg}, or direct application of
  \cref{lem:phiWW}, we note
  \begin{subequations}
  \begin{align}
    \label{eqA:ppsdelta}
    &\pp{}{\phi} \bigg( \delta(g,h)[t+1] - \delta(g,h)[t] \bigg) \beq = \bigg(
    \pp{\phi}{\avg{s}} \bigg)^{-1} \pp{\delta(g,h)[t + 1] -
      \delta(g,h)[t]}{\avg{s}[t]} \beq \\ &\quad = \frac{1}{\weq}
    \pp{}{\phi}(W_{1} - W_{0})\eq \bigg( s_{g}(1 - s_{g}) - s_{h}(1 - s_{h})
    \bigg) \beq \\ &\quad= \frac{1}{\weq} \pp{}{\phi}(W_{1} - W_{0}) \beq
    \delta(g, h)(1 - s_{g} - s_{h})
  \end{align}
  \end{subequations}
  Likewise, pairing the chain rule with \cref{lem:savgsavg} or directly applying
  \cref{lem:phiWW}, we note
  \begin{subequations}
  \begin{align}
      \label{eqA:ppsphi}
      &\pp{}{\phi} \bigg( \avg{s}[t+1] -  \avg{s}[t] \bigg) \beq = \bigg(
      \pp{\phi}{\avg{s}} \bigg)^{-1} \pp{(\avg{s}[t + 1] -
        \avg{s}[t])}{\avg{s}[t]} \beq \\ &\quad= \frac{1}{\weq}
      \pp{}{\phi}(W_{1} - W_{0}) \beq \sum_{g} \mu_{g}s_{g}(1 - s_{g})
  \end{align}
  \end{subequations}
  Together, our observations imply
  \begin{align}
    \pp{}{\phi}(\mathbf{s}[t + 1] - \mathbf{s}[t])\beq = \frac{1}{\weq}
    \pp{}{\phi}(W_{1} - W_{0}) \beq \mathbf{v}
  \end{align}
  and perturbation of \(\phi\) induces motion parallel to \(\mathbf{v}\). For
  readers familiar with gradient descent but new to linear stability analysis,
  we offer the intuition that \(\mathbf{v}\) is parallel to the gradient of
  \(\phi\) in state space.
\end{proof}

\begin{center}
\textbf{Proof of \cref{thm:dem-parity}}
\end{center}
\setcounterref{thm}{thm:dem-parity} \textbf{\cref{thm:dem-parity} Statement.}
\thmdemparity

\begin{proof}[Proof of \cref{thm:dem-parity}]
  The policy adopted by a classifier subject to demographic parity is given by
  \begin{equation}
  \begin{aligned}
    \pi = \argmax_{\pi} &\sum_{y,\hat{y} = 0}^{1} V_{y\hat{y}}
    \Pr_{\hat{Y}=\pi(X)}(Y = y, \hat{Y} = \hat{y}) \\ \text{subject to}&~~
    \Pr(\hat{Y} = 1 \mid G = g) = \Pr(\hat{Y} = 1 \mid G = h) &\forall g, h \in
    \mathcal{G}
  \end{aligned}
  \end{equation}
  Without allowing group-dependent values of $\phi$, the only solutions to $\pi$
  when groups have differing qualification rates are the trivial policies $\pi =
  0$ and $\pi = 1$.  We therefore consider a solution that permits
  group-dependent thresholds $\phi_g$. We solve for these thresholds using the
  method of Lagrange multipliers.  In the $s_{g}$ state basis, this requires
  that we satisfy, for Lagrange multipliers $L_{h} \in (-\infty, \infty)$, $h
  \in \{1, 2, ... n - 1\}$, the set of equations
  \begin{align}
    \label{eqA:lagrange}
    \nabla_{\boldsymbol{\phi}} u &=  \nabla_{\boldsymbol{\phi}} \left( \sum_{h =
      1}^{n - 1} L_{h} c_{h} \right)
  \end{align}
  where $\nabla_{\boldsymbol{\phi}}$ denotes the vector operator such that the
  $g$th component is the partial derivative with respect to $\phi_{g}$; $u$ is
  the utility to be maximized; and each $c_{h} = 0$ represents a pairwise
  constraint between the probabilities of accepting an agent from two different
  groups ($h$ and $h + 1$).
  \begin{subequations}
  \begin{align}
    u &\define \sum_{y,\hat{y} = 0}^{1} V_{y\hat{y}} \Pr_{\hat{Y}=\pi(X)}(Y = y,
    \hat{Y} = \hat{y}) \\ &= \sum_{g} \mu_{g} \left(
        \begin{aligned}
            &V_{0\hat{0}} (1 - s_{g}) Q_{0}(\phi_{g}) \\ + &V_{0\hat{1}} (1 -
          s_{g}) (1 - Q_{0}(\phi_{g})) \\ + &V_{1\hat{0}} s_{g} Q_{1}(\phi_{g})
          \\ + &V_{1\hat{1}} s_{g} (1 - Q_{1}(\phi_{g}))
        \end{aligned}
        \right)
  \end{align}
  \end{subequations}
  \begin{align}
    c_{h} &\define \left( \begin{aligned} s_{h} Q_{1}(\phi_{h}) &+ (1 -
      s_{h})Q_{0}(\phi_{h}) \\ - s_{h+1} Q_{1}(\phi_{h+1}) &- (1 -
      s_{h+1})Q_{0}(\phi_{h+1})
      \end{aligned} \right)
  \end{align}

  Defining $L_{0} = L_{n} = 0$ for notational convenience, \cref{eqA:lagrange}
  simplifies to a set $n$ equations indexed by $g \in \{1, 2, ..., n\}$
  \begin{subequations}
  \begin{align}
    &\mu_{g}\Big((V_{0\hat{0}} - V_{0\hat{1}}) q_{0}(\phi_{g})(1 - s_{g}) +
    (V_{1\hat{0}} - V_{1\hat{1}}) q_{1}(\phi_{g})s_{g}\Big) \\ &\quad=(L_{g} -
    L_{g-1})\Big( s_{g}q_{1}(\phi_{g}) + (1 - s_{g})q_{0}(\phi_{g}) \Big)
  \end{align}
  \end{subequations}
  From which the perturbed values $\phi_{g}$ may be derived:
  \begin{subequations}
  \begin{align} \label{eqA:demparityperturb}
    \frac{q_1(\phi_g)}{q_0(\phi_{g})} &= \bigg(\frac{V_{0\hat{0}} - V_{0\hat{1}}
      - \gamma_{g}} {V_{1\hat{1}} - V_{1\hat{0}} + \gamma_{g}} \bigg)
    \bigg(\frac{1 - s_{g}}{s_{g}}\bigg) \\ \gamma_{g} &\define \frac{L_{g} -
      L_{g - 1}}{\mu_{g}}; \quad L_g = \pp{u}{c_g}
  \end{align}
  \end{subequations}
  We compare this equation with \cref{eqA:q1-q0-s-theta}, noting that when each
  \(\gamma_g = 0\) (\ie, requiring that constraints \(c_g\) are not active at
  locally optimal utility \(u\)), we recover a Laissez-fair policy:
  \begin{align} \label{eqA:LZ}
    \frac{q_{1}(\phi)}{q_{0}(\phi)} &= \bigg( \frac{V_{0\hat{0}} -
      V_{0\hat{1}}}{V_{1\hat{1}} - V_{1\hat{0}}} \bigg) \bigg( \frac{1 -
      s_g}{s_g} \bigg)
  \end{align}
  For interpretation of the Lagrange multipliers \(L_g\), also known as the
  \emph{dual variables}, we refer the reader to \citet{boyd2004convex}.  By the
  monotonicity of $q_{1}/q_{0}$, the effect of $\gamma_{g}$ in determining
  $\phi_{g}$ is therefore a perturbation to $\phi$, the sign of which is
  inverted relative to the sign of $\gamma_g$.  Finally, having defined \(L_0 =
  L_n = 0\), as a telescoping sum,
  \begin{align}
    \sum_{g = 1}^{n} (L_{g} - L_{g-1}) = 0
  \end{align}
  Therefore,
  \begin{align}
    \sum_{g=1}^{n} \mu_{g} \gamma_{g} = 0
  \end{align}
  This guarantees in turn that set of group-specific \emph{changes} to the
  group-specific values $\phi_g$ defined by a Laissez-Fair policy
  (\cref{eqA:LZ}) must be sign-heterogeneous to satisfy Demographic Parity.
\end{proof}

We comment that a solution for each $\gamma_g$ that satisfying constraints
$c_{g}$ requires the solution of differential equation(s) in $q_y$ (\ie,
equation(s) involving both \(q_y(\phi_g)\) and \(Q_y(\phi_g)\) simultaneously).
This most apparent if we appeal to the chain rule to write
\begin{subequations}
\begin{align}
  L_g &= \pp{u}{c_g} = \pp{s_g}{c_g} \pp{u}{s_g} + \pp{s_{g+1}}{c_g}
  \pp{u}{s_{g+1}} \\ &\quad+ \pp{Q_0(\phi_g)}{c_g} \pp{u}{Q_0(\phi_g)} +
  \pp{Q_1(\phi_g)}{c_g} \pp{u}{Q_1(\phi_g)} \\ &\quad+ \pp{Q_0(\phi_{g+1})}{c_g}
  \pp{u}{Q_0(\phi_{g+1})} + \pp{Q_1(\phi_{g+1})}{c_g} \pp{u}{Q_1(\phi_{g+1})}
\end{align}
\end{subequations}
which, after some simplification, yields an expression in terms of \(Q_y\):
\begin{subequations}
\begin{align}
  & L_g - L_{g-1} =\\ &\quad \bigg(Q_1(\phi_{g+1}) - Q_0(\phi_{g+1})
  \bigg)\bigg(Q_0(\phi_{g+1})\Big( V_{0\hat{1}} - V_{0\hat{0}}\Big) +
  Q_1(\phi_{g+1})\Big( V_{1\hat{0}} - V_{1\hat{1}} \Big)\bigg) \\ &\quad-
  \bigg(Q_1(\phi_{g-1}) - Q_0(\phi_{g-1}) \bigg)\bigg(Q_0(\phi_{g-1})\Big(
  V_{0\hat{1}} - V_{0\hat{0}}\Big) + Q_1(\phi_{g-1})\Big( V_{1\hat{0}} -
  V_{1\hat{1}} \Big)\bigg)
\end{align}
\end{subequations}

Considering that we treat arbitrary $q_y$ subject to \cref{asm:q1-q0-monotonic},
analytically solving an equation in \(q\) and \(Q\) simultaneously is not
practical for our purposes.

\begin{center}
\textbf{Proof of \cref{thm:hyperplane-intervention}}
\end{center}
\setcounterref{thm}{thm:hyperplane-intervention}
\textbf{\cref{thm:hyperplane-intervention} Statement.}
\thmhyperplaneintervention

\begin{proof}[Proof of \cref{thm:hyperplane-intervention}]
  For convenience, on an equilibrium hyperplane, we will write as equivalent
  statements
  \begin{align}
    \pp{}{\phi}(W_1 - W_0) \eq = \pp{}{\phi_g}(W_1^g - W_0^g) \eq
  \end{align}
  We first generalize \cref{lem:phiWW} for group-dependent feature thresholds
  $\phi_{g}$, each perturbed from $\phi_{g} = \phi$ at equilibrium and but
  applied only to the corresponding group $g$.
  \begin{align}
    \label{eqA:ppphig}
    \pp{}{\phi_{g}} \bigg( s_{h}[t+1] - s_{h}[t] \bigg) \eq &= \frac{1}{\weq}
    \pp{}{\phi}(W_{1} - W_{0})\eq  \begin{cases} s_{g}(1 - s_{g}) & g = h \\ 0 &
      h \neq g
    \end{cases}
  \end{align}
  It follows from the definition of \(\avg{s}\) that
  \begin{align}
    \pp{}{\phi_{g}}\Big( \avg{s}[t+1] - \avg{s}[t] \Big) \eq &=
    \frac{1}{\weq}\pp{}{\phi}(W_{1} - W_{0})\eq \mu_{g}s_{g}(1 - s_{g})
  \end{align}
    and, by the definition of \(\delta(h, h+1)\),
  \begin{subequations}
  \begin{align} \label{eqA:ppphigdelta}
    \pp{}{\phi_{g}}\Big( \delta(h,h+1)[t+1] - \delta(h,h+1)[t] \Big) \eq =
    \\ \bigg( \frac{1}{\weq}\pp{}{\phi}(W_{1} - W_{0})\eq \bigg)
      \begin{cases}
        s_{g}(1 - s_{g}) & g = h \\ -s_{g}(1 - s_{g}) & g = h + 1 \\ 0 &
        \text{otherwise}
      \end{cases}
  \end{align}
  \end{subequations}

  We may now prove that perturbation of the vector $\Phi$ by the vector
  $\Delta_{g} \Phi = (\Delta_{g} \phi_{1}, \Delta_{g} \phi_{2}, ..., \Delta_{g}
  \phi_{n})$ causes the system to maintain its current value of $\avg{s}$. We
  sum the contribution due to each $\Delta_{g} \phi_{h}$, where $\langle \cdot,
  \cdot \rangle$ denotes the inner product, $\nabla_{\Phi}$ denotes a gradient
  taken with respect to the components of $\Phi$, and by linearity, \(\langle
  \Delta_{g} \Phi, \nabla_{\Phi}\rangle\) is an operator that perturbs the
  system with change \(\Delta_g \Phi\). Linear proportionality is denoted with
  $\propto$.
  \begin{subequations}
\begin{align}
  \Delta_{g} (\avg{s}[t + 1] - \avg{s}[t]) \eq &= \langle \Delta_{g} \Phi,
  \nabla_{\Phi}\rangle \Big( \avg{s}[t+1] - \avg{s}[t] \Big) \eq \\ &=
  \sum_{h=1}^{n} (\Delta_{g}\phi_{h}) \pp{}{\phi_{h}}\Big(  \avg{s}[t+1] -
  \avg{s}[t] \Big) \eq \\ &= \frac{-\epsilon \delta(g, g +
    1)}{\weq}\pp{}{\phi}(W_{1} - W_{0})\eq \bigg( \sum_{h=1}^{g} \mu_{h}
  \alpha_{g} + \sum_{h=g+1}^{n} \mu_{h} \beta_{g} \bigg) \\ &\propto (-\beta_g
  \alpha_g + \alpha_g \beta_g) \\ &= 0
\end{align}
  \end{subequations}
Next, by \cref{eqA:ppphigdelta}, we consider the effect that the perturbation
$\Delta_{g} \Phi$ has on each $\delta(h,h+1)$ at equilibrium.
  \begin{subequations}
\begin{align}
& \Delta_g (\delta(h, h + 1)[t + 1] - \delta(h, h + 1[t]))\eq \\ &=\langle
  \Delta_{g} \Phi, \nabla_{\Phi}\rangle \Big( \delta(h, h+1)[t+1] - \delta(h,
  h+1)[t] \Big) \eq \\ &=\sum_{i=1}^{n} (\Delta_{g} \phi_{i}) \pp{}{\phi_{i}}
  \Big( \delta(h, h+1)[t+1] - \delta(h, h+1)[t] \Big) \eq \\ &= \bigg(
  \frac{-\epsilon \delta(g, g + 1)}{\weq}\pp{}{\phi}(W_{1} - W_{0})\eq \bigg)
      \begin{cases}
        \alpha_{g} & h \leq g \\ \beta_{g} & h > g
      \end{cases} \\
  &\quad\quad\quad -     \bigg( \frac{-\epsilon \delta(g, g +
        1)}{\weq}\pp{}{\phi}(W_{1} - W_{0})\eq \bigg)
      \begin{cases}
        \alpha_{g} & h + 1 \leq g\\ \beta_{g}  & h + 1 > g
      \end{cases} \\
  &= \bigg( \frac{-\epsilon \delta(g, g + 1)}{\weq}\pp{}{\phi}(W_{1} - W_{0})\eq
      \bigg)
      \begin{cases}
        \alpha_{g} - \beta_{g} & g = h \\ 0 & g \neq h \\
      \end{cases}
\end{align}
  \end{subequations}
Since $\alpha_{g} - \beta_{g} = 1$ by \cref{eq:def-mu}, We see that the discrete
velocity in $\delta(g,g+1)$ induced by $\Delta_{g} \Phi$ is
\begin{align}
 -\epsilon \frac{\delta(g, g+1)}{\weq}\pp{}{\phi}(W_{1} - W_{0})\eq
\end{align}
On the stable equilibrium hyperplane, where $\pp{}{\phi}(W_{1} - W_{0}) > 0$ and
initial discrete velocity in \(\delta(g, g+1)\) is zero, the prescribed
perturbation proportionately \emph{opposes} $\delta(g,g+1)$.
\end{proof}

\newpage
\section{Additional Figures}\label{Asec:figures}

For all settings, we display the simulated dynamics for two groups, subject to
different global interventions. Streamlines approximate system time evolution.
$q_{0}$ and $q_{1}$ are Gaussians with unit variance and have means $-1$ and
$1$, respectively. The figures included herein are provided with little analysis
and are intended to prompt further consideration for the curious reader.

\subsection{Additional Variables of Interest}
In addition to the acceptance rate for Group 1 (blue; first row), we plot the
false positive rate for Group 1 (red; second row) and the false negative rate
for Group 1 (green; third row).

\begin{figure}[H]
\centering \hfill
\begin{subfigure}{0.3\textwidth}
\begin{align*}
    &~\begin{bmatrix} \mu_1 = 0.5 & \mu_2 = 0.5
    \end{bmatrix}
    \\ &\begin{bmatrix} U_{ 0, \hat{0} } = 0.1 & U_{ 0, \hat{1} } = 5.5 \\ U_{
        1, \hat{0} } = 0.5 & U_{ 1, \hat{1} } = 1.0
    \end{bmatrix}
    \\ &\begin{bmatrix} V_{ 0, \hat{0} } = 0.5 & V_{ 0, \hat{1} } = -0.5 \\ V_{
        1, \hat{0} } = -0.25 & V_{ 1, \hat{1} } = 1.0
    \end{bmatrix}
\end{align*}
\end{subfigure}
\hfill
\begin{subfigure}{0.5\textwidth}
    \includegraphics[width=0.8\textwidth]{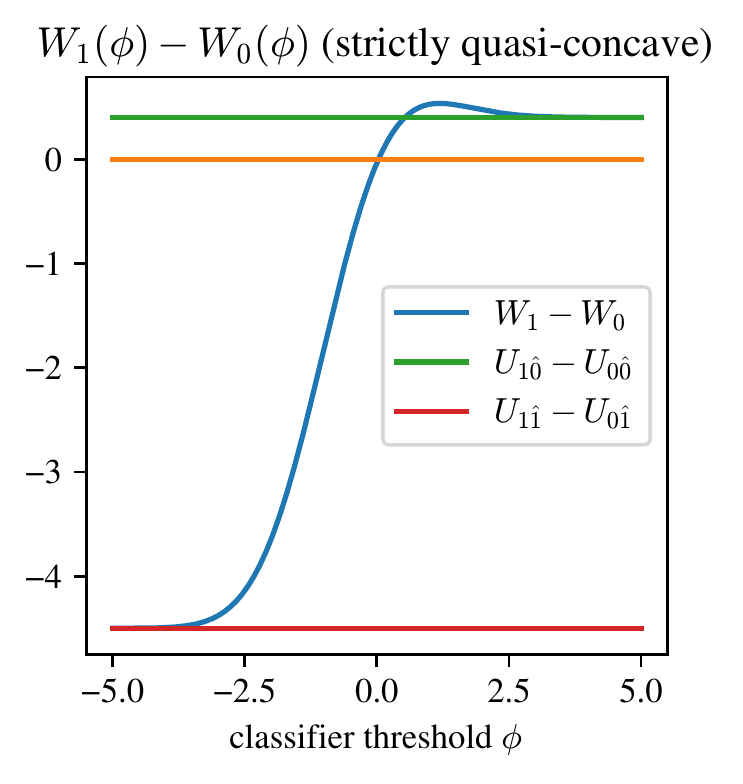}
\end{subfigure}
\hfill
\begin{subfigure}{\textwidth}
    \includegraphics[width=\textwidth, trim=5 5 5 5, clip]{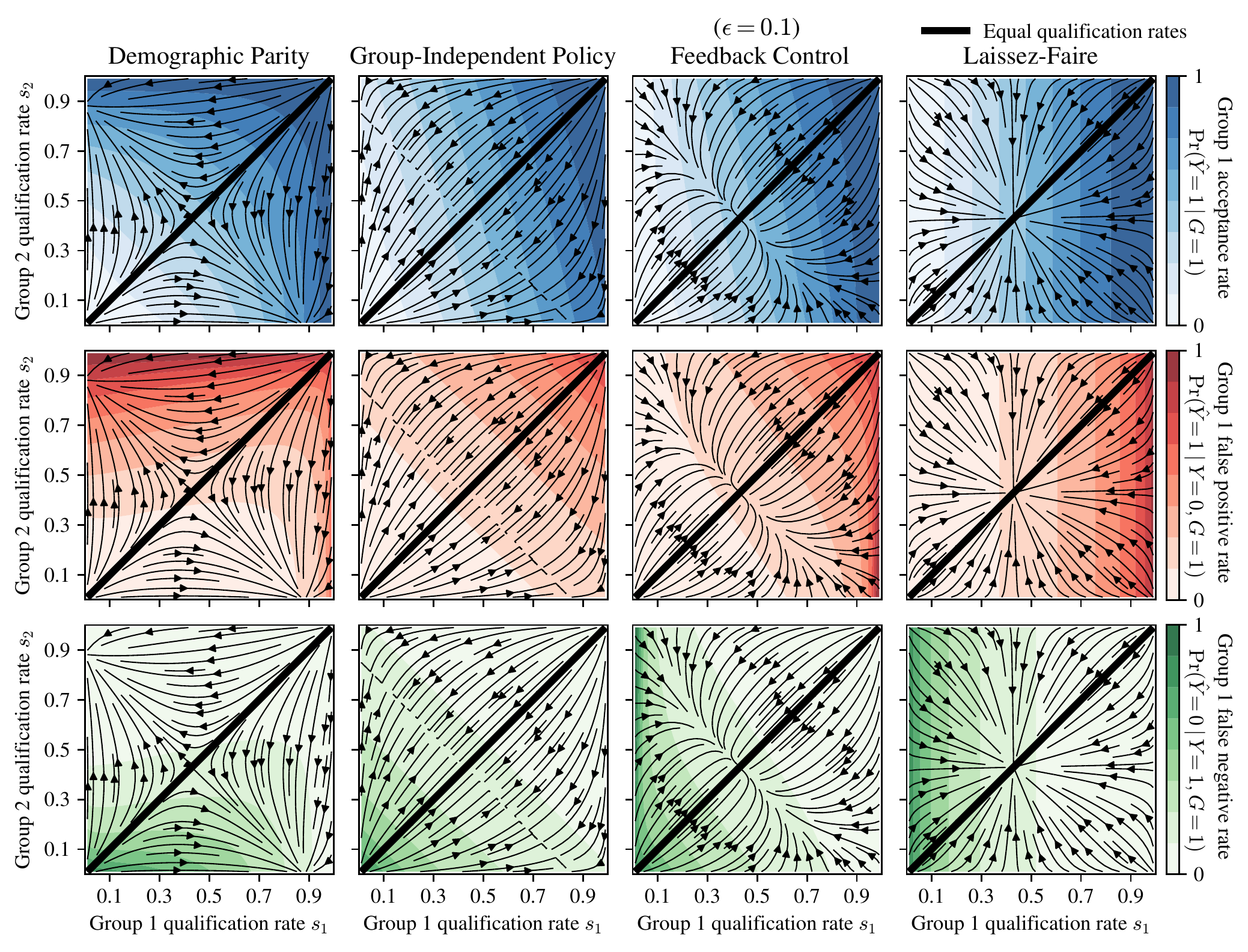}
\end{subfigure}

\caption{\textbf{Setting 1} (Analyzed in \cref{sec:interventions})}

\end{figure}

\newpage
\subsection{Different \texorpdfstring{\(U\)}{U} and \texorpdfstring{\(V\)}{V}}
Classifier decisions will differ for a given state \(\mathbf{s}\) when \(V\) is
modified. Similarly, the success of different strategies update with different
\(U\) values. The qualitative behavior of the system ultimately depends on the
shape of \(W_1 - W_0\) as a function of \(\phi\).

\vspace*{\fill}
\begin{figure}[H]
\centering \hfill
\begin{subfigure}{0.3\textwidth}
  \begin{align*}
    &~\begin{bmatrix} \mu_1 = 0.5 & \mu_2 = 0.5
    \end{bmatrix}
    \\ &\begin{bmatrix} U_{ 0, \hat{0} } = 0.5 & U_{ 0, \hat{1} } = 1.5 \\ U_{
        1, \hat{0} } = 0.1 & U_{ 1, \hat{1} } = 1.0
    \end{bmatrix}
    \\ &\begin{bmatrix} V_{ 0, \hat{0} } = 1 & V_{ 0, \hat{1} } = 0 \\ V_{ 1,
        \hat{0} } = 0 & V_{ 1, \hat{1} } = 1
    \end{bmatrix}
      \end{align*}
\end{subfigure}
\hfill
\begin{subfigure}{0.5\textwidth}
    \includegraphics[width=0.8\textwidth]{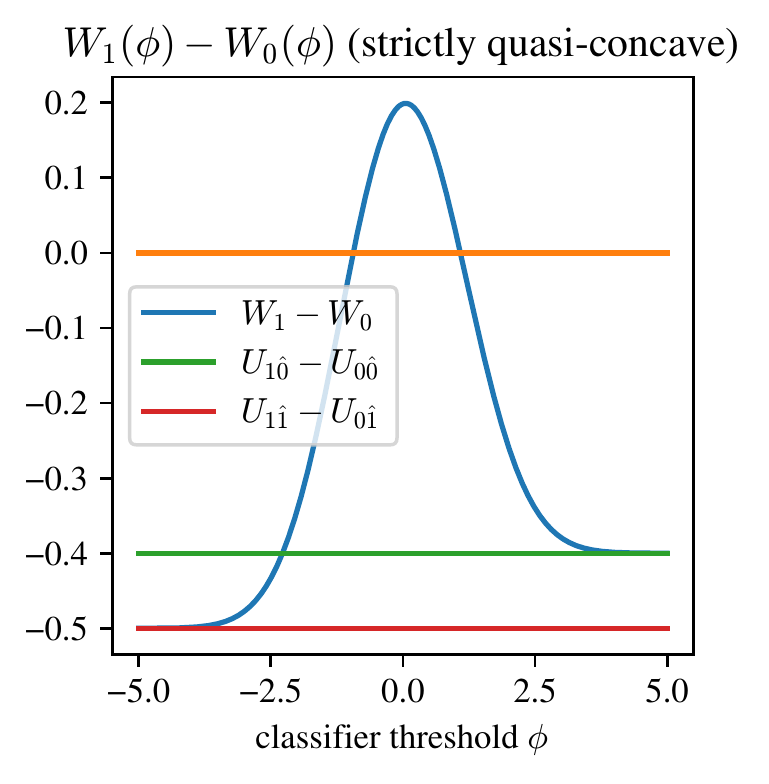}
\end{subfigure}
\hfill
\begin{subfigure}{\textwidth}
    \includegraphics[width=\textwidth, trim=5 5 5 5, clip]{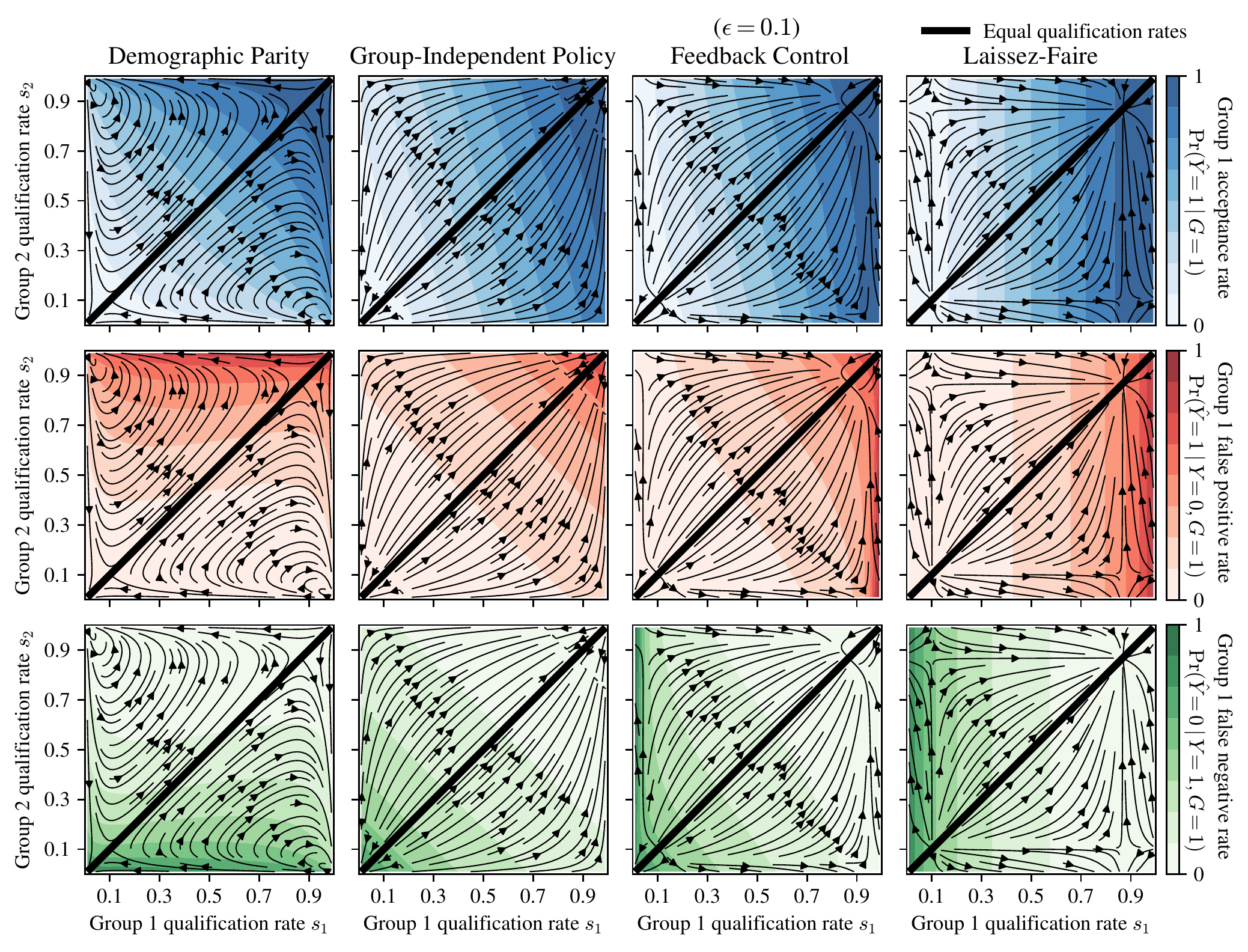}
\end{subfigure}

\caption{\textbf{Setting 2} (Stable \emph{and} unstable hyperplanes)}

\end{figure}
\vspace*{\fill}

\newpage
\vspace*{\fill}
\begin{figure}[H]
\hfill \centering
\begin{subfigure}{0.3\textwidth}
  \begin{align*}
    &~\begin{bmatrix} \mu_1 = 0.5 & \mu_2 = 0.5
    \end{bmatrix}
    \\ &\begin{bmatrix} U_{ 0, \hat{0} } = 0.5 & U_{ 0, \hat{1} } = 0.5 \\ U_{
        1, \hat{0} } = 0.1 & U_{ 1, \hat{1} } = 1.5
    \end{bmatrix}
    \\ &\begin{bmatrix} V_{ 0, \hat{0} } = 10.0 & V_{ 0, \hat{1} } = 0.0 \\ V_{
        1, \hat{0} } = 1.0 & V_{ 1, \hat{1} } = 1.5
    \end{bmatrix}
  \end{align*}
\end{subfigure}
\hfill
\begin{subfigure}{0.5\textwidth}
    \includegraphics[width=0.8\textwidth]{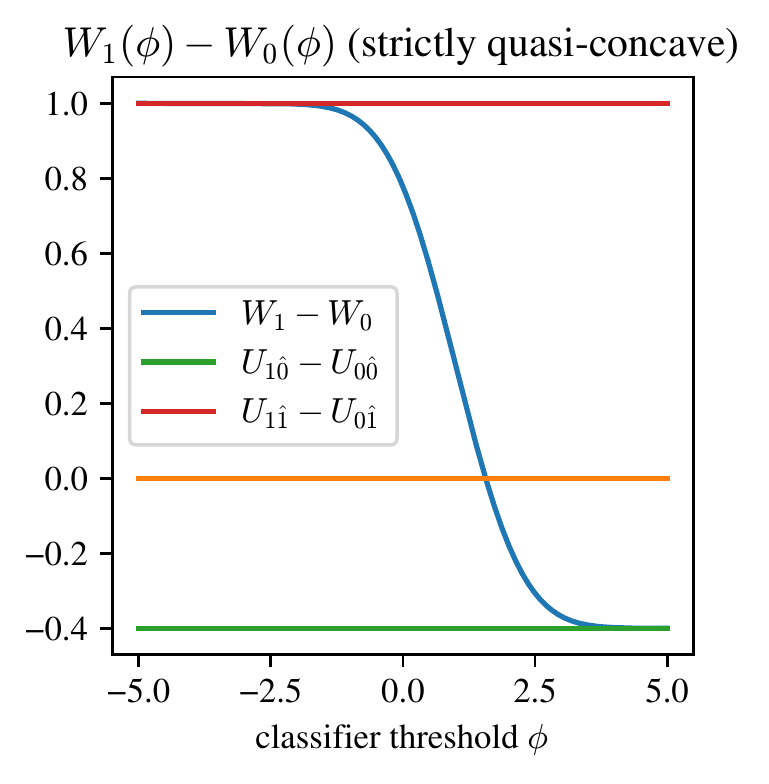}
\end{subfigure}
\hfill
\begin{subfigure}{\textwidth}
    \includegraphics[width=\textwidth, trim=5 5 5 5, clip]{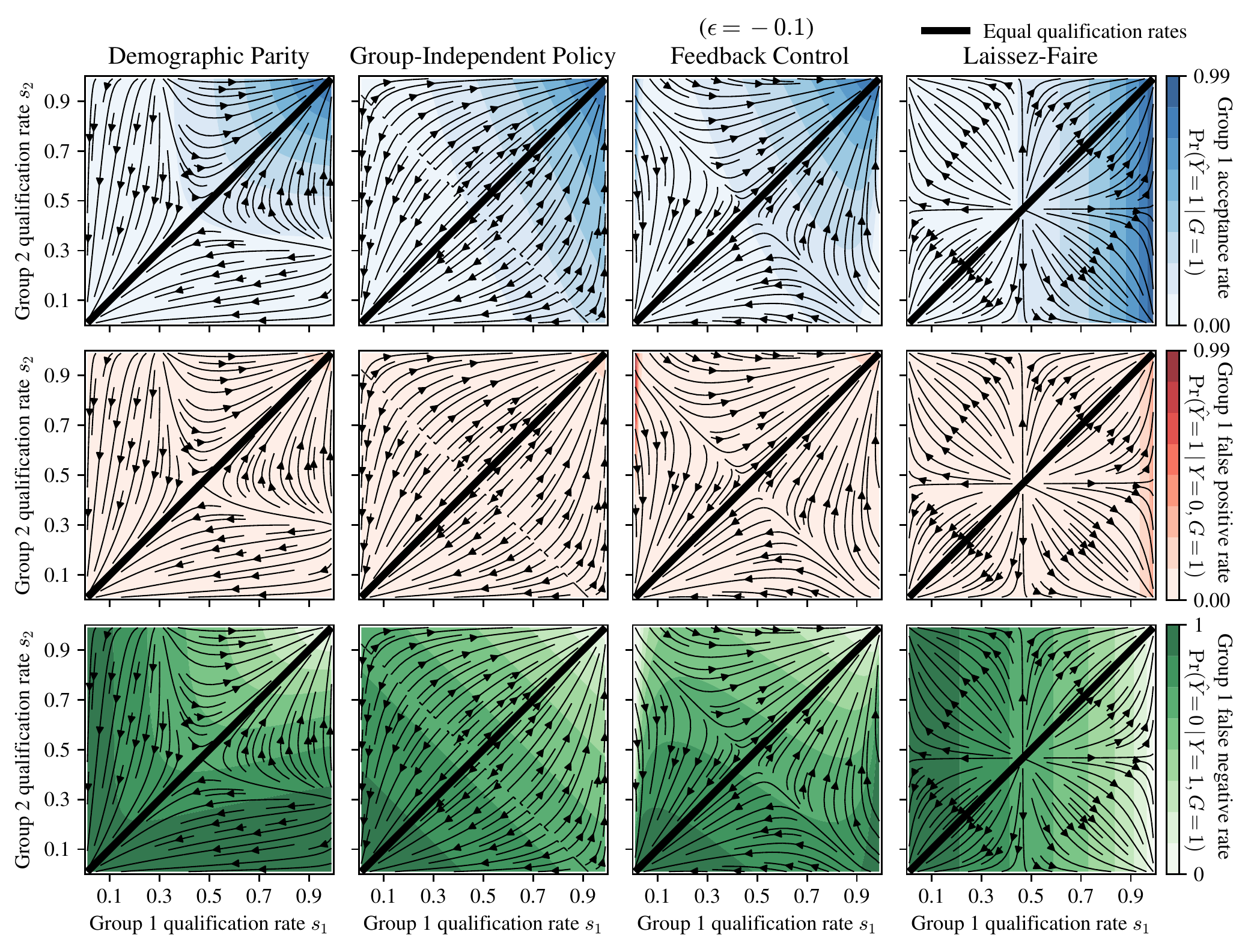}
\end{subfigure}
\caption{\textbf{Setting 3} (Only an unstable hyperplane; Note the negative
  value of \(\epsilon\) for Feedback Control.)}
\end{figure}
\nocaption
\vspace*{\fill}
\newpage
\subsection{Different Group Sizes}
\begin{figure}[H]
\begin{subfigure}{0.98\textwidth}
    \includegraphics[width=\textwidth, trim=5 5 5 5, clip]{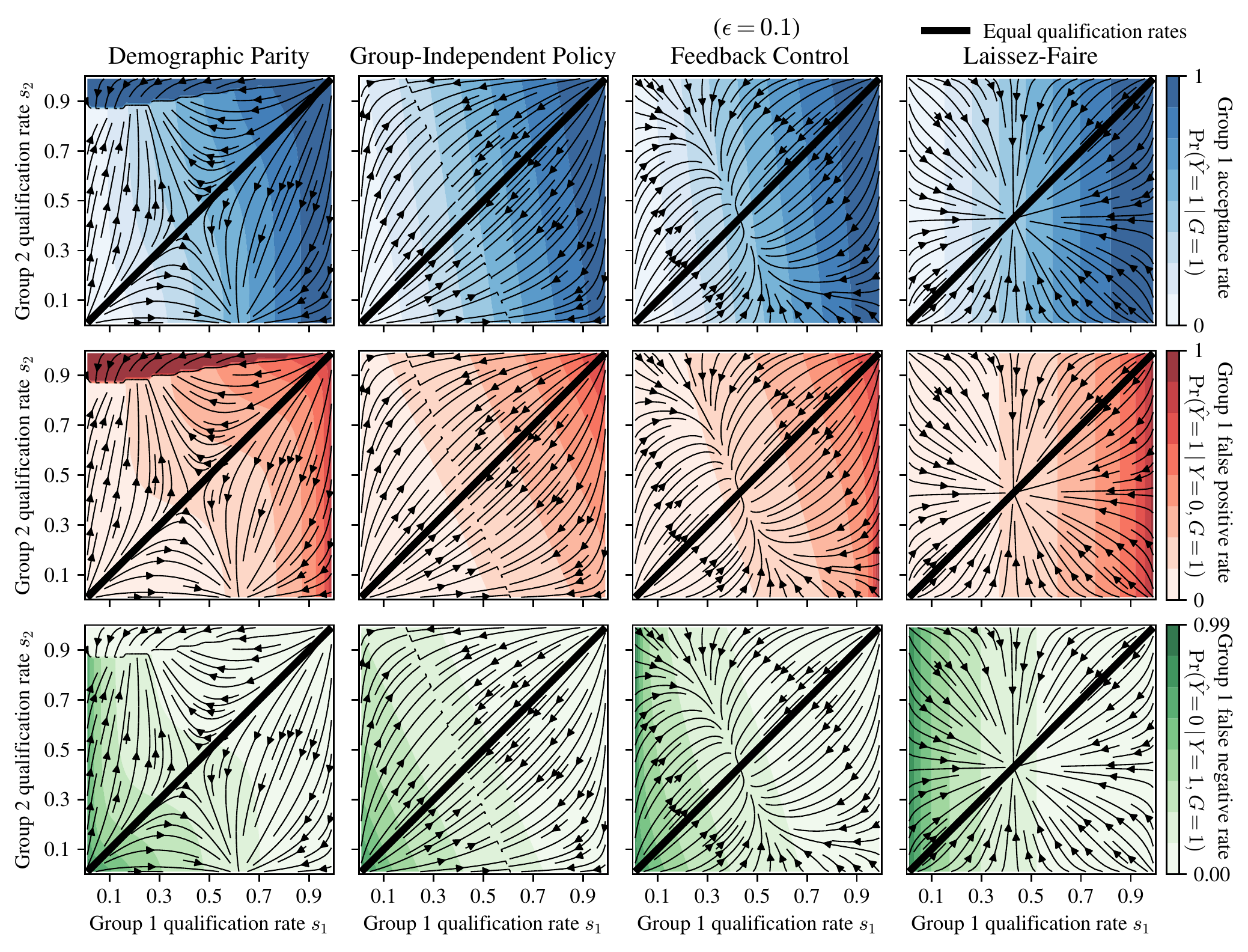}
    \caption{Setting 1 with asymmetric \(\boldsymbol{\mu}\): \(\mu_1 = 0.7,
      \mu_2 = 0.3\)}
\end{subfigure}
\begin{subfigure}{0.98\textwidth}
    \includegraphics[width=\textwidth, trim=5 5 5 5, clip]{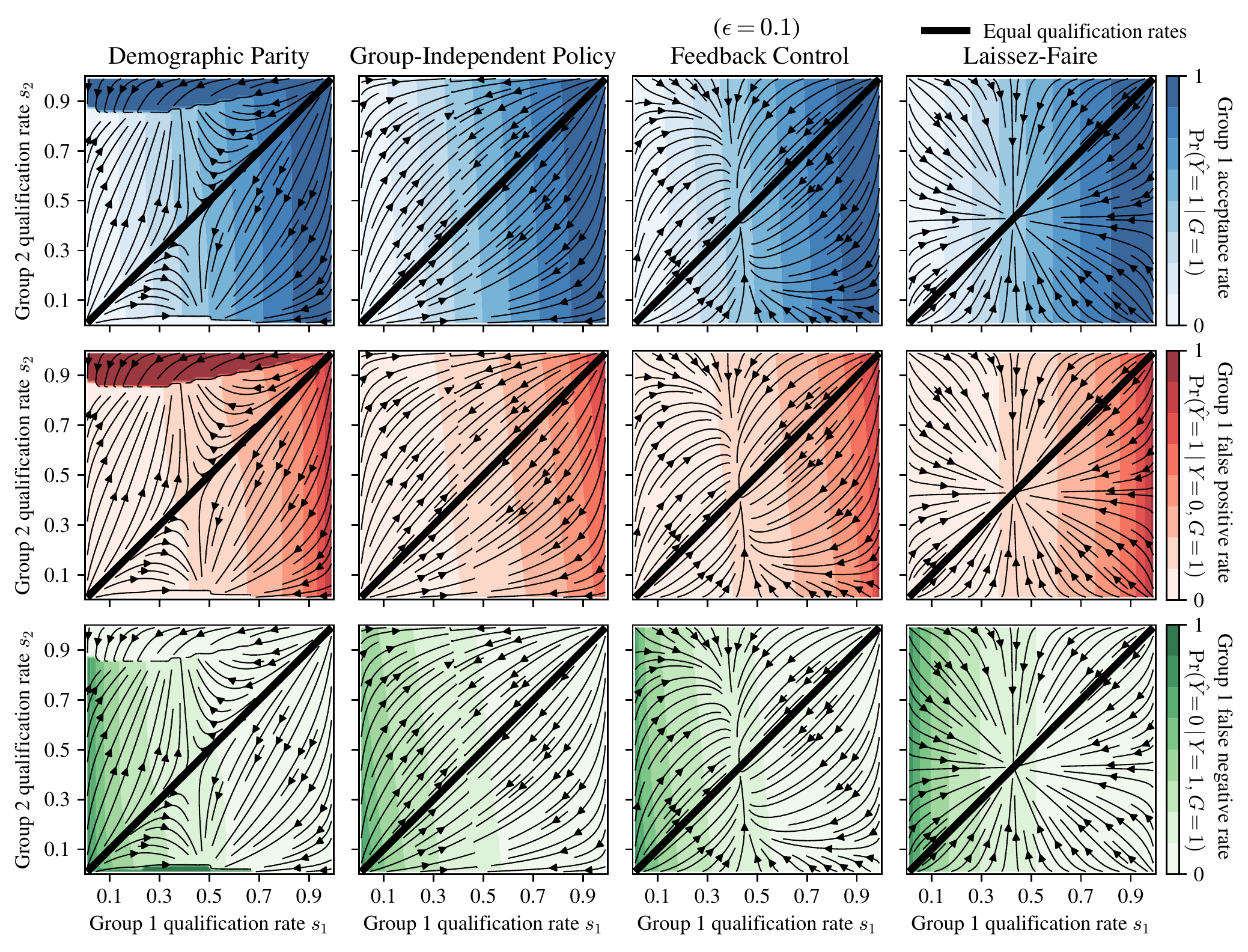}
    \caption{Setting 1 with asymmetric \(\boldsymbol{\mu}\): \(\mu_1 = 0.9,
      \mu_2 = 0.1\)}
\end{subfigure}
\end{figure}
\newpage
\subsection{Limited Space for Acceptance}
\begin{figure}[H]
\begin{subfigure}{0.98\textwidth}
    \includegraphics[width=\textwidth, trim=5 5 5 5,
      clip]{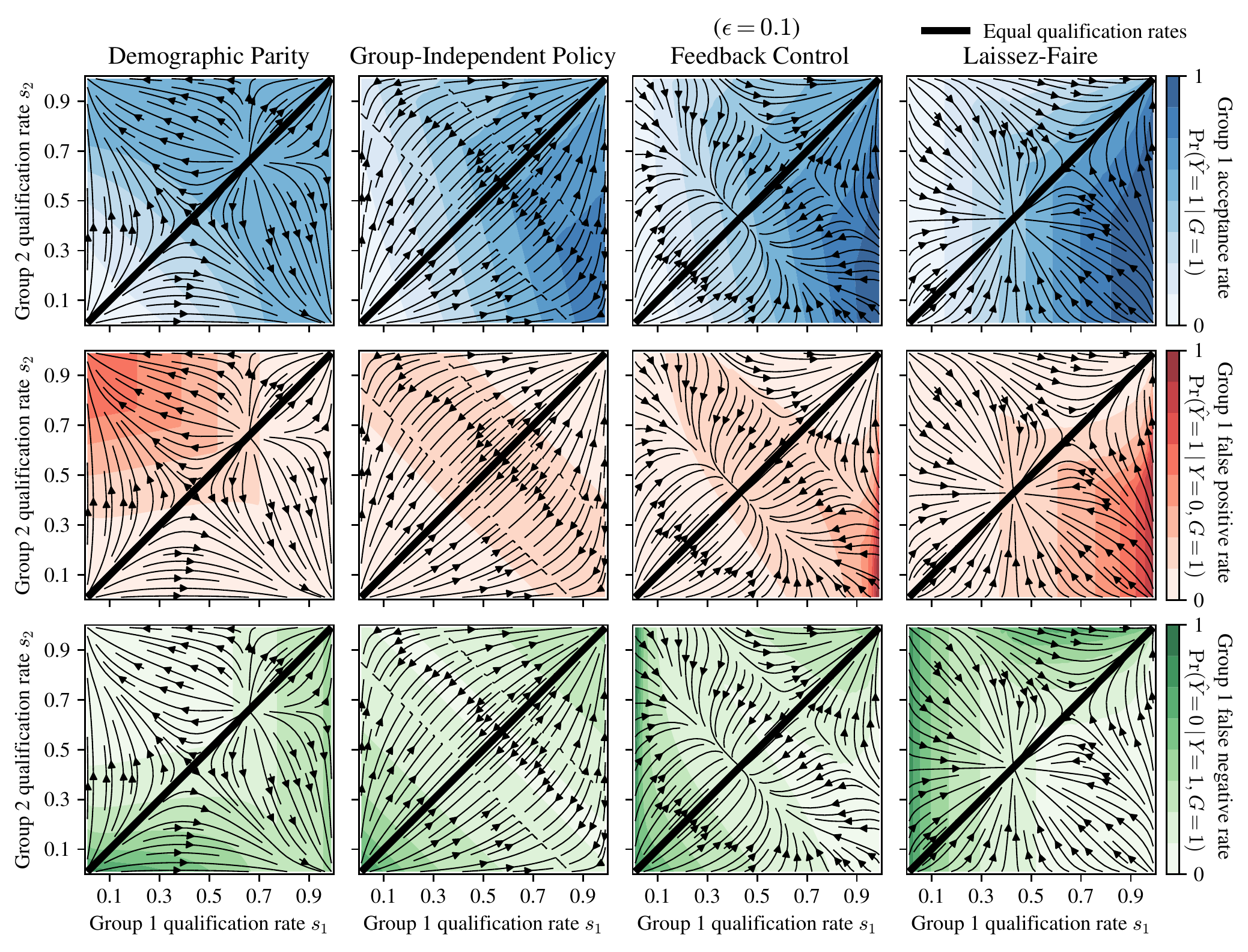}
    \caption{Setting 1, but the classifier is limited to accepting
      \(\Pr(\hat{Y}) < 0.6\)}
\end{subfigure}
\begin{subfigure}{0.98\textwidth}
    \includegraphics[width=\textwidth, trim=5 5 5 5,
      clip]{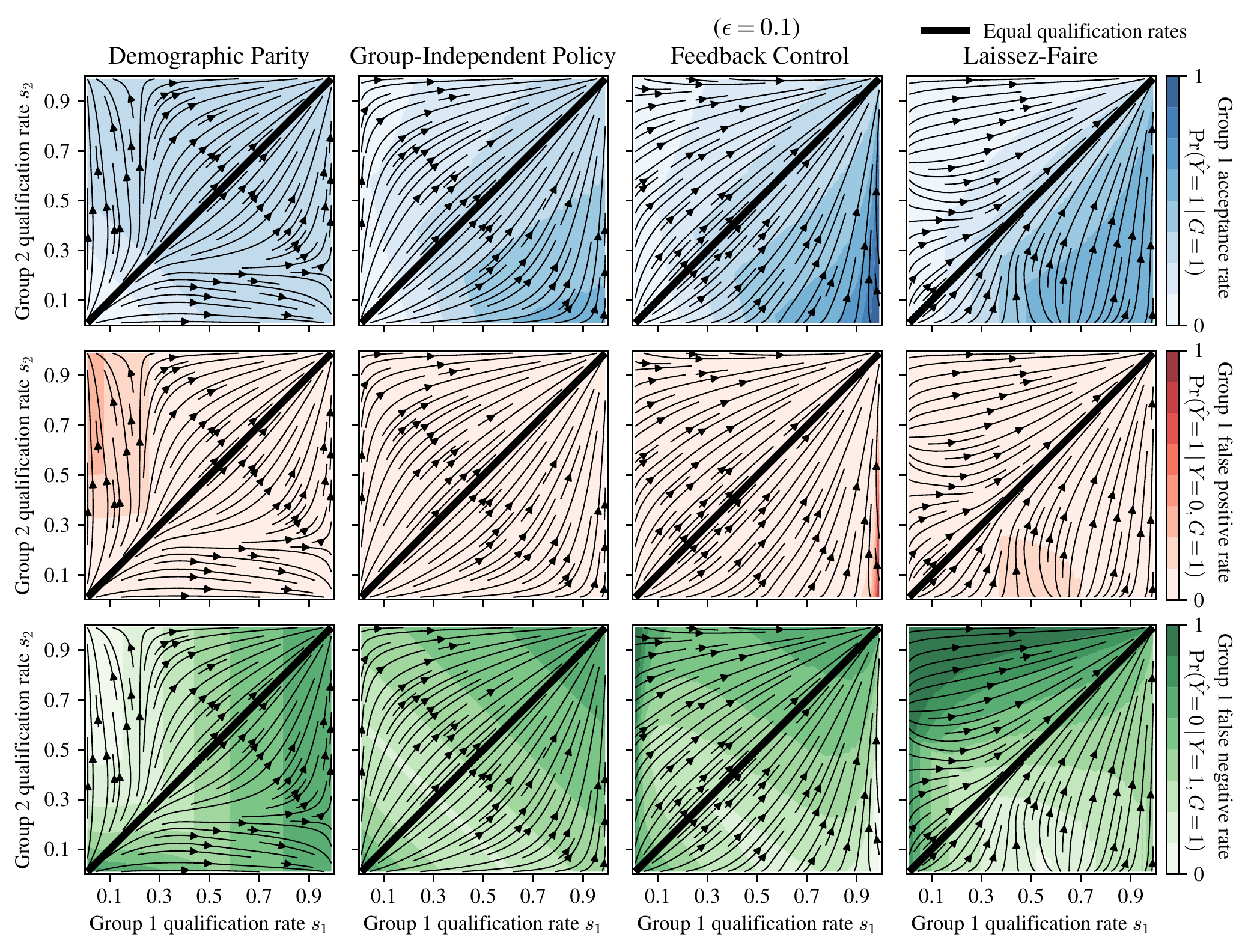}
    \caption{Setting 1, in the classifier is limited to accepting \(\Pr(\hat{Y})
      < 0.3\)}
\end{subfigure}
\end{figure}
\newpage
\subsection{Other Models}

\begin{figure}[H]
\centering
\begin{subfigure}{\textwidth}
    \includegraphics[width=\textwidth, trim=5 5 5 5, clip]{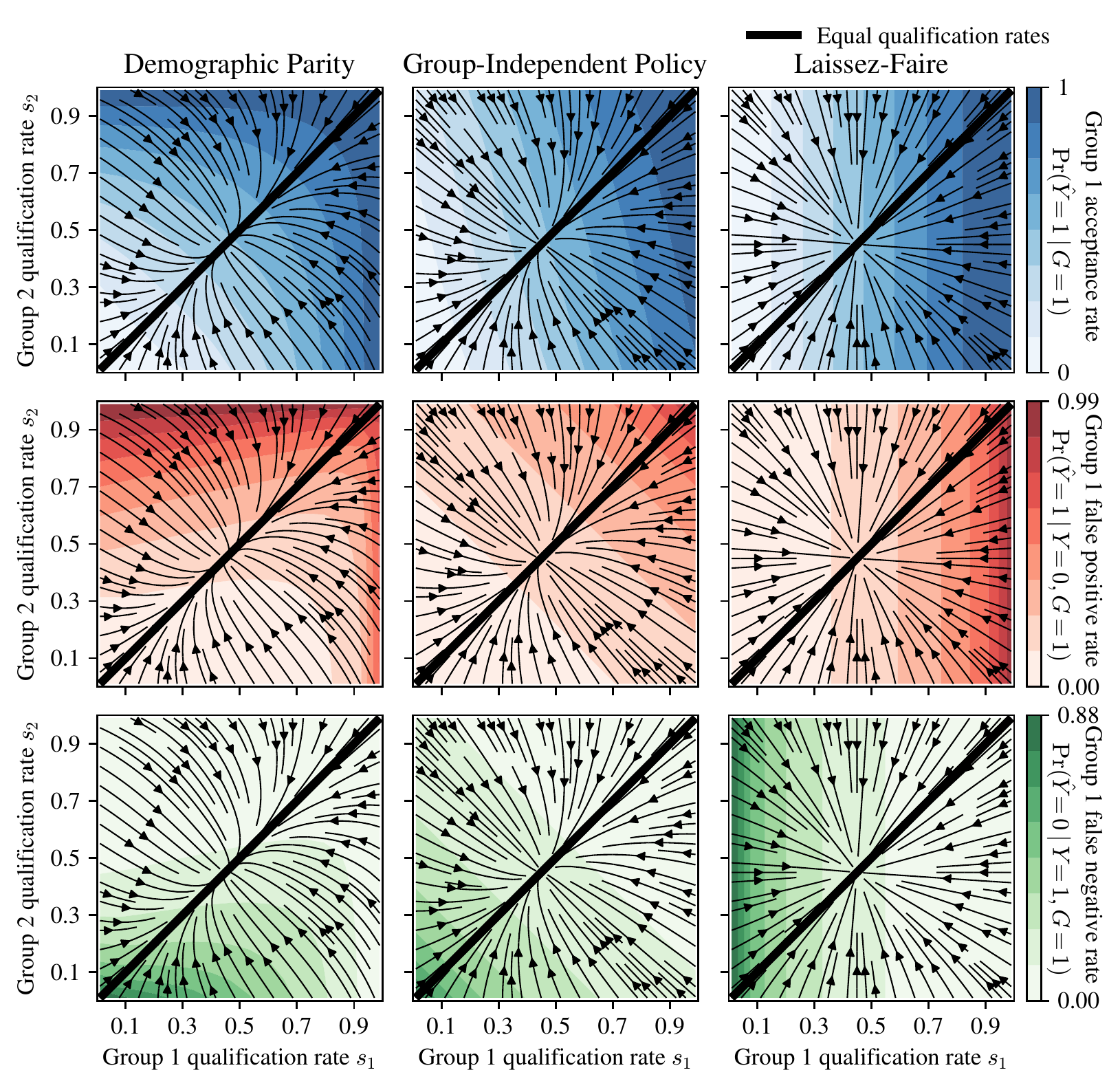}
\end{subfigure}

\caption{The classifier of Setting 1, but the dynamics of \citet{zhang2020fair},
  where the probability of an agent becoming qualified in the next round given
  outcome \(y, \hat{y}\), denoted \(T_{y,\hat{y}}\), given as above.  We assume
  \(T\) is group-independent; under this assumption, disparity in qualification
  rates cannot persist.}
\label{fig:markov}

\end{figure}

\begin{figure}[H]
\hfill
\begin{subfigure}{0.3\textwidth}
\begin{align*}
    &~\begin{bmatrix} \mu_1 = 0.5 & \mu_2 = 0.5
    \end{bmatrix}\\
    &\begin{bmatrix} V_{ 0, \hat{0} } = 0.0 & V_{ 0, \hat{1} } = -500.0 \\ V_{
         1, \hat{0} } = 0.0 & V_{ 1, \hat{1} } = 1.0
    \end{bmatrix}
\end{align*}
  \vfill
\end{subfigure}
\hfill
\begin{subfigure}{0.46\textwidth}
    \includegraphics[width=\textwidth]{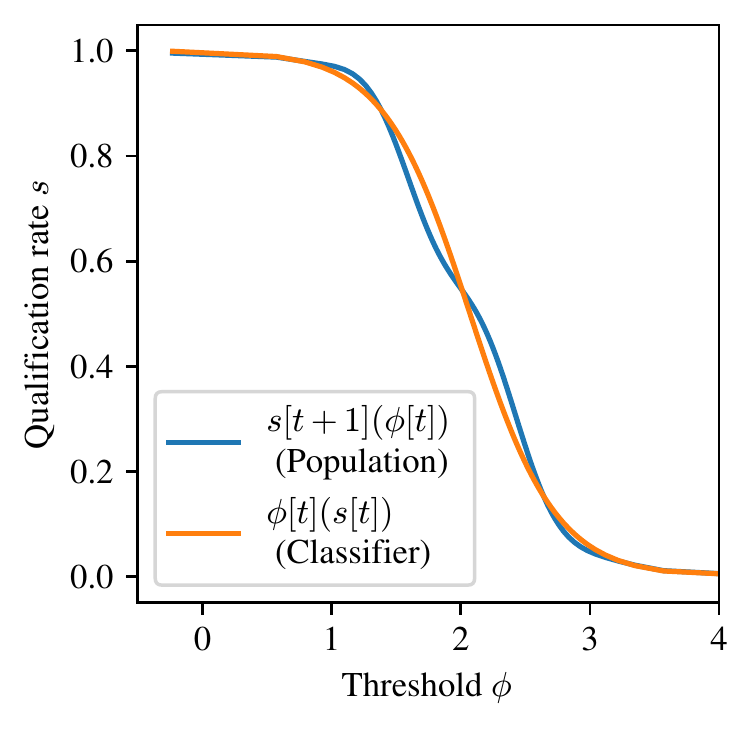}
\end{subfigure}
\hfill
\begin{subfigure}{\textwidth}
    \includegraphics[width=\textwidth, trim=5 5 5 5, clip]{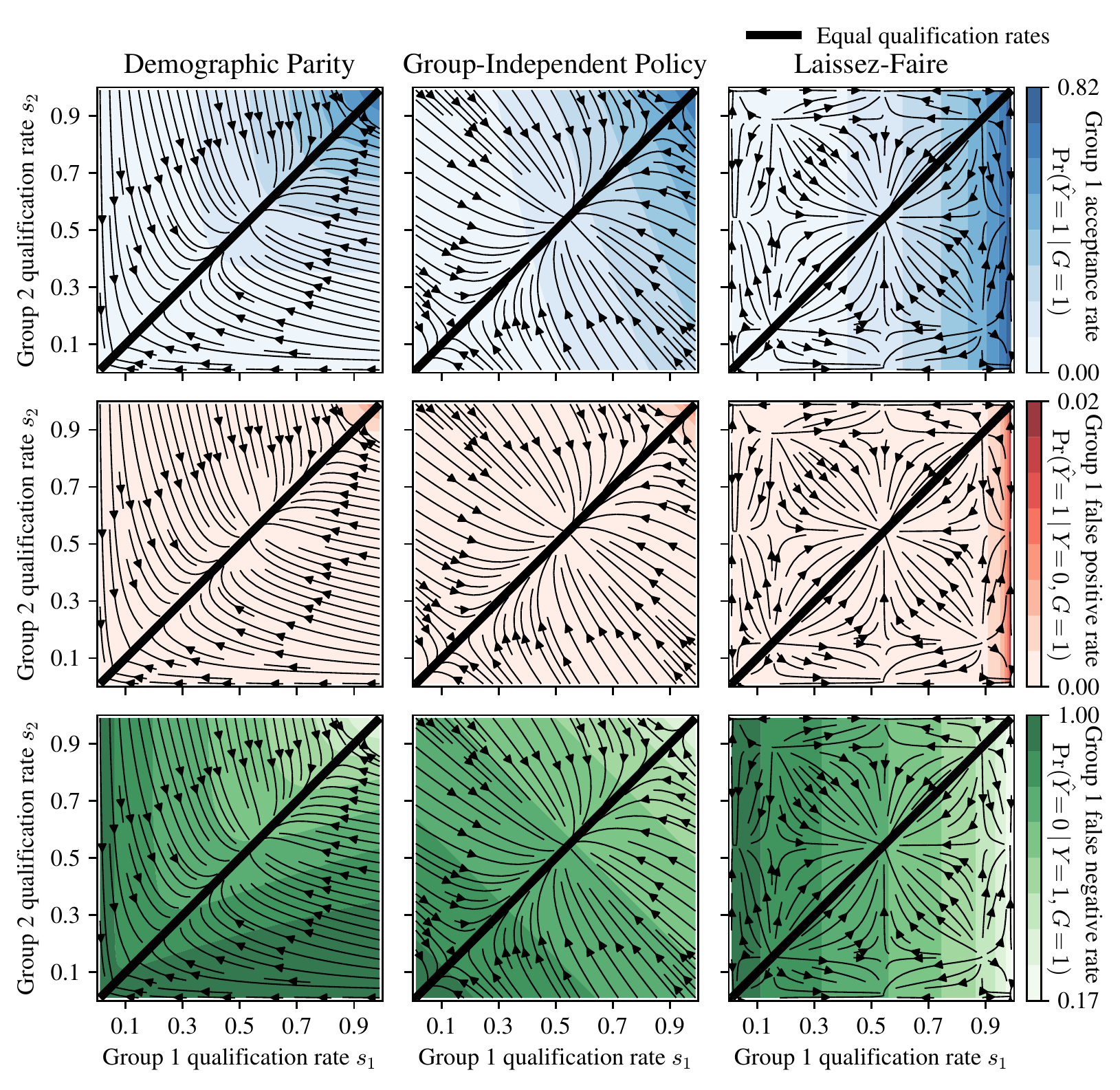}
\end{subfigure}
\caption{The classifier of Setting 1, but the population response model of
  \citet{coate1993will}. The intersections of the curves shown above the phase
  portraits correspond to the possible fixed points of the system in
  qualification rate; these intersections had to be manufactured with a
  distribution of costs, known to agents privately, associated with
  qualification.}
\label{fig:loury}
\end{figure}

For completeness, we picture the dynamics of other models. Specifically, a
Markov model like that of \citet{zhang2020fair} (\cref{fig:markov}) and the
``best response'' model of \citet{coate1993will} (\cref{fig:loury}). Note that
the setting of \citet{coate1993will} assumes that agents privately know their
own costs for becoming qualified, which are sampled from a group-independent
distribution, rather than being uniform for all agents. We use the following set
of parameters.

\begin{align*}
    &~\begin{bmatrix} \mu_1 = 0.5 & \mu_2 = 0.5
    \end{bmatrix}\\
        &\begin{bmatrix} T_{ 0, \hat{0} } = 0.2 & T_{ 0, \hat{1} } = 0.5 \\ T_{
             1, \hat{0} } = 0.1 & T_{ 1, \hat{1} } = 0.8
        \end{bmatrix}\\
    &\begin{bmatrix} V_{ 0, \hat{0} } = 0.0 & V_{ 0, \hat{1} } = -1.0 \\ V_{ 1,
         \hat{0} } = 0.0 & V_{ 1, \hat{1} } = 1.3
    \end{bmatrix}
\end{align*}

\vspace*{\fill}
\vspace*{\fill}

\newpage

\vspace*{\fill}
\vspace*{\fill}

\end{document}